\documentclass{ieeeaccess}
\usepackage{amsmath,amsfonts}
\usepackage{algorithmic}
\usepackage{algorithm}
\usepackage{array}
\usepackage[caption=false,font=scriptsize,labelfont=sf,textfont=sf]{subfig}
\usepackage{textcomp}
\usepackage{stfloats}
\usepackage{url}
\usepackage{verbatim}
\usepackage{graphicx}
\usepackage{cite}
\usepackage{bf_style}
\usepackage{amsmath,amssymb,amsfonts}
\usepackage{bm}
\usepackage{textcomp}
\usepackage{here}
\usepackage{url}
\usepackage[acronym,shortcuts]{glossaries}
\usepackage{multirow}
\usepackage{graphicx}
\usepackage[normalem]{ulem}
\usepackage{booktabs}
\usepackage{graphicx}
\usepackage{subcaption}
\usepackage{booktabs}
\usepackage{threeparttable}
\usepackage{empheq}

\usepackage{multirow}
\usepackage{soul}
\soulregister\ref7 
\soulregister\cite7
\soulregister\textbf7 
\soulregister\ac7 

\newcommand{\down}{\text{down}}
\newcommand{\Imask}{I_M}
\newcommand{\Irange}{I}
\newcommand{\Idepth}{I_D}

\newcommand{\hatImask}{\hat{I}_M}
\newcommand{\hatIrange}{\hat{I}}
\newcommand{\hatIdepth}{\hat{I}_D}

\newcommand{\datasetMask}{\mathcal{D}_M}
\newcommand{\datasetDepth}{\mathcal{D}_D}
\newacronym{2D}{2D}{two-dimensional}
\newacronym{3D}{3D}{three-dimensional}
\newacronym{LiDAR}{LiDAR}{Light Detection and Ranging}
\newacronym{RI}{RI}{range image}
\newacronym{JPEG}{JPEG}{Joint Photographic Experts Group}
\newacronym{JPEG2000}{JPEG2000}{Joint Photographic Experts Group 2000}
\newacronym{INR}{INR}{implicit neural representation}
\newacronym{NNs}{NNs}{neural networks}
\newacronym{MSE}{MSE}{mean squared error}
\newacronym{BCE}{BCE}{binary cross entropy}
\newacronym{PCC}{PCC}{point cloud compression}
\newacronym{GFT}{GFT}{graph Fourier transform}
\newacronym{PCL}{PCL}{point cloud library}
\newacronym{G-PCC}{G-PCC}{geometry-based point cloud compression}
\newacronym{V-PCC}{V-PCC}{video-based point cloud compression}
\newacronym{BT}{BT}{binary-tree}
\newacronym{QT}{QT}{quad-tree}
\newacronym{ReLU}{ReLU}{Rectified Linear Unit}
\newacronym{NeRV}{NeRV}{Neural Representations for Videos}
\newacronym{HEIF}{HEIF}{High-Efficiency Image File Format}
\newacronym{AVIF}{AVIF}{AV1 Image File Format}
\newacronym{COIN}{COIN}{COmpression with Implicit Neural representations}
\newacronym{MPEG}{MPEG}{Moving Picture Experts Group}
\newacronym{CD}{CD}{chamfer distance}
\newacronym{R-D}{R-D}{rate-distortion}
\newacronym{BD-CD}{BD-CD}{Bj{\o}ntegaard delta chamfer distance}
\newacronym{BD-D1 PSNR}{BD-D1 PSNR}{Bj{\o}ntegaard delta D1 PSNR}
\newacronym{BD-D2 PSNR}{BD-D2 PSNR}{Bj{\o}ntegaard delta D2 PSNR}
\newacronym{DCT}{DCT}{discrete cosine transform}
\newacronym{DNN}{DNN}{deep neural network}

\def\BibTeX{{\rm B\kern-.05em{\sc i\kern-.025em b}\kern-.08em
    T\kern-.1667em\lower.7ex\hbox{E}\kern-.125emX}}
\begin{document}
\history{Received 30 December 2025, accepted 7 January 2026, date of publication 14 January 2026, date of current version 22 January 2026.}
\doi{10.1109/ACCESS.2026.3654068}

\title{Range Image-Based Implicit Neural Compression for LiDAR Point Clouds}

\author{\uppercase{Akihiro Kuwabara}\authorrefmark{1}, 
\uppercase{Sorachi Kato}\authorrefmark{1},
\IEEEmembership{Graduate Student Member, IEEE}, \\
\uppercase{Toshiaki Koike-Akino}\authorrefmark{2},
\IEEEmembership{Senior Member, IEEE},
\uppercase{Takuya Fujihashi}\authorrefmark{1}, \IEEEmembership{Member, IEEE}}
\address[1]{Graduate School of Information Science and Technology, The University of Osaka, Japan (e-mail: kuwabara.akihiro@ist.osaka-u.ac.jp)}
\address[2]{Mitsubishi Electric Research Laboratories (MERL), 201 Broadway, Cambridge, MA 02139, USA}
\tfootnote{This work was supported by JST\textperiodcentered ASPIRE Grant Number JPMJAP2432 and JSPS KAKENHI Grant Number JP22H03582.}

\markboth
{Kuwabara \headeretal: Range Image-Based Implicit Neural Compression for LiDAR Point Clouds}
{Kuwabara \headeretal: Range Image-Based Implicit Neural Compression for LiDAR Point Clouds}

\corresp{Corresponding author: Akihiro Kuwabara (e-mail: kuwabara.akihiro@ist.osaka-u.ac.jp).}

\begin{abstract}
This paper presents a novel scheme to efficiently compress Light Detection and Ranging~(LiDAR) point clouds, enabling high-precision 3D scene archives, and such archives pave the way for a detailed understanding of the corresponding 3D scenes. 
We focus on {2D} range images~(RIs) as a lightweight format for representing 3D LiDAR observations.
Although conventional image compression techniques can be adapted to improve compression efficiency for RIs, their practical performance is expected to be limited due to differences in bit precision and the distinct pixel value distribution characteristics between natural images and RIs.
We propose a novel implicit neural representation~(INR)--based RI compression method that effectively handles floating-point valued pixels.
The proposed method divides RIs into depth and mask images and compresses them using patch-wise and pixel-wise INR architectures with model pruning and quantization, respectively. 
Experiments on the KITTI dataset show that the proposed method outperforms existing image, point cloud, RI, and INR-based compression methods in terms of 3D reconstruction and detection quality at low bitrates and decoding latency.
\end{abstract}

\begin{keywords}
LiDAR, Point Clouds, Range Image INR.
\end{keywords}

\titlepgskip=-15pt

\maketitle

\section{Introduction}
\label{sec:introduction}
\PARstart{L}{iDAR} sensors have gained significant attention not only in online applications but also in offline applications. 
In such offline applications, memory-efficient and precise \ac{3D} scenes should be stored in advance and the 3D scenes should be smoothly retrieved from the storage based on the user demand for applications of \ac{3D} scene understanding such as digital archiving, environmental monitoring, navigation, and geological surveying ~\cite{env_moniter,modeling,urban}.
{LiDAR} sensors scan the physical space with the ego-centric coordinate and measure the distance to the closest point on surrounding objects for each angle, allowing the creation of a point cloud with \ac{3D} points corresponding to the intersection of laser beams with objects ahead.
As the resolution of {LiDAR} sensors increases, effectively storing and transmitting {LiDAR} scans becomes a significant challenge, primarily due to the substantial volume of each {LiDAR} sequence. 

Although {LiDAR} scans are typically represented as \ac{3D} point clouds, they can also be expressed as a single-channel image, referred to as a \ac{2D} \ac{RI}~\cite{RI}, where point clouds captured in an ego-centric spherical coordinate system are projected onto a panoramic image.
The x-y axes of the \ac{RI} image correspond to the azimuth and elevation angles in the 3D spherical coordinate system, while each pixel value represents the distance to the corresponding point in that direction.
Whereas 3D point clouds require $3N$ values to represent the locations of $N$ point measurements, \ac{RI}s require only $N$ pixels at most, thus demonstrating their compactness.

We can pursue methods to further compress \ac{RI}s.
One potential solution is to adapt conventional lossy image compression techniques, such as \ac{JPEG}~\cite{jpeg} and \ac{JPEG2000}~\cite{jpeg2000}.
However, these methods are based on integer precision for pixel values, which is incompatible for \ac{RI}s whose pixels are represented by single or double precision floating-point numbers to precisely express the distance to the corresponding point measurements.
We can still utilize these compression methods by adjusting the bit precision of \ac{RI}s to align with them, but this naturally leads to degradation in 3D point cloud reconstruction performance due to inadequate distance resolution.
Another drawback of conventional compression methods lies in their strategy: they use block-based \ac{DCT} and apply coarse quantization to high-frequency components based on human visual perception.
This approach leads to significant degradation in the decoding performance of \ac{RI}s, as \ac{RI}s are characterized by large and sudden changes in the pixel value between foreground objects and distant backgrounds or pixels without any assigned point measurement.

To effectively compress \ac{RI}s while preserving their fine precision and high-frequency changes in pixel value, this paper presents a novel \ac{RI} compression method inspired by \ac{INR}-based image compression technique~\cite{dupont2021coin}. 
\ac{INR}~\cite{sitzmann2020implicit,tancik2020fourier} is a lightweight representation of multidimensional signals by compressing them into shallow \ac{NNs}.
Specifically, \ac{INR} overfits \ac{NNs} with a limited number of parameters to the signals of interest through supervised learning, and the trained parameters become the compressed signal representation by providing the mapping function from signal indices, for example, coordinates on the image plane, to the corresponding signals.
A primary challenge for \ac{INR}-based signal compression is to ensure the precision of high-frequency details while simultaneously managing the constraints imposed by the limited model size.
To address this challenge, we propose an extended \ac{INR} training approach that incorporates both a mask image and an \ac{RI} for point depth information.
The mask image is a binary map that indicates whether each pixel corresponds to a projected 3D point ({one}) or not ({zero}).
We begin by generating a mask image from the \ac{RI}, followed by learning two separate coordinate-to-value mappings using distinct \ac{INR} architectures: one for depth \ac{INR} and one for the mask \ac{INR}.
During decoding, these trained \ac{INR}s reconstruct both the depth and mask images, and the final reconstructed \ac{RI} is obtained by applying the mask to the depth image.
Although the reconstructed depth image may contain values for pixels that do not correspond to any 3D points, applying the mask enforces hard thresholding, effectively removing these artifacts.
This process ensures a high-quality reconstructed \ac{RI}, with sharp edges accurately preserved.


The contributions of our study are three-fold:
\begin{itemize}
    \item To the best of our knowledge, this is the first paper to propose an \ac{INR}-based intra \ac{RI} compression method specifically designed for LiDAR measurements projected onto high-precision floating-point images.
    \item We extend the \ac{INR} compression approach to incorporate the mask \ac{INR} that explicitly represents whether any point measurements are assigned to each pixel on the \ac{RI} or not, allowing efficient elimination of false point estimation on reconstructed depth images in the decoding process. 
    \item We evaluate our proposed method using KITTI dataset~\cite{kitti} and compare its performance with existing baselines including conventional image compression, \ac{PCC}~\cite{bib:vpcc_gpcc}, and \ac{RI}-based and \ac{INR}-based image compression, and show the better \ac{R-D} performance and downstream task quality of our method than the baselines, especially at low bitrates.
\end{itemize}

\section{Related Work}

\subsection{Point Cloud Compression}
The measured distance from LiDAR sensors is usually represented as a 3D point cloud. 
Each point cloud consists of a set of \ac{3D} points, and each point is defined by \ac{3D} coordinates, \textit{i.e.}, (X, Y, Z). 
The graph-based and tree-based compression methods have been proposed to compress coordinate information, that is, geometry information.  
The graph-based methods regard the 3D points as graph signals and define \ac{GFT} for frequency conversion in the graph domain. 
\cite{bib:graph_PCC2,bib:holoplus,bib:kirihara} utilized \ac{GFT} for the geometry compression. 
Other studies~\cite{bib:ueno,bib:gsp_inr} reduce the storage and transmission costs for graph signal reconstruction. 
The tree-based compression method is another popular strategy for compressing geometry information.
The typical way is octree-based representation, such as \ac{PCL} and \ac{G-PCC}~\cite{bib:PCL,bib:vpcc_gpcc}. 
Some recent studies have been proposed to improve the efficiency of geometry compression using traditional signal processing~\cite{bib:quad} and \ac{DNN}~\cite{OctAttention,huang2020octsqueeze} solutions, respectively. 
For example, the study in~\cite{bib:quad} adaptively adopts the \ac{QT} and \ac{BT} block partitions in addition to those of octrees to improve the efficiency of the coding. 

\subsection{LiDAR Range Image Compression}
Many recent works consider projecting the measured {LiDAR} information onto 2D \ac{RI} to represent the measured distance information in a compact format. 
There are two types of input LiDAR information to obtain the corresponding \ac{RI}s: 1) raw packet containing the LiDAR laser IDs~\cite{riddle}, the rotation angle of the LiDAR sensors and the distance values, and 2) 3D point clouds~\cite{wang2022r,bib:inter_rpcc,bib:segmentation}.
Our study utilizes 3D point clouds. 
The obtained \ac{RI}s are then intra-coded~\cite{wang2022r} or inter-coded~\cite{bib:inter_rpcc,bib:segmentation} in lossless and lossy manners. 
Here, intra-coding reduces the spatial redundancy in each \ac{RI}, whereas inter-coding reduces the temporal redundancy across \ac{RI}s. 
In R-PCC~\cite{wang2022r}, which is an intra-coding method, each \ac{RI} can be coded by using a lossless coding method, such as LZ4 and Deflate, to compress the floating-point format.
Our study is designed for \ac{RI} intra-coding and exploits the INR-based compression to represent {LiDAR} measurements in small storage and transmission costs.

\begin{figure*}[t]
  \begin{center}
  \subfloat[Patched depth image and mask image from LiDAR point cloud]
  {\includegraphics[width=\linewidth]{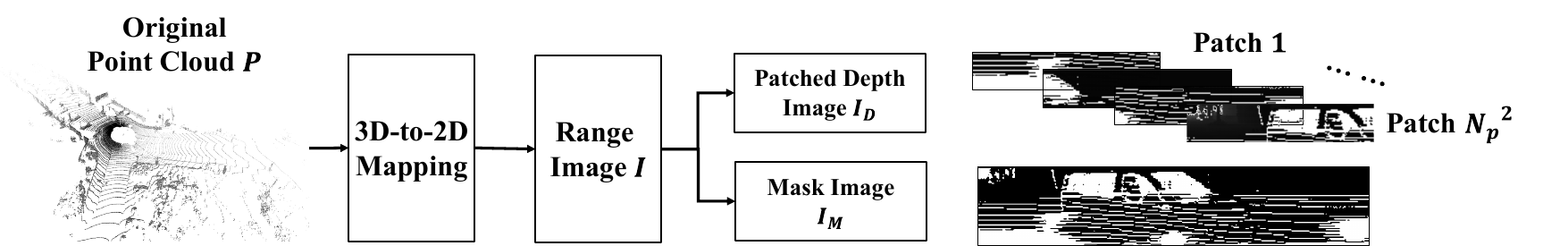}} \\
  \subfloat[Proposed encoder and decoder]
  {\includegraphics[width=\linewidth]{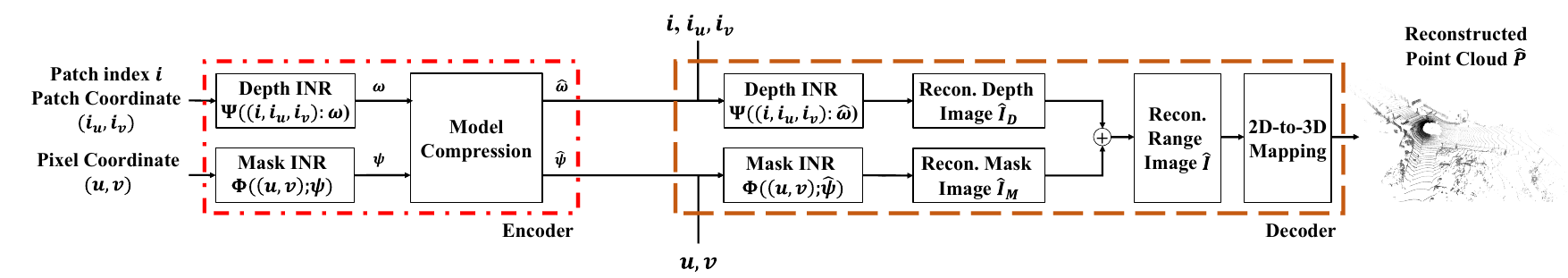}}
  \end{center}
  \caption{Overview of the proposed scheme.}
  \label{fig:proposed_overview}
\end{figure*}

\subsection{Implicit Neural Compression}
Since the concept of \ac{INR} overfits multi-dimensional signals to a small NN architecture, recent studies exploit INR architectures for image compression.
Specifically, each \ac{INR} architecture takes a spatial/time index of the target signals and/or the corresponding feature vector to reconstruct the corresponding attribute values, such as color information.
The overfitted weights of the INR architecture are shared with the receiver's side for signal reconstruction.  

The existing \ac{INR}-based compression can be classified into pixel-wise, patch-wise, and frame-wise architectures. 
The frame-wise architectures realized inter-coding between multiple video frames to remove temporal redundancy. 
They feed the frame index and/or the corresponding embeddings to the \ac{NNs} to generate each frame.
\ac{NeRV}~\cite{chen2021nerv} is the first work on frame-wise video compression, and various extensions~\cite{hinerv,lee2023ffnerv, he2023dnerv,2024VQNeRV,nirvana,NeRI,yan2024ds, Gomes_2023_CVPR,chen2023hnerv,2024boosting} are proposed to improve the quality of reconstruction. 
However, \ac{NeRV} architectures are large models when used for intra-coding each image.  

The pixel-wise INR architecture~\cite{dupont2021coin,dupont2022coin++} takes the pixel index as input and reconstructs the corresponding pixel value. 
The patch-wise INR architecture was first proposed in~\cite{yunpengps-nerv}. Specifically, each image is divided into multiple patches, and the \ac{INR} architecture takes the patch index as input to exploit the similarity of local adjacent pixels for high-quality reconstruction under the same model size. 
A key issue in such INR architectures is the lack of precision in high-frequency details with a small NN architecture. 
To represent high-frequency details under a small NN architecture, SIREN in~\cite{sitzmann2020implicit} argues that sinusoidal activations work better than \ac{ReLU} networks because sinusoidal activations can fit signals contained in higher-order derivatives. 
To address the same problem, our paper extends the training process of INRs by separating \ac{RI}s into depth and mask images.

\section{Proposed Scheme}
\subsection{Overview}
Fig.~\ref{fig:proposed_overview}~shows an end-to-end architecture of the proposed scheme.
Fig.~\ref{fig:proposed_overview}~(a) specifically shows the procedure to obtain \ac{RI} and corresponding depth and mask images from the {LiDAR} 3D point cloud.
We consider that the {LiDAR} measurement to be compressed is a \ac{3D} point cloud consisting of $N$ points, denoted as $\Pbf = \{\pbf_i = [x_i, y_i, z_i] \ | \ i=1, \cdots, N\}$, where $x_i, y_i, z_i \in \mathbb{R}$ represent the Cartesian coordinates of the $i$-th point.
The point cloud is first transformed into the spherical coordinate system.
Subsequently, each point is projected onto a 2D image plane by mapping it to a pixel in a single-channel image $I \in \mathbb{R}^{W \times H}$, producing an \ac{RI}.
{Here, $W$ and $H$ represent the width and height of the range image, respectively.}
The \ac{RI} is then divided into a depth image $\Idepth \in \mathbb{R}^{W \times H}$ and a mask image $\Imask \in \{0, 1\}^{W \times H}$.
The depth image $\Idepth$ is further segmented into small rectangular regions, or patches, with a resolution of $\frac{W}{N_p} \times \frac{H}{N_p}$, where $N_p$ is the scaling factor.

Fig.~\ref{fig:proposed_overview}~(b) shows the sequential operations of the encoder and decoder.
In the encoder, two distinct \ac{INR}s, namely the mask \ac{INR} $\Phi(\cdot;\psibf)$ and the depth \ac{INR} $\Psi(\cdot;\omegabf)$ with learnable parameters $\psibf$ and $\omegabf$, are trained to be overfitted to the depth image and the mask image, respectively.
This training process is a pixel-wise process, which means that the parameters are trained to obtain a mapping from the pixel coordinates or patch indices on the images to their corresponding pixel values.
This is achieved by sequentially providing a pair of indices to the networks.
The well-trained parameters $\psibf$ and $\omegabf$ are subsequently pruned and quantized as $\hat{\psibf}$ and $\hat{\omegabf}$ to enhance their compactness, and we assume that these parameters are stored in storage or transmitted to content receivers as the lightweight format of the {LiDAR} measurements.
In the decoding process, the decoder uses compressed parameters to reconstruct a mask image $\hatImask$ and a depth image $\hatIdepth$ from individual INR architectures $\Phi(\cdot;\hat{\bm{\psi}})$ and $\Psi(\cdot;\hat{\bm{\omega}})$, respectively.
Similarly to the encoding process, the images are reconstructed by sequentially feeding a pair of coordinates and indices to the \ac{INR}s and collecting all estimated values to form the shape of the image.
We obtain the final result of \ac{RI}, $\hatIrange$, by applying the mask image to the depth image to mask out any values of the pixels in the depth image corresponding to the pixels with a mask value of {0}, indicating ``no point''.
Finally, the {LiDAR} point cloud is reconstructed as $\hat{\mathbf{P}}$ from $\hatIrange$ via a reverse coordinate projection process from the 2D image plane, through the spherical coordinate system, to the Cartesian coordinate system.


\subsection{3D-to-2D Mapping}
Our proposed method first performs a coordinate transformation for all points in the 3D point cloud $\Pbf$ measured by {LiDAR} sensors to obtain a \ac{2D} \ac{RI} $I$.
Specifically, the 3D-to-2D mapping consists of two steps: 1) mapping points in the 3D Cartesian coordinate system $x$-$y$-$z$ to the spherical coordinate $\rho$-$\phi$-$\theta$, and 2) mapping points in the spherical coordinate $\rho$-$\phi$-$\theta$ to an image coordinate system $u$-$v$.

Each 3D point in the point cloud $\pbf \in \Pbf$ consists of the \ac{3D} Cartesian coordinate $(x,y,z)$ first.
This point is transformed into a point in the spherical coordinate $\pbf^\prime = (\rho, \phi, \theta)$, where $\rho, \phi, \theta $ denotes the length, pitch, and yaw of the coordinate system, as follows:
\begin{align}
\rho = \sqrt{x^2 + y^2 + z^2}, \;
\phi = \arcsin\left(\frac{z}{\rho}\right), \; \theta = \arctan\left(\frac{y}{x}\right).
\label{eq:rho}
\end{align}
The point in the spherical coordinate is further transformed to the image coordinate $(u, v)$ to generate \ac{2D} \ac{RI} $I$ as follows:
\small
\begin{align}
u &= \left \lfloor \frac{W}{2} \times \left(\frac{\theta}{\pi} + 1\right) \right \rfloor, \nonumber \\
v &= \left \lfloor H \times \left(1 - \frac{\phi + |\phi_\mathrm{down}|}{\phi_\mathrm{up} + |\phi_\mathrm{down}|}\right) \right \rfloor,
\label{eq:v}
\end{align}
\normalsize
where $\phi_\mathrm{up}$ and $\phi_\mathrm{down}$ are the maximum and minimum value of $\phi$ in the dataset, $| \cdot |$ is the absolute value, and $\lfloor \cdot \rfloor$ is a floor function.
$H$ and $W$ in Eq.~(\ref{eq:v}) are the height and width of \ac{RI}, and they are determined by the angular resolution of the {LiDAR} sensor for the elevation and azimuth axes. 
In this study, we set $H = 64$ and $W = 1024$.
The value of each pixel $I(u,v)$ on the \ac{RI} is the measured distance $\rho$, derived in Eq.~(\ref{eq:rho}), with arbitrary unit for physical length.

Due to the sparsity of {LiDAR} measurements, not all pixels on the \ac{RI} are guaranteed to be assigned to any 3D point.
Therefore, if a pixel on $(u^\prime, v^\prime)$ remains unassigned after performing the 3D-to-2D mapping for all points, we set $I(u^\prime, v^\prime) = \rho_{\rm null}$ where $\rho_{\rm null}$ is the arbitrary value indicating that no 3D point is assigned to the pixel.
In practice, $\rho_{\rm null}$ should be selected to be greater than the maximum value of $\rho$ in the {LiDAR} measurements or a negative value.

\subsection{Depth/Mask Image Construction}
After 3D-to-2D mapping, the \ac{RI} is then divided into a mask image $\Imask \in \{0, 1\}^{W \times H}$ and a depth image $\Idepth \in \mathbb{R}^{W \times H}$.

The mask image $\Imask$ is to indicate whether any 3D point is assigned to each pixel on the \ac{RI} or not and is defined as:
\begin{empheq}[]{align}
\Imask(u,v) = 
\begin{cases}
1 & \text{if } \Irange(u,v) \neq \rho_{\rm null}, \\
0 & \text{otherwise}.
\end{cases}
\end{empheq}
Given the mask image, we construct a dataset $\datasetMask$ for training the mask \ac{INR} $\Phi(\cdot;\bm{\psi})$ which consists of pairs of the coordinates of pixels and corresponding binary values as
\small
\begin{align}
\datasetMask = \{((u,v), \Imask(u,v)) \mid \ & u \in \{1, \ldots, W\}, v \in \{1, \ldots, H\} \}.
\end{align}
\normalsize

The depth image $\Idepth$ is the masked version of the \ac{RI}.
Pixels without any 3D point assignment are considered as ``Do not care'' ($\emptyset$) and are defined as such.
\begin{align}
\Idepth(u,v) = 
\begin{cases} 
\emptyset & \text{if } \Irange(u,v) = \rho_{\rm null}, \\
\Irange(u,v) & \text{otherwise}.
\end{cases}
\end{align}
In addition, we divide the depth image into small rectangular areas, or patches, inspired by recent works~\cite{yunpengps-nerv} to improve the decoding performance and the quality of the reconstructed depth image.
Specifically, the \ac{RI} is evenly segmented into patches $\Idepth^\prime (i) \in \mathbb{R}^{\frac{W}{N_{p}} \times \frac{H}{N_{p}}}$, where $N_p$ is a scaling factor and $i = 1, \cdots, N_p^2$.
Each patch is assigned a patch index, allowing us to specify a pixel in the patched RI as $\Idepth^\prime(i, i_u, i_v)$, where $i$ represents the patch index and $i_u, i_v$ denotes the in-patch pixel coordinates whose origin is the top left pixel in the $i$-th patch.
Similarly to the mask image, we also construct a dataset $\datasetDepth$ for training depth INR $\Psi(\cdot;\bm{\omega})$ which consists of pairs of a patch index, in-patch coordinates, and the corresponding depth values, excluding unassigned pixels, as follows:
\begin{align}
\datasetDepth = \{ ((i,i_u,i_v), \Idepth(i,i_u,i_v)) \mid \ & i \in \{1, \ldots, N_p^2\}, \nonumber \\
& i_u \in \{1, \ldots, W/N_p\}, \nonumber \\ 
& i_v \in \{1, \ldots, H/N_p\}, \nonumber \\
& \Idepth(i,i_u,i_v) \neq \emptyset \}
\end{align}

\begin{figure}[t]
  \begin{center}
  \subfloat[Mask INR]
  {\includegraphics[width=0.8\linewidth]{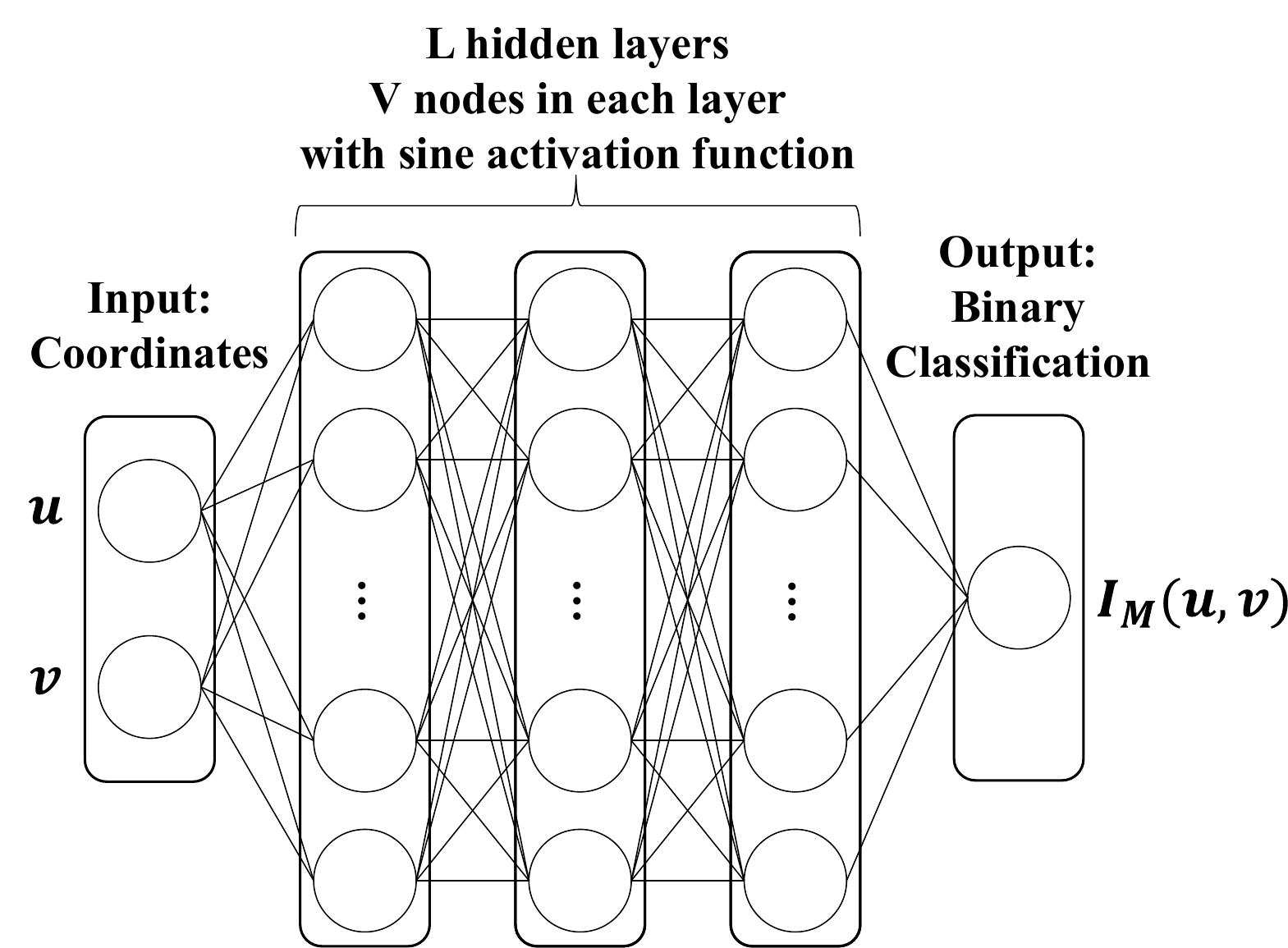}} \\
  \subfloat[Depth INR]
  {\includegraphics[width=0.8\linewidth]{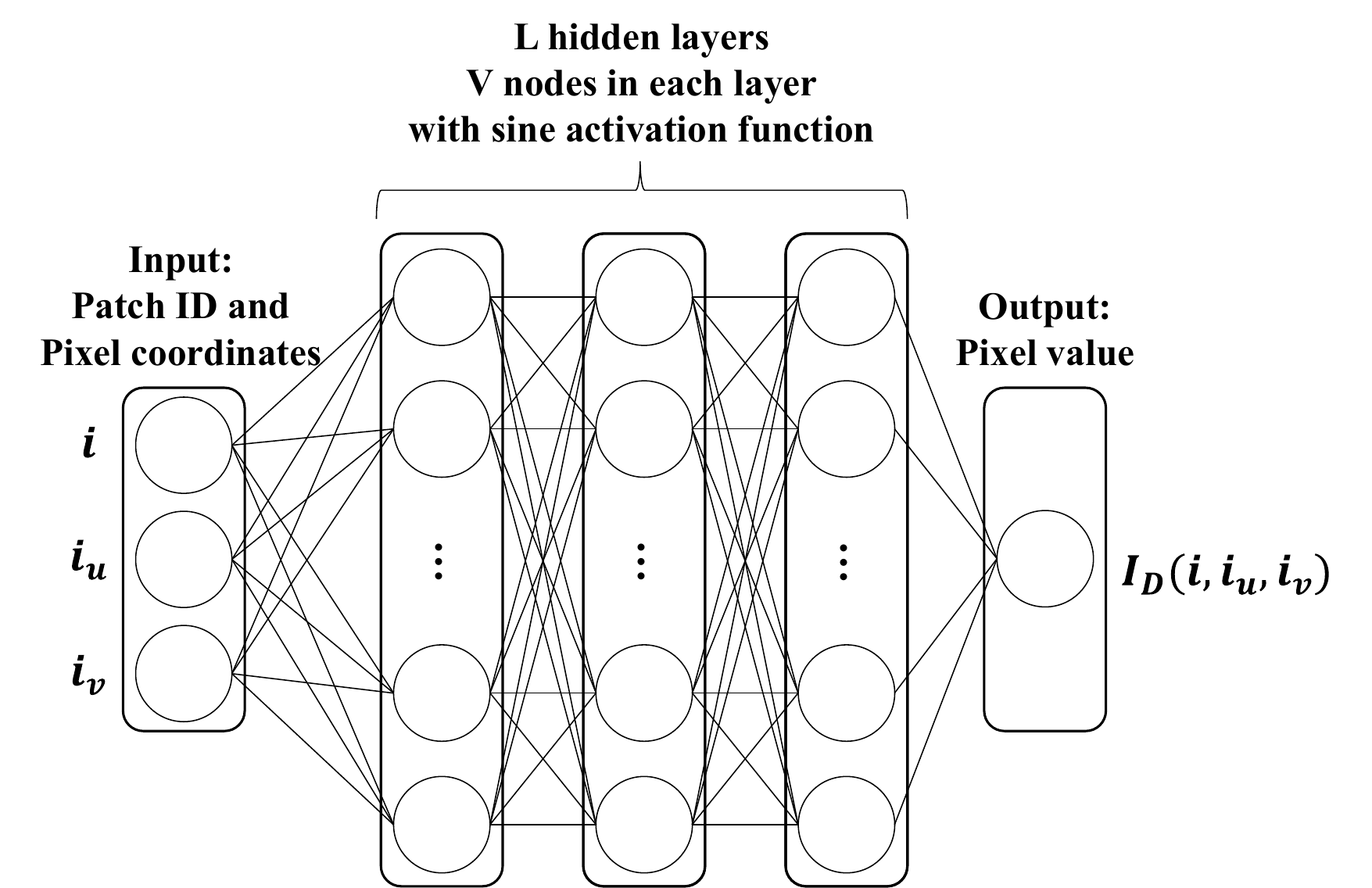}}
   \caption{Architectures of the proposed mask and depth INRs.}
   \label{fig:INR}
  \end{center}

\end{figure}

\subsection{INR-based RI Encoder}
In the encoding process, the mask \ac{INR} $\Phi(\cdot; \psi)$ and the depth \ac{INR} $\Psi(\cdot; \omega)$ are trained to obtain good parameters to express the coordinate-to-value relationships contained in the mask dataset $\datasetMask$ and $\datasetDepth$.
Figs.~\ref{fig:INR}~(a) and (b) show the detailed architecture of the proposed depth INR and mask INR, respectively.

\subsubsection{Mask INR}
Regarding the mask image, we assume the existence of a function $\Phi_M$, which maps each coordinate on the image to a binary value as
\begin{equation}
    \Phi_M: \mathbb{R}^2 \longrightarrow \{0, 1\},
\end{equation}
and the objective of the mask \ac{INR} is to obtain parameters $\psibf$ that well approximate that mapping function as $\Phi(\cdot;\psibf) \approx \Phi_M$ through supervised learning with the dataset $\datasetMask$.
Fig.~\ref{fig:INR}~(a) shows the detailed architecture of the proposed mask INR.  
The mask INR is a multi-layer perceptron~(MLP) with $L$ hidden layers and $V$ nodes, and after each hidden layer, a sinusoidal function layer is used as the activation function.
The network sequentially receives the coordinate $(u, v)$ from the mask dataset $\datasetMask$ and regresses a binary value as its output.
Regression loss is computed using \ac{BCE} loss function between all output values $\Phi((u,v);\bm{\psi})$ and the corresponding true values $\Imask(u, v)$ as follows:
\begin{align}
    \mathcal{L}_{\mathrm{BCE}}(\bm{\psi}) = - \frac{1}{HW} \sum_{u}^{H} \sum_{v}^{W} [\Imask(u,v) \log \left(\Phi((u,v);\bm{\psi})\right) \nonumber \\ + 
    (1-\Imask(u,v)) \log \left(1-\Phi((u,v);\bm{\psi})\right)].
\end{align}

\subsubsection{Depth INR}
Similar to the mask INR, we also assume the existence of another function $\Psi_D$ regarding the depth image, which maps the pair of a patch index and an in-patch pixel coordinate to a depth value as
\begin{equation}
    \Psi_D: \mathbb{R}^3 \longrightarrow \mathbb{R}^1.
\end{equation}
and the objective of the depth \ac{INR} is to obtain parameters $\omegabf$ for good approximation of $\Psi_D$ as $\Psi((i,i_u,i_v);\omegabf) \approx \Psi_D$.
Figs.~\ref{fig:INR}~(b) shows the detailed architecture of the proposed depth INR.
The depth \ac{INR} shares its structure with the mask \ac{INR}, with the exception of the input layer, which accepts 3 values.
In the training process, pairs of patch index $i$ and an in-patch coordinate $(i_u, i_v)$ are sequentially passed to the depth INR network from the depth dataset $\datasetDepth$, and the corresponding depth values are regressed.
We employ \ac{MSE} loss as a regression loss function for the depth \ac{INR} as

{
\footnotesize
\begin{align}
\mathcal{L}_{\mathrm{MSE}}(\omegabf) = \frac{1}{HW} \sum_i^{N_p^2} \sum_{i_u}^{W/N_p} \sum_{i_v}^{H/N_p} \|\Psi((i,i_u,i_v);\omegabf) - \Idepth(i,i_u,i_v)\|^2.
\label{eq:mse}
\end{align}
}

\subsection{Model Compression}
Parameters $\psibf$ and $\omegabf$ become effectively compressed representations of the depth and mask images after a thorough training.
We introduce a series of parameter compression processes for both to further improve their compactness.

\subsubsection{Model Pruning}
As an initial step in our parameter compression procedure, we implement global unstructured pruning for parameters in both depth and mask \ac{INR}s.
Given a threshold $w_q$ for the magnitude of parameters, each parameter $w$ is determined to be retained or pruned based on the following criteria:
\begin{align}
    \hat{\omegabf} = 
    \begin{cases}
        \bm{\omega} & \bm{\omega} \geq \bm{\omega}_q, \\
        \mathbf{0} & \mathrm{otherwise}. 
    \end{cases}
\end{align}
To guarantee that the pruned parameters are of good expression, we subsequently retrain the parameters to fine-tune using the same dataset $\datasetMask$ and $\datasetDepth$.

\subsubsection{Model Quantization and Encoding}
The pruned and fine-tuned parameters are uniformly quantized to a bit depth of $N_b$.
This quantization is layer-wise, meaning that given a parameter set corresponding to each layer in the depth and mask INRs as $\mubf \in \hat{\bm{\omega}}$, a quantized parameter set $\mubf_q$ is obtained as follows:
\begin{align}
    \mubf_q = \mathrm{round}\left(\frac{\mubf - \mubf_{\min}}{2^{N_b}}\right) s + \mubf_{\min}, \; s = \frac{\mubf_{\max} - \mubf_{\min}}{2^{N_b}},
\end{align}
where $\mathrm{round}(\cdot)$ is a rounding function to the nearest integer and $\mubf_{\max}$ and $\mubf_{\min}$ are the maximum and minimum values in $\mubf$.
The quantized tensor $\mubf_q$ is finally coded into a binary sequence using Huffman coding.
It is noteworthy that the quantized parameters $\mubf_q$ are likely to assume values near zero, particularly for smaller bit depths.
Consequently, Huffman coding demonstrates its effectiveness in reducing the overall size of encoded parameters.

\subsection{RI Decoder}

The decoding process of \ac{RI} is a simple feedforward process involving the mask \ac{INR} and depth \ac{INR} with optimized parameters $\hat{\psibf}$ and $\hat{\omegabf}$.
The mask image is reconstructed by feeding the coordinate sets $\{(u, v) \ | \ u\in \{1,\cdots, W\}, v\in \{1, \cdots, H\}\}$ to the mask \ac{INR}.
The resulting binary values are then reshaped to construct the $W \times H$ mask image $\hatImask$.

The depth image reconstruction is in a two-stage manner.
Each patch is first reconstructed by feeding the sets of pairs of a patch index and an in-patch coordinate $\{(i, i_u, i_v) \ | \ i\in \{1, \cdots, N_p^2\}, i_u \in \{1,\cdots, W/N_p\}, i_v\in \{1, \cdots, H/N_p\}\}$ to the depth \ac{INR}.
The reconstructed patches are then gathered to build the complete depth image $\hatIdepth$.
Finally, the reconstructed RI $\hat{I}$ is obtained as
\begin{empheq}[]{align}
    \hat{I}(u, v) =
    \begin{cases}
        \hatIdepth(u,v) & \hatImask(u,v) = 1, \\
        \rho_{\mathrm{null}} & \mathrm{otherwise}.
    \end{cases}
\end{empheq}

\subsection{2D-to-3D Mapping}

The concluding phase of our decoding procedure is the reconstruction of a 3D point cloud through a 2D-to-3D mapping against the reconstructed \ac{RI}.
When the pixel $(u,v)$ on the \ac{RI} has a valid point depth, i.e., is not $\rho_{\rm null}$, the corresponding point in the spherical coordinate $\hat{\pbf}^\prime = (\hat{\rho}, \hat{\phi}, \hat{\theta})$ is obtained as follows:
\begin{align}
\hat{\rho} & = \hatIrange(u,v), \nonumber \\
\hat{\phi} & = \left(1 - \frac{v}{H}\right) \left(\phi_{up} + |\phi_\down|\right) - |\phi_\down|, \nonumber \\ 
\hat{\theta} & = - \left(2 \frac{u}{W} - 1\right)  \pi.
\end{align}
Finally, the 3D points in the spherical coordinate are transformed into the 3D Cartesian coordinate $\hat{\pbf}=(\hat{x}, \hat{y}, \hat{z})$ as
\begin{align}
\hat{x} = \hat{\rho}\cos\hat{\phi}\cos\hat{\theta}, \;
\hat{y} = \hat{\rho} \cos\hat{\phi} \sin\hat{\theta} \;,
\hat{z} = \hat{\rho} \sin\hat{\phi}.
\end{align}

\section{Experiments} 

\subsection{Settings}

\noindent \textbf{Dataset:}
We use the KITTI dataset~\cite{kitti} as our source of 3D point cloud data.
For \ac{R-D} performance, we evaluate the KITTI Odometry dataset, using frames 00, 25, 50, 75, and 100 from sequences 00 to 06. 
{For downstream tasks, we use the KITTI 3D Object Detection dataset. 
We split the official training set into 3,712 training samples and 3,769 validation samples, and evaluate detection performance on the validation split to quantify the impact of compression.
We use OpenPCDet~\cite{openpcdet2020} v0.6.0 to train and evaluate three representative detectors: PointPillars~\cite{lang2019pointpillars}, SECOND~\cite{yan2018second}, and PointRCNN~\cite{shi2019pointrcnn}.
}

All data were collected with a Velodyne 64 scanner that features 64 laser scan lines and an azimuth resolution of 0.09 degrees.
{In the proposed scheme, these 3D point clouds are projected into an RI for compression.
Note that projecting points into a range image may cause point loss.
We evaluate the point retention ratio defined as $r_N = |\hat{\mathcal{P}}|/|\mathcal{P}|$, where $\mathcal{P}$ is the original input point cloud and $\hat{\mathcal{P}}$ is the reconstructed point cloud via RI projection and back-projection.
On the KITTI odometry dataset, we observe $r_N = 41.02\% \pm 0.38\%$ using 35 selected frames from sequences 00--06.
Similarly, we observe $r_N = 74.33\% \pm 0.69\%$ on the KITTI object detection validation split of 3,769 frames, measured on the cropped point clouds used as inputs for 3D object detectors.}

\begin{figure*}[t]
  \centering
  {\includegraphics[width=\hsize]{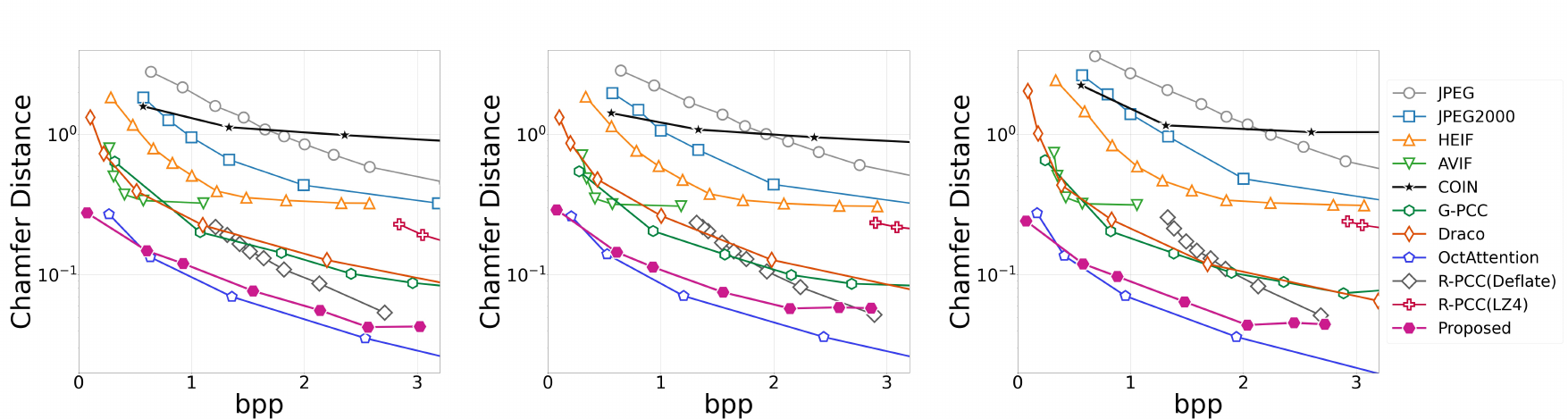}}
    \caption[]{Chamfer distance as a function of bitrate across different sequences of the KITTI LiDAR point clouds{, where the bitrate is measured in bits per point (bpp). From left to right: (left) performance in frame 00 of sequence 00, (middle) performance in frame 25 of sequence 00, and (right) performance in frame 50 of sequence 00.}}
  \label{fig:rate-distortion}
\end{figure*}

\noindent \textbf{Metric:}
Regarding the metrics for the decoded 3D point clouds, we follow the common practice in the community
using \ac{CD}. 
{CD has been widely adopted as a distortion measure for 3D point cloud
reconstruction and sensing~\cite{REF01}.} 

\ac{CD} is defined as:

\small
\begin{align}
\label{eq:CD}
\mathrm{CD} = \frac{1}{2} \bigg\{ \frac{1}{|\mathbf{P}|} \sum_{\bm{p}\in \mathbf{P}} \min_{\widehat{\bm{p}} \in \widehat{\mathbf{P}}} \|\bm{p} - \widehat{\bm{p}}\|_2 +   \frac{1}{|\widehat{\mathbf{P}}|} \sum_{\bm{\widehat{p}}\in \widehat{\mathbf{P}}} \min_{\bm{p} \in \mathbf{P} } \|\bm{p} - \widehat{\bm{p}}\|_2 \bigg\},
\end{align}
\normalsize
where $\mathbf{P}$ is the set of 3D points in the original point cloud and $\widehat{\mathbf{P}}$ is the set of 3D points in the decoded point set.

For the \ac{R-D} performance assessment between the proposed method and the baselines, we use the \ac{BD-CD}~\cite{bjotegaard2001calculation} for calculating average \ac{CD} improvement between \ac{R-D} curves for the same bitrate, where positive values denote \ac{CD} improvement compared to the baselines.
{For downstream tasks, we evaluate 3D object detection accuracy using the Car 3D bounding-box average precision~(AP).
}

\noindent \textbf{Network Architecture Details:}
Both INR architectures are designed to effectively approximate the coordinate-to-value mappings in the mask and depth images derived from \ac{RI}.
The mask INR is an MLP with a fixed depth of $L = 6$ layers.
We experimented with varying the number of nodes $V$ in each hidden layer to evaluate the impact on performance and compression efficiency. 
The values of $V$ considered are $\{10, 19, 24, 28, 31, 34, 37, 40\}$.

The network takes as input the pixel coordinates $(u, v) \in \mathbb{R}^2$ from the mask dataset $\mathcal{D}_{M}$ and outputs a scalar value representing the mask at that coordinate. 
The architecture is structured as follows:
\begin{itemize} 
    \item \textbf{Input Layer}: The coordinates of the pixels $(u, v)$. 
    \item \textbf{Hidden Layers}: It consists of $L=6$ hidden layers, each with $V$ nodes. Each hidden layer employs the sinusoidal activation function to introduce periodicity and enable the network to model high-frequency variations in the mask image.  
    \item \textbf{Output Layer}: A single node with the sigmoid activation function produces an output in the range $(0, 1)$, suitable for binary classification of mask values. 
\end{itemize}

The Depth INR is also implemented as an MLP with a fixed depth of $L = 6$ layers. 
We also consider the different number of nodes $V$ in each hidden layer, choosing $\{28, 31, 34, 37, 40, 42, 45\}$ to evaluate the trade-off between model capacity and compression.

The input to the depth INR is a concatenation of the patch index $i$ and the in-patch pixel coordinates $(i_u, i_v) \in \mathbb{R}^2$, resulting in a 3-dimensional input vector. 
Since we set the patch scaling factor to $N_p = 16$, the patch index $i$ ranges from $0$ to $255$.
The architecture of the depth INR is as follows:
\begin{itemize} 
    \item \textbf{Input Layer}: The concatenated input vector $(i, i_u, i_v)$. 
    \item \textbf{Hidden Layers}: It contains $L=6$ hidden layers, each with $V$ nodes. Similarly to the mask INR, the sinusoidal activation function is applied to each hidden layer to capture the complex variations in the depth image.  
    \item \textbf{Output Layer}: A single node with a linear activation function (identity function) to output the estimated depth value $\hat{I}_{D}(i, i_u, i_v) \in \mathbb{R}$. 
\end{itemize}

\noindent \textbf{Hyperparameter Details:}
We use separate hyperparameter settings for mask and depth INRs. 
The general settings for both INRs include the Adam optimizer, an initial learning rate of $1 \times 10^{-3}$, 3{,}000 training epochs, and a batch size of 1. 
For depth INR, we adopt the cosine annealing scheduler with a warmup phase. The initial learning rate is set to $1 \times 10^{-4}$, and the warmup period lasts for 300 epochs. 
The minimum learning rate is set to $1 \times 10^{-12}$. 

\noindent \textbf{Model Compression Details:}
A global unstructured pruning is used for model pruning. 
The pruning ratio (sparsity) was varied from 0 to 1 to adjust the sparsity of the model parameters. 
For each pruning ratio, we determined the corresponding threshold $\omega_q$ to control which parameters were pruned. 
A higher pruning ratio results in more parameters being set to zero. After pruning, we fine-tuned the model using the same dataset to recover any potential performance loss.

To further compress the pruned and fine-tuned model, we perform uniform quantization to the parameters. 
The quantization bit depth $N_b$ was varied from 4 to 32 bits to balance compression performance and model precision.
The quantized parameters were then encoded using Huffman coding to further reduce the model size.

\noindent \textbf{Baselines:}
We evaluate our proposed method by comparing it with existing baselines in both geometric 3D point cloud compression and 2D image compression.

\begin{enumerate}
    \item[1)]As baselines for 3D point cloud compression, we select \textbf{G-PCC} within the \ac{PCC} family.
We refer to the MPEG reference implementation TMC13-v14.0 for octree geometry compression.

    \item[2)] We also select \textbf{Draco}~\cite{bib:draco} as the 3D point cloud compression baseline which also belongs to the \ac{PCC} family.
    We use the official implementation of the Draco encoder that performs KD-tree-based compression~\cite{kd-tree}.


    \item[3)] 
    {\textbf{OctAttention}~\cite{OctAttention} is an octree-based autoencoder within the \ac{PCC} family.  
    This method improves the conventional octree structure by incorporating attention mechanisms for better context modeling.  
    To evaluate its performance across different compression levels, we set the octree depth to values from 8 to 13.}

    \item[4)] As conventional image compression baselines, we select \textbf{\ac{JPEG}}, \textbf{\ac{JPEG2000}}, \textbf{\ac{HEIF}}, and \textbf{\ac{AVIF}}.
We convert floating-point valued \ac{RI}s into 8-bit precision in advance when using these methods.

    \item[5)] \textbf{R-PCC}~\cite{wang2022r} is an \ac{RI} based {LiDAR} compression baseline.
    It maps {LiDAR} point clouds to \ac{RI}s and performs intra-coding using floating-point lossless coding methods.
    Here, we use LZ4 and Deflate for coding methods due to their fast decompression. 

    \item[6)] \textbf{\ac{COIN}}~\cite{dupont2021coin} is an \ac{INR}--based image compression baseline.
The INR architecture is trained to obtain a direct mapping of the pixel coordinate to each pixel value of \ac{RI}. 
We assume that \ac{COIN} serves as a reliable indicator of the efficiency of our depth/mask separation strategy, as it does not employ the process.
\end{enumerate}


\noindent \textbf{Implementation Detail:}
All the evaluations exhibited in this paper are performed with CPUs of Intel Core i9-10850K and i9-13900KF and with GPUs of NVIDIA GeForce RTX 3080 and 4070.
\ac{NNs} for \ac{COIN} and our proposed method are implemented, trained, and evaluated using PyTorch 2.2.0 with Python 3.10.

\begin{figure*}[t!]
    \centering
    \begin{tabular}{cccccc}
        \begin{minipage}[t]{0.142\hsize}
            \centering
            \subfloat[{Original \\ Seq:~00 \\ Frame:~00}]
            {\includegraphics[width=\hsize]{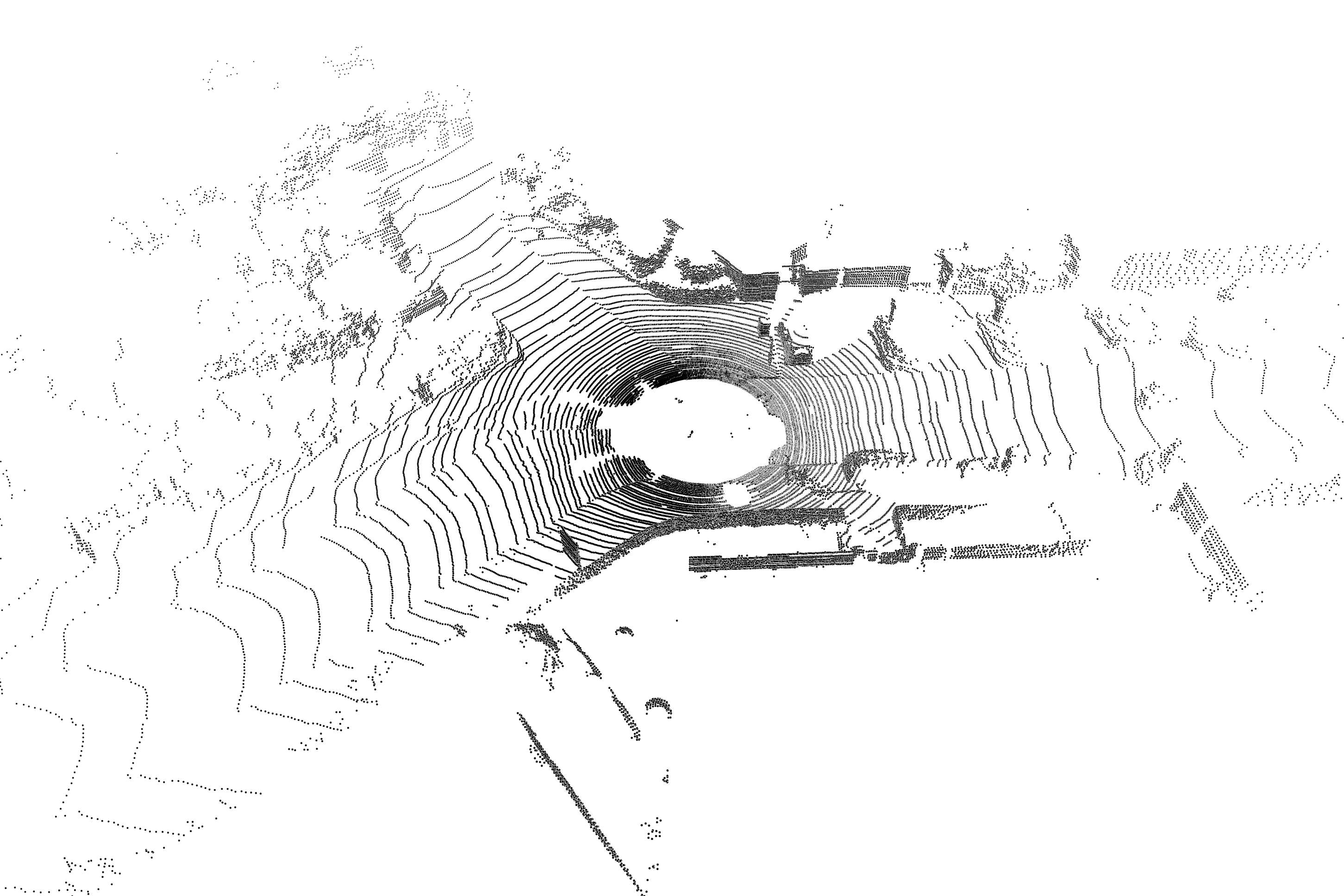}}
        \end{minipage} &

        \begin{minipage}[t]{0.142\hsize}
            \centering
            \subfloat[\ac{JPEG} \\
            bpp:1.46 \\
            CD:1.32]
            {\includegraphics[width=\hsize]{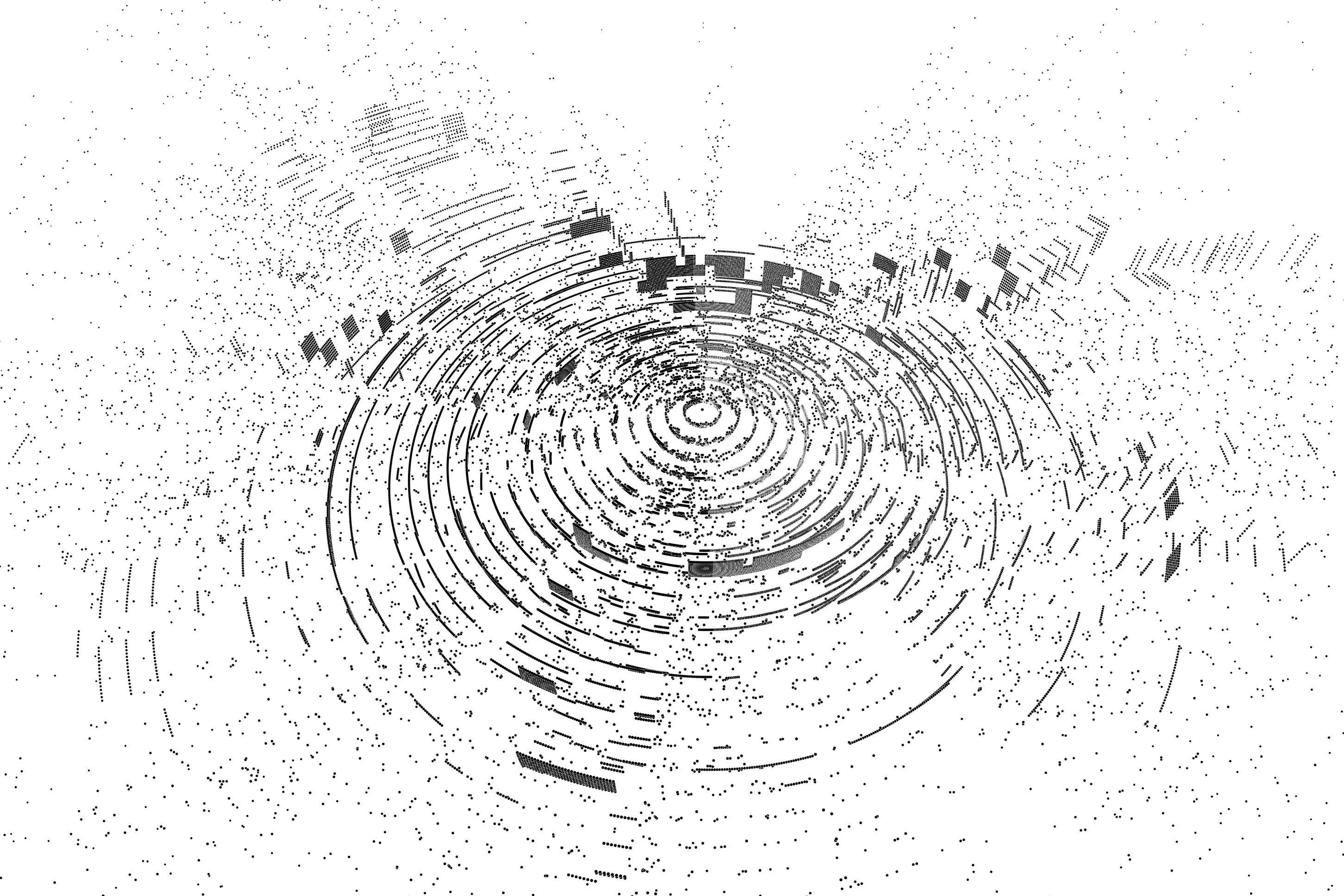}}
        \end{minipage} &

        \begin{minipage}[t]{0.142\hsize}
            \centering
            \subfloat[\ac{JPEG2000} \\
            bpp:1.33 \\
            CD:0.662]
            {\includegraphics[width=\hsize]{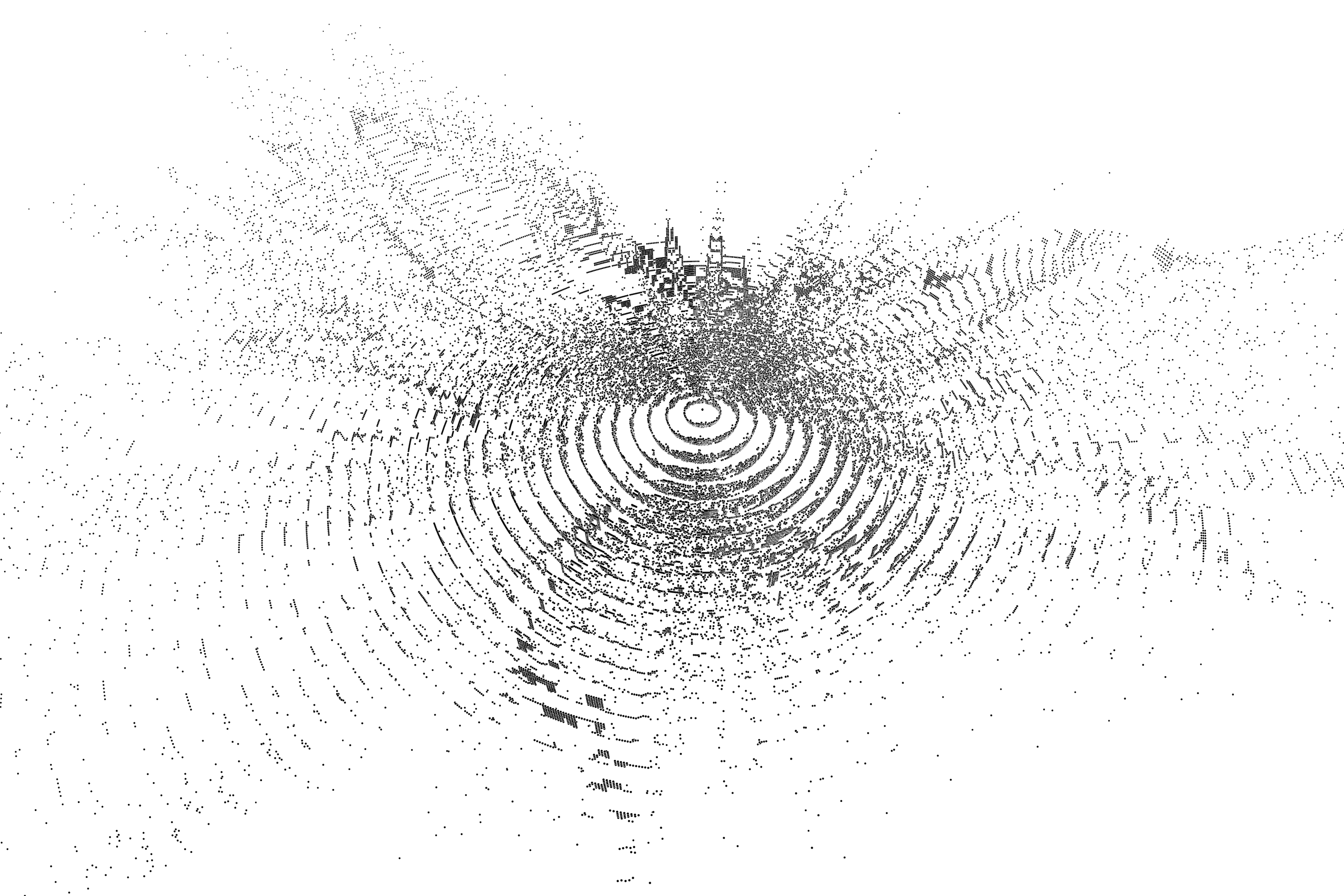}}
        \end{minipage} &

        \begin{minipage}[t]{0.142\hsize}
            \centering
            \subfloat[HEIF \\
            bpp:1.49 \\
            CD:0.355]
            {\includegraphics[width=\hsize]{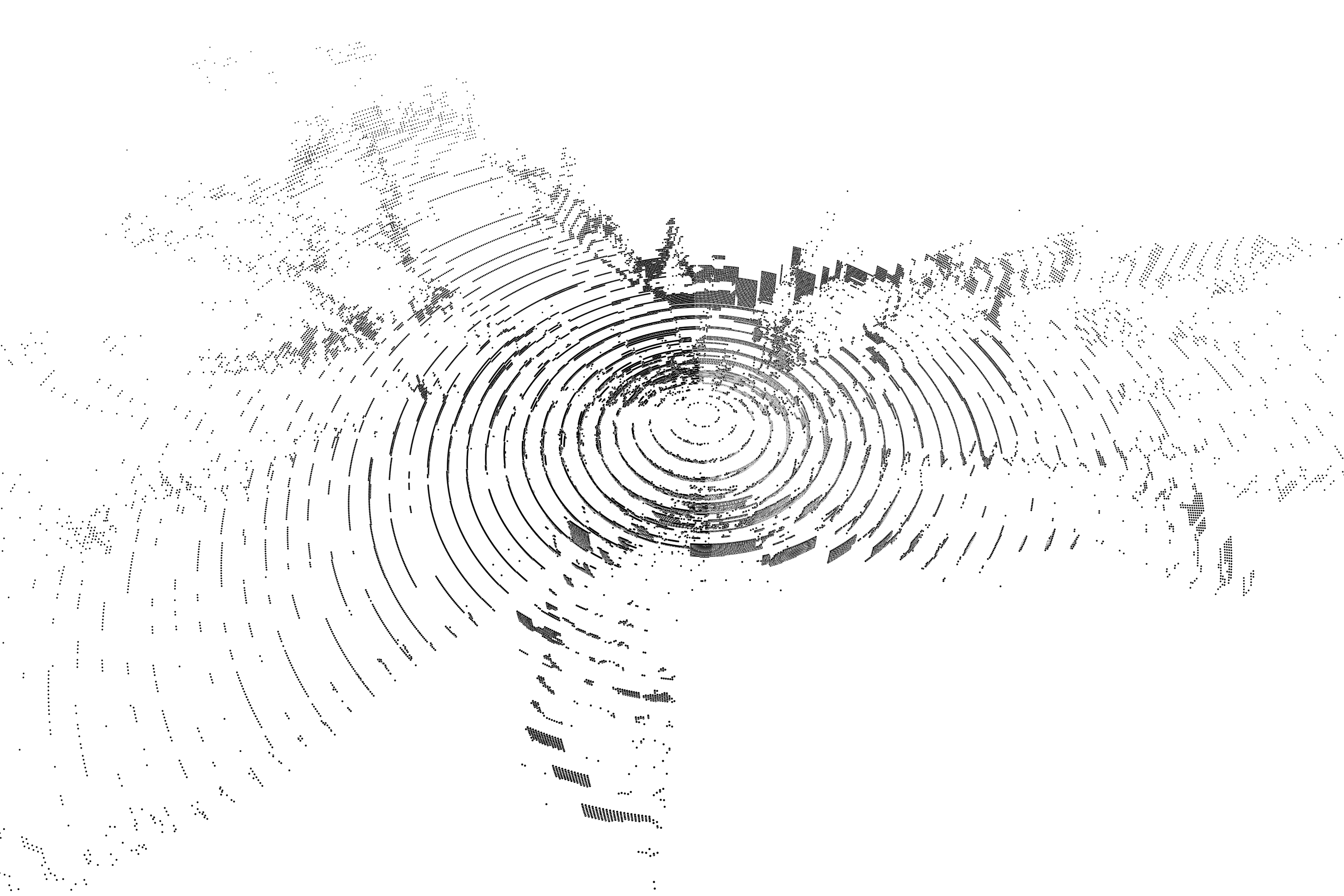}}
        \end{minipage} &

        \begin{minipage}[t]{0.142\hsize}
            \centering
            \subfloat[AVIF \\
            bpp:1.10 \\
            CD:0.324]
            {\includegraphics[width=\hsize]{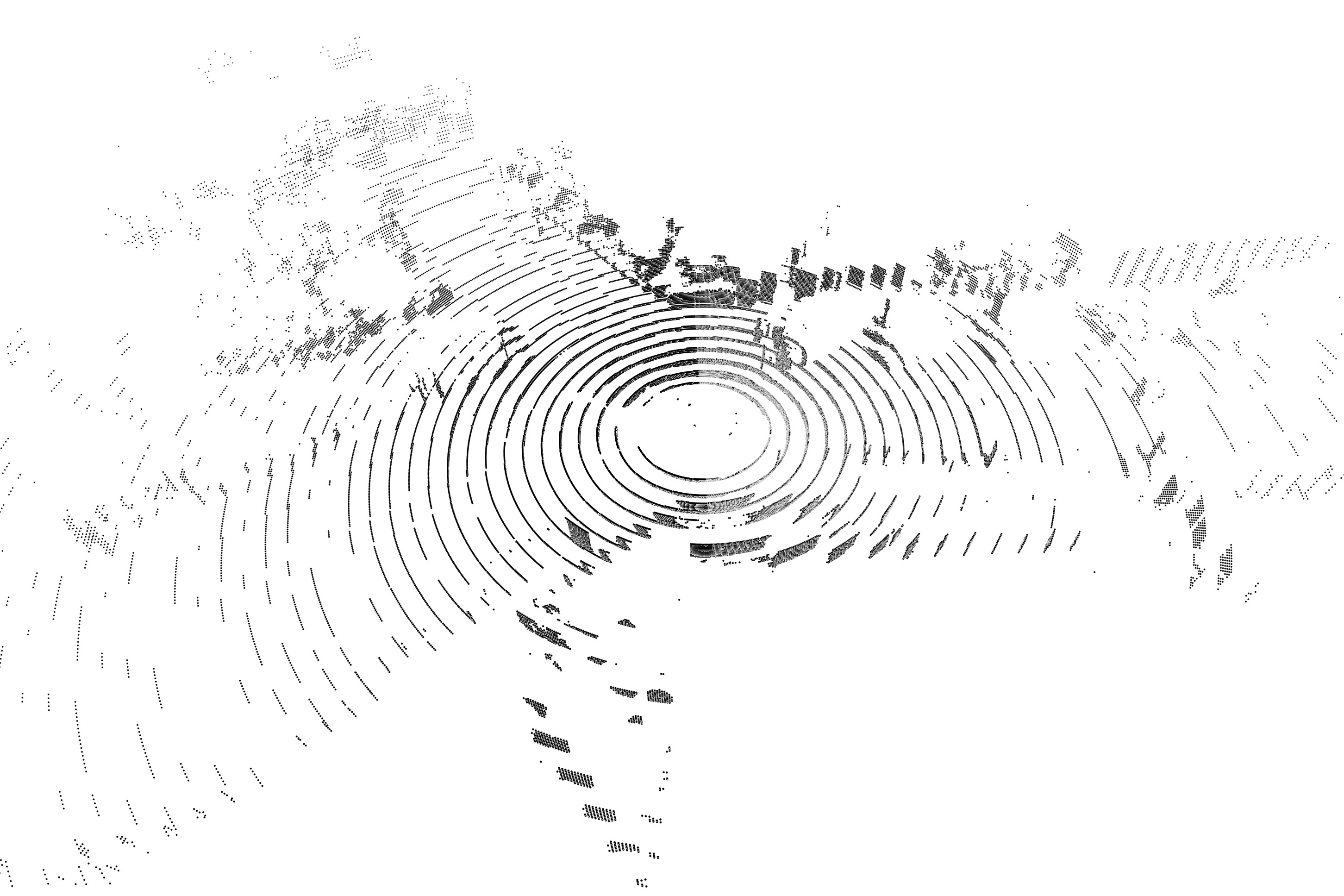}}
        \end{minipage} &

        \begin{minipage}[t]{0.142\hsize}
            \centering
            \subfloat[COIN \\
            bpp:1.33 \\
            CD:1.13]
            {\includegraphics[width=\hsize]{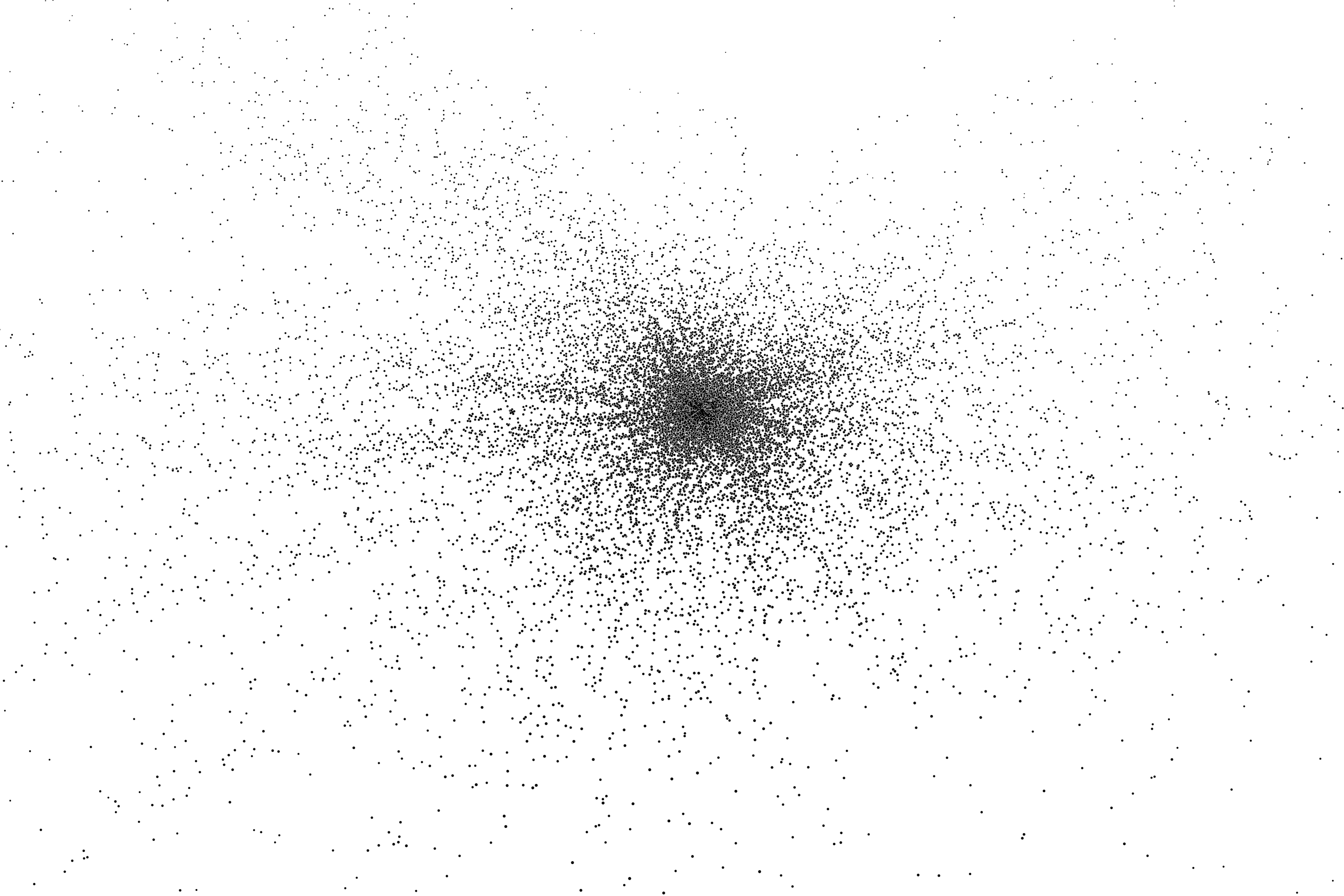}}
        \end{minipage} \\
        
        \begin{minipage}[t]{0.142\hsize}
            \centering
            \subfloat[\ac{G-PCC} \\
            bpp:1.46 \\
            CD:0.152]
            {\includegraphics[width=\hsize]{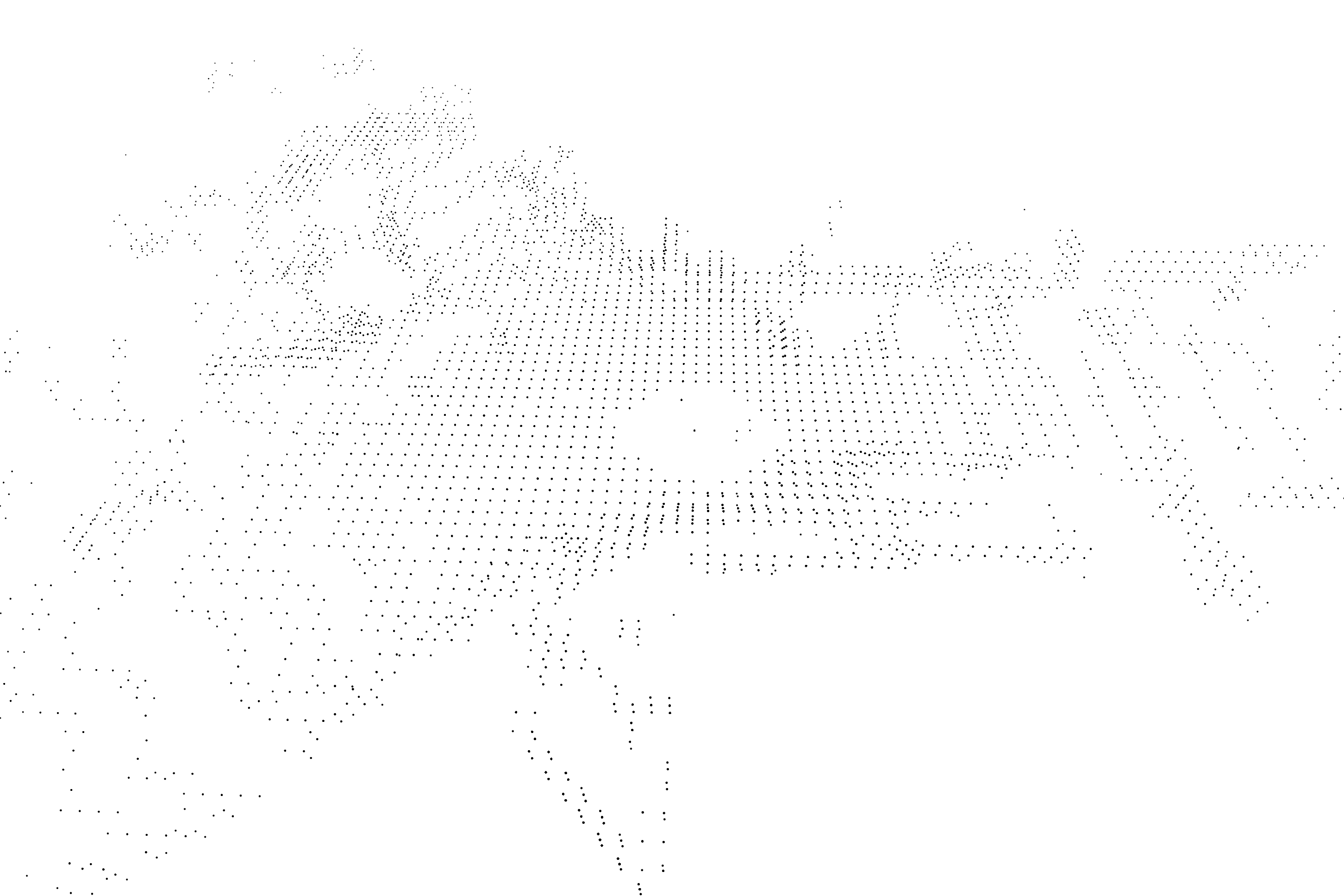}}
        \end{minipage} &

        \begin{minipage}[t]{0.142\hsize}
            \centering
            \subfloat[Draco \\
            bpp:1.10 \\
            CD:0.226]
            {\includegraphics[width=\hsize]{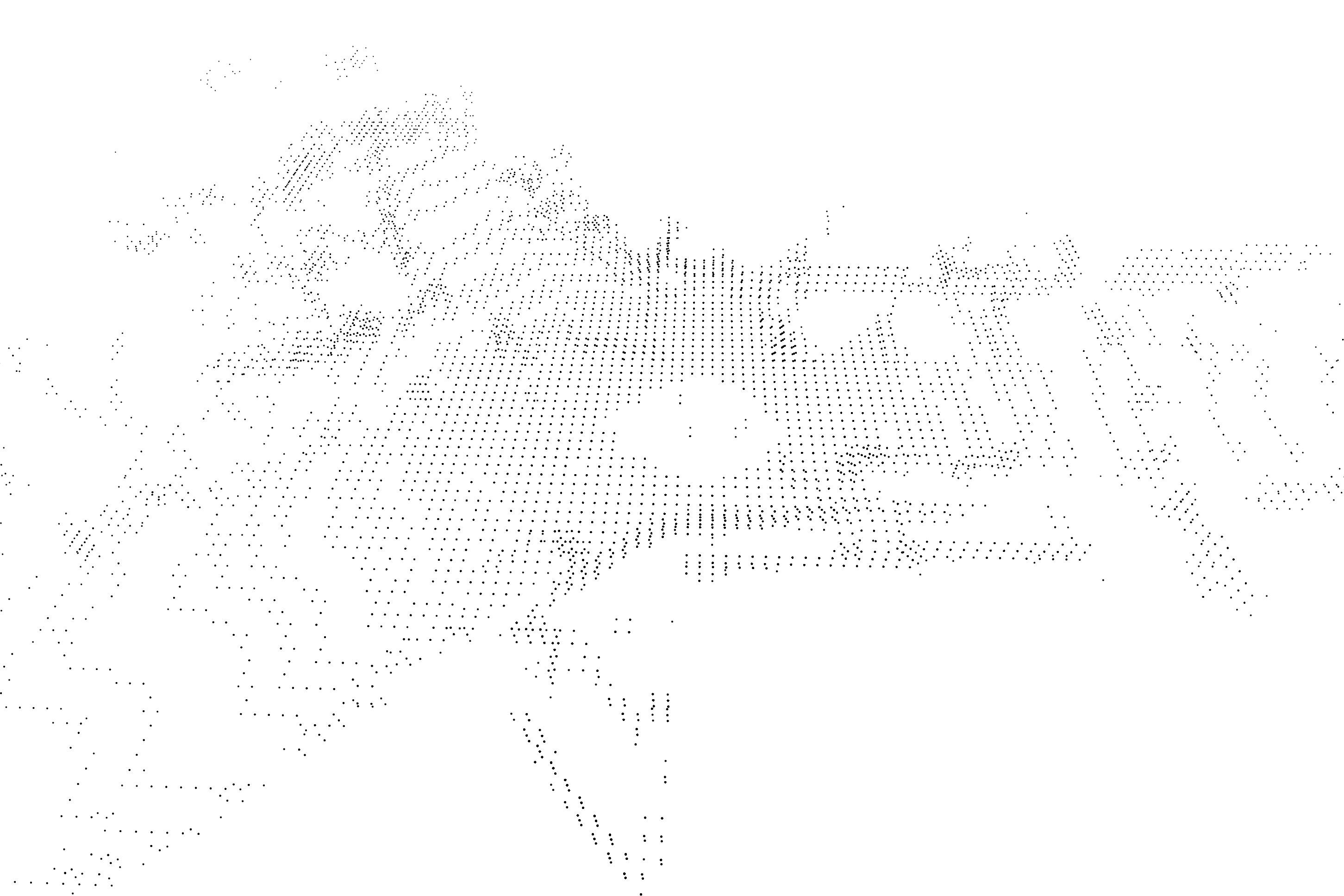}}
        \end{minipage} &

        \begin{minipage}[t]{0.142\hsize}
            \centering
            \subfloat[OctAttention \\
            bpp:1.36 \\
            CD:0.070]
            {\includegraphics[width=\hsize]{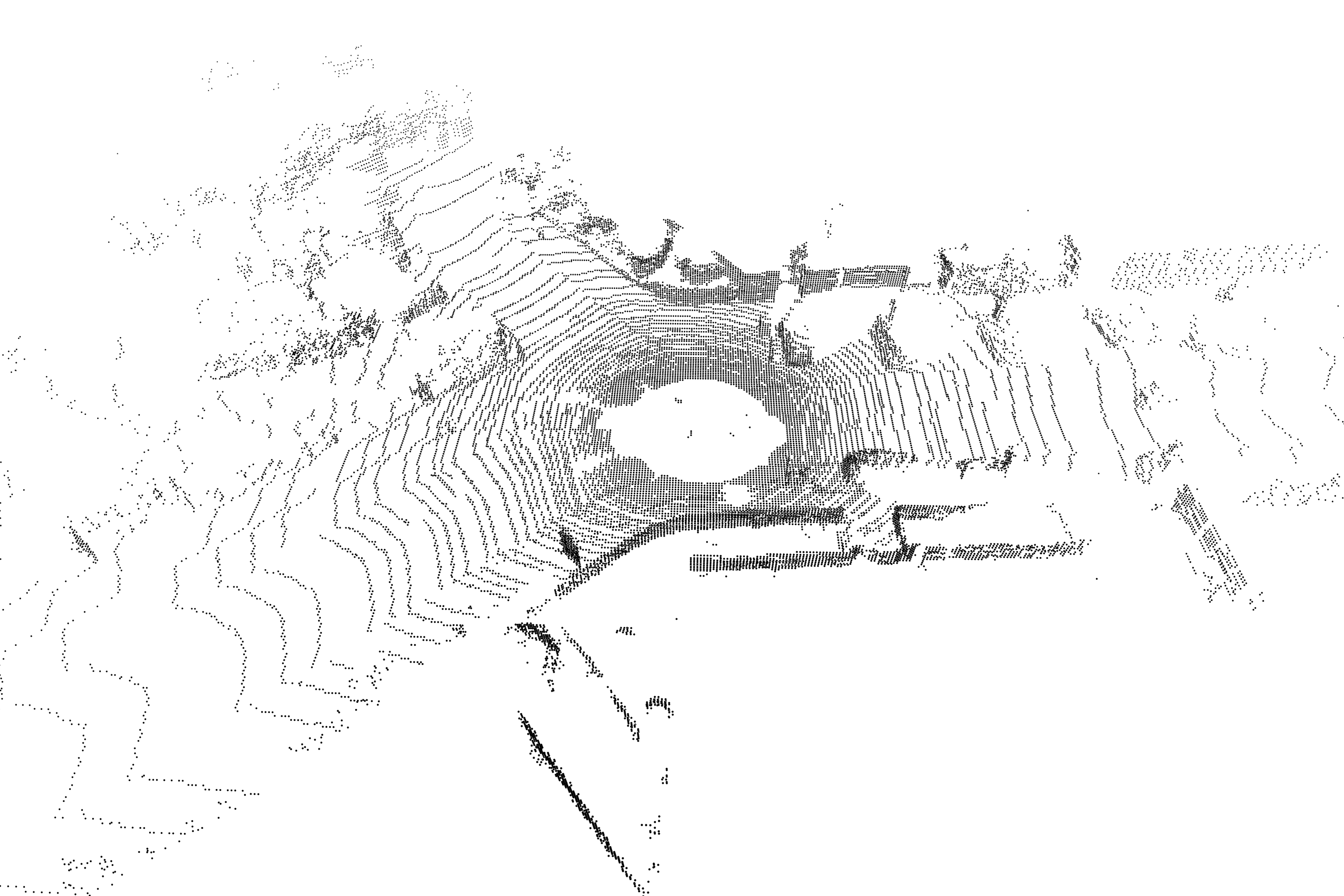}}
        \end{minipage} &

        \begin{minipage}[t]{0.142\hsize}
            \centering
            \subfloat[R-PCC~(Deflate) \\
            bpp:1.52 \\
            CD:0.147]
            {\includegraphics[width=\hsize]{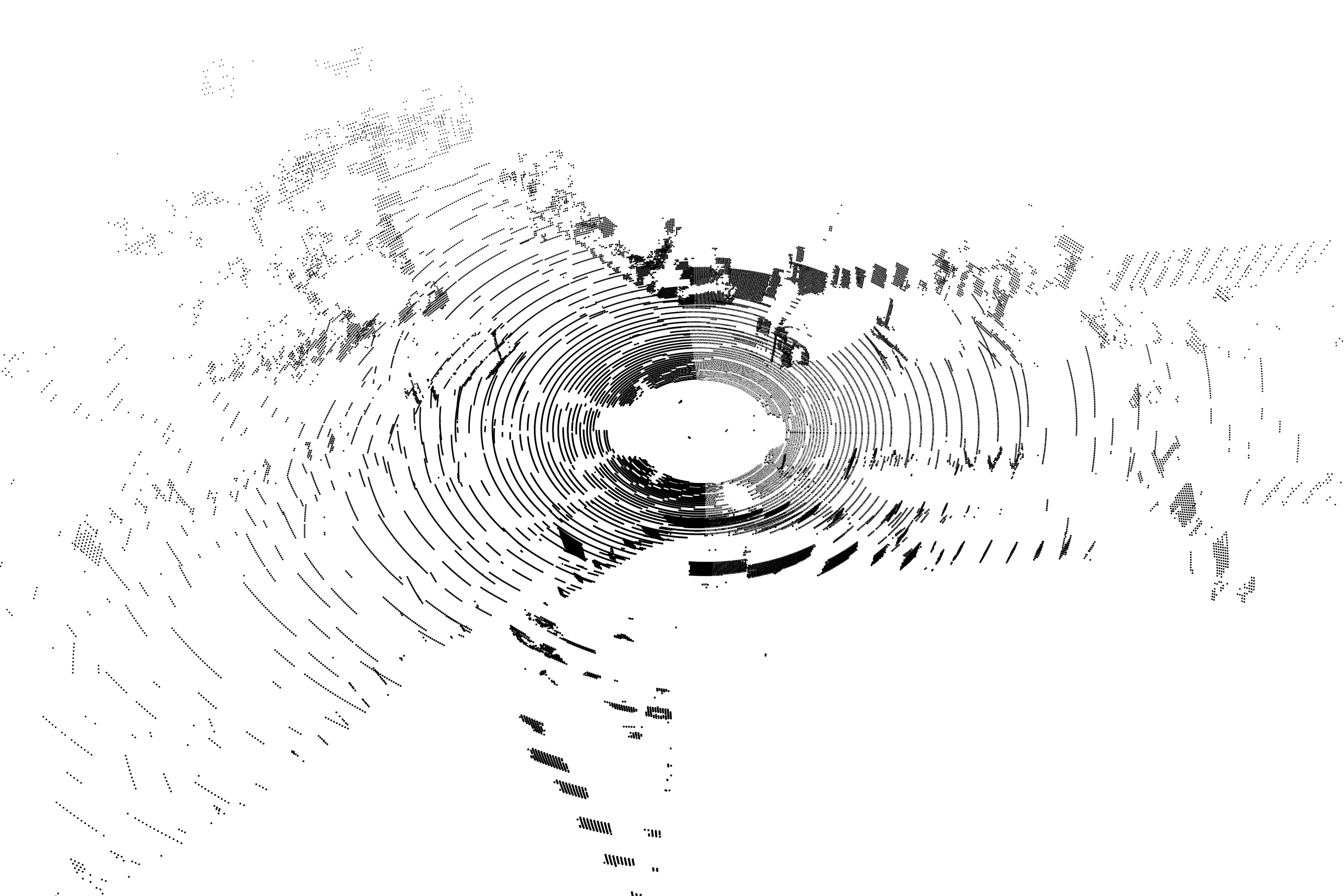}}
        \end{minipage} &

        \begin{minipage}[t]{0.142\hsize}
            \centering
            \subfloat[R-PCC~(LZ4) \\
            bpp:2.84 \\
            CD:0.229]
            {\includegraphics[width=\hsize]{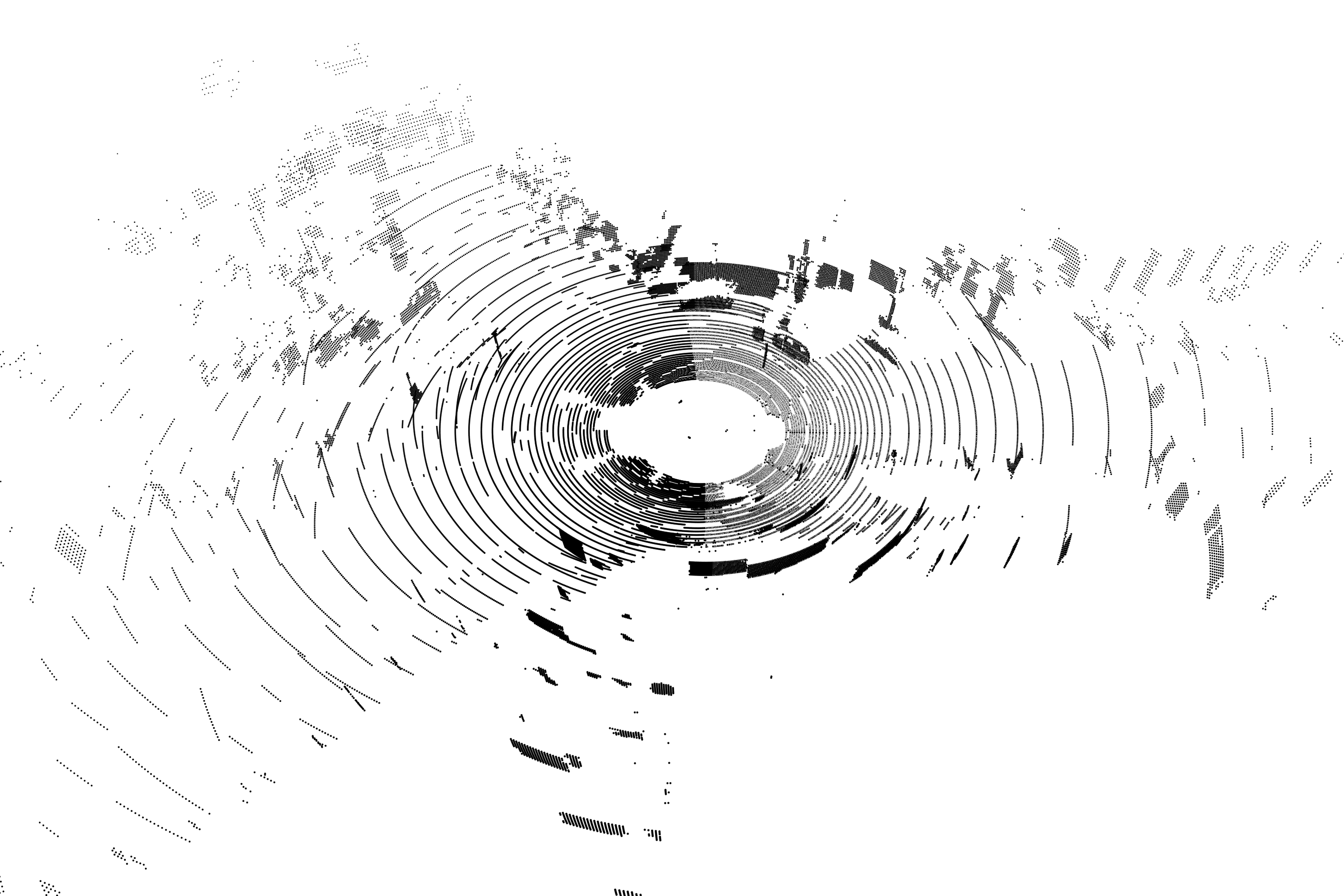}}
        \end{minipage} &

        \begin{minipage}[t]{0.142\hsize}
            \centering
            \subfloat[Proposed \\
            bpp:1.54 \\
            CD:0.077]
            {\includegraphics[width=\hsize]{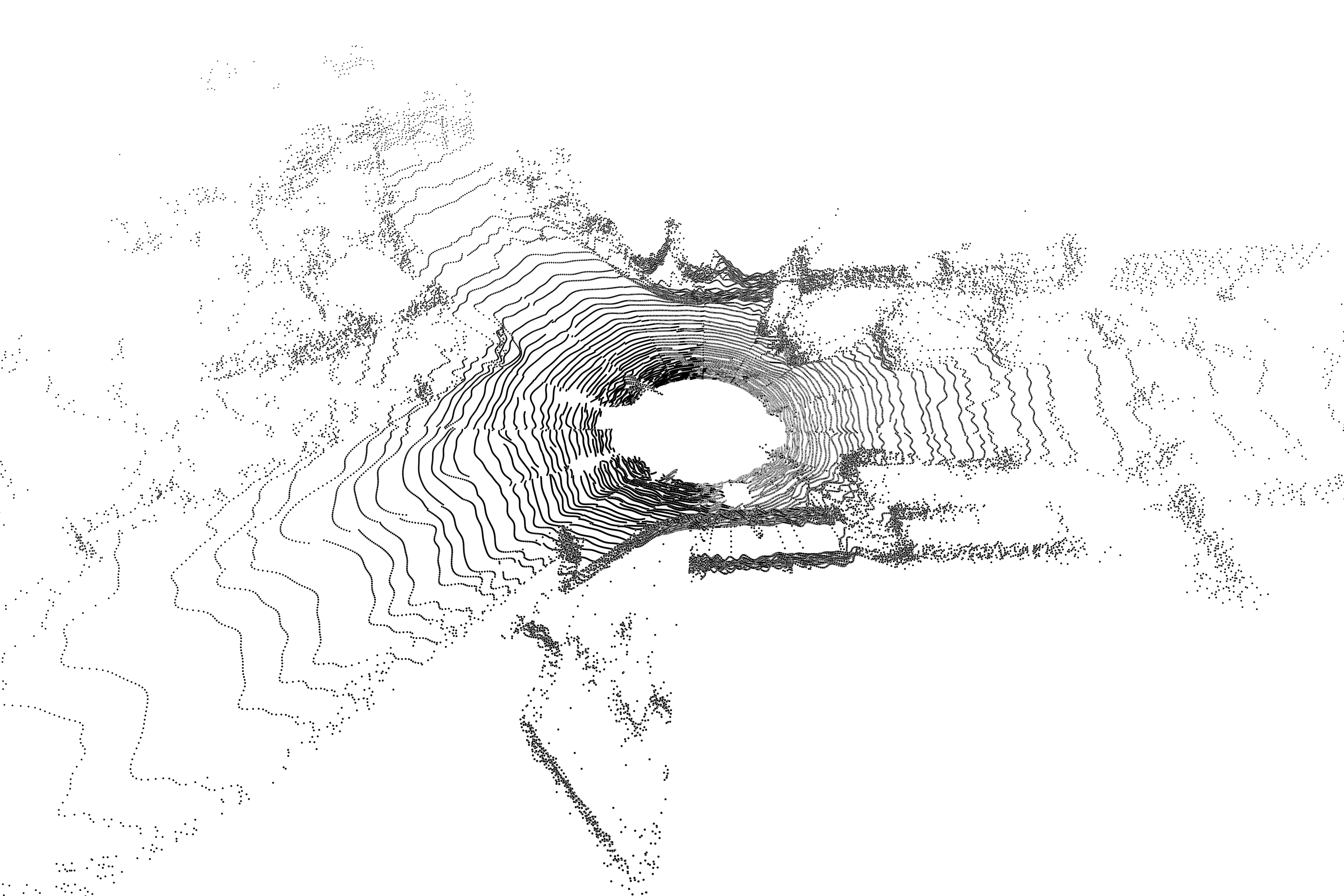}}
        \end{minipage} \\

        \begin{minipage}[t]{0.142\hsize}
            \centering
            \subfloat[{Original \\ Seq:~00 \\Frame:~25}]
            {\includegraphics[width=\hsize]{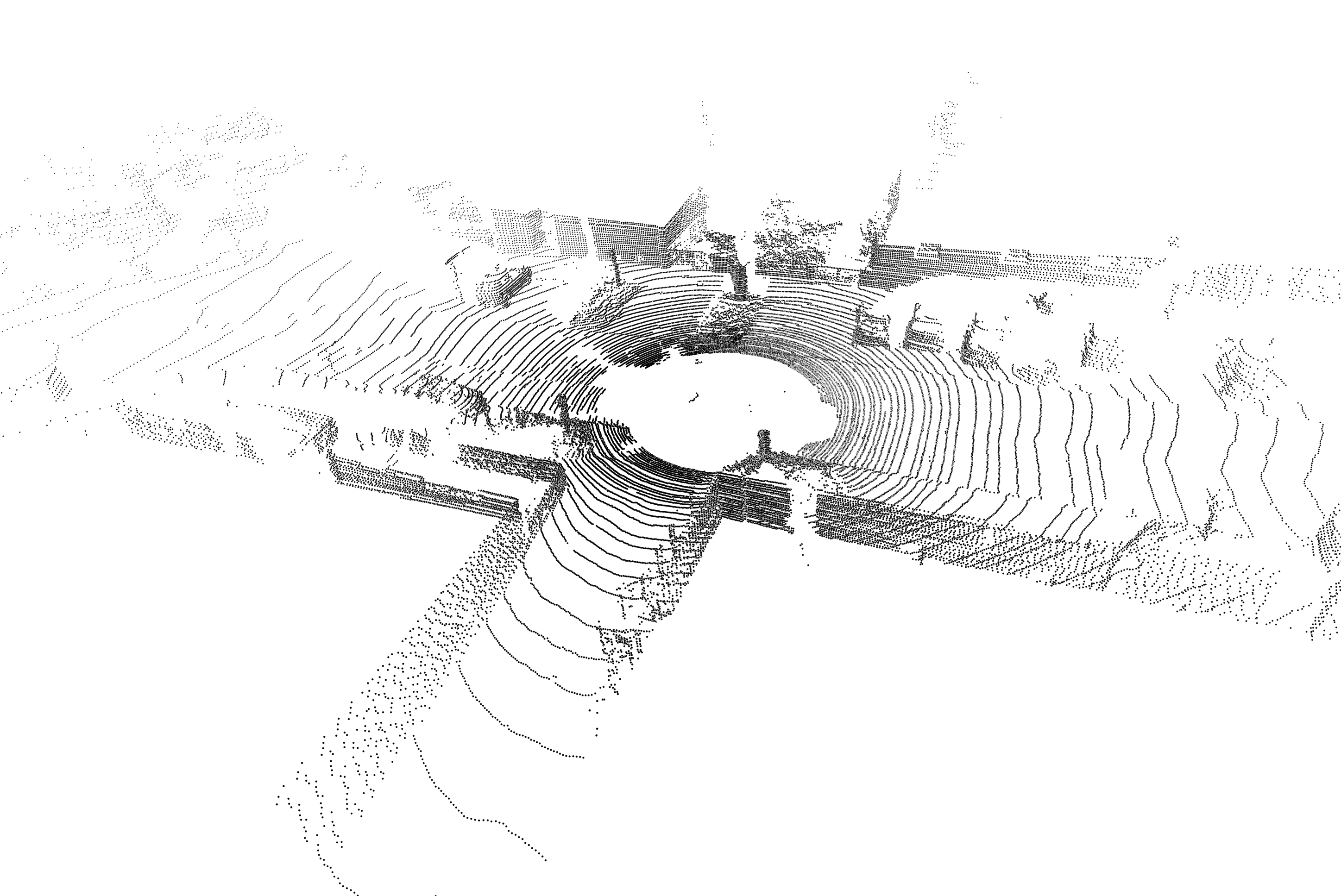}}
        \end{minipage} &

        \begin{minipage}[t]{0.142\hsize}
            \centering
            \subfloat[\ac{JPEG} \\
            bpp:1.93 \\
            CD:1.01]
            {\includegraphics[width=\hsize]{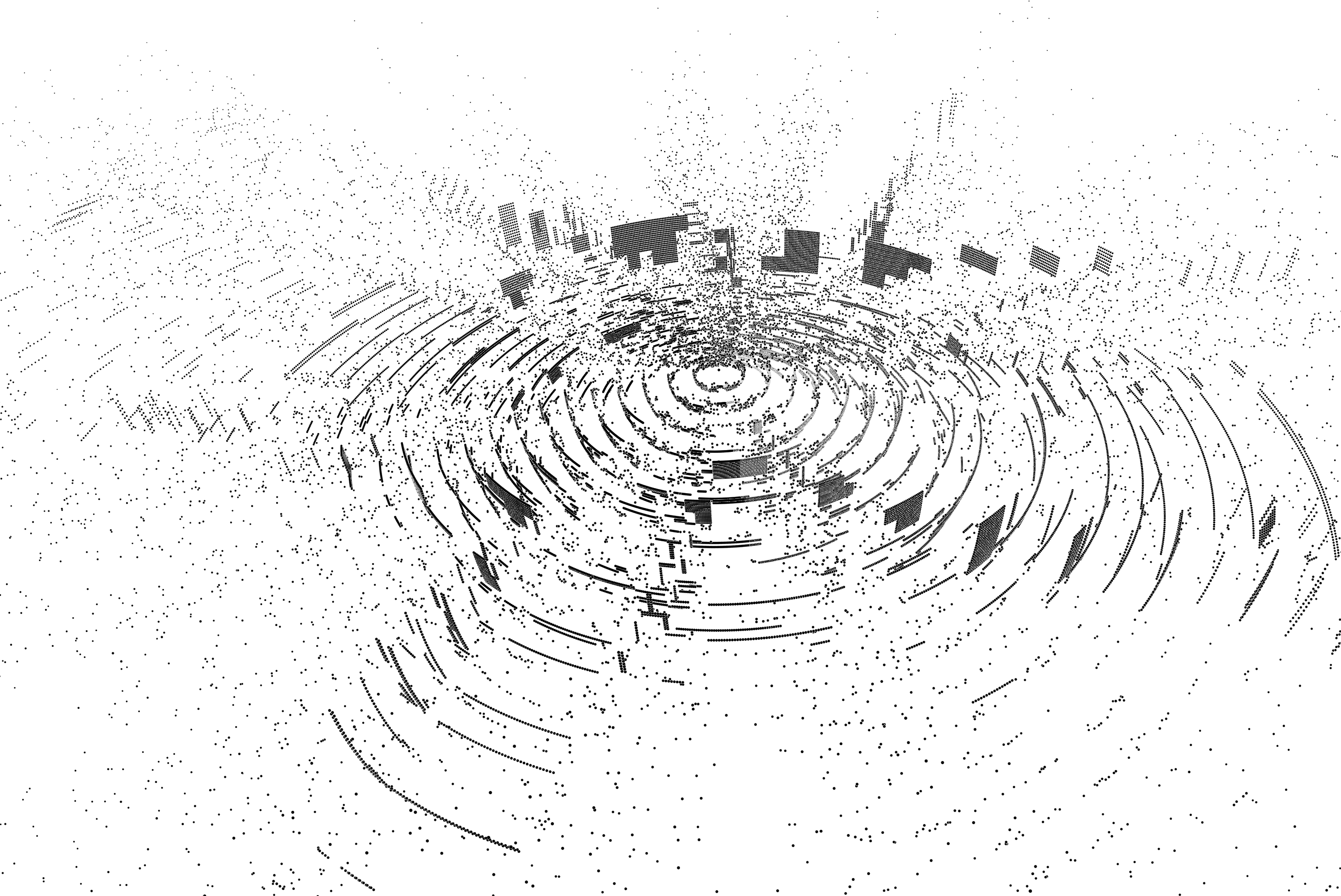}}
        \end{minipage} &

        \begin{minipage}[t]{0.142\hsize}
            \centering
            \subfloat[\ac{JPEG2000} \\
            bpp:1.99 \\
            CD:0.441]
            {\includegraphics[width=\hsize]{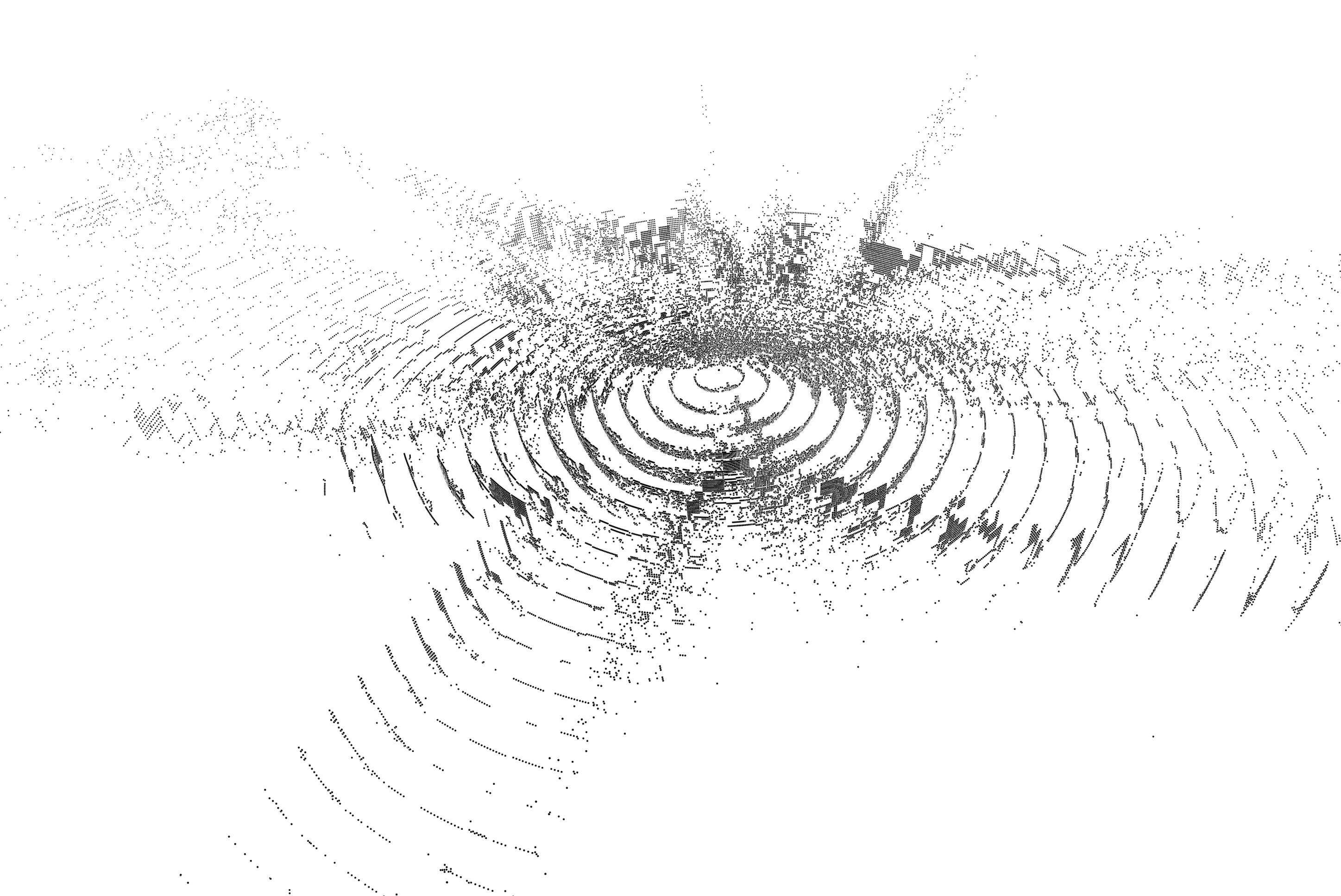}}
        \end{minipage} &

        \begin{minipage}[t]{0.142\hsize}
            \centering
            \subfloat[HEIF \\
            bpp:2.09 \\
            CD:0.323]
            {\includegraphics[width=\hsize]{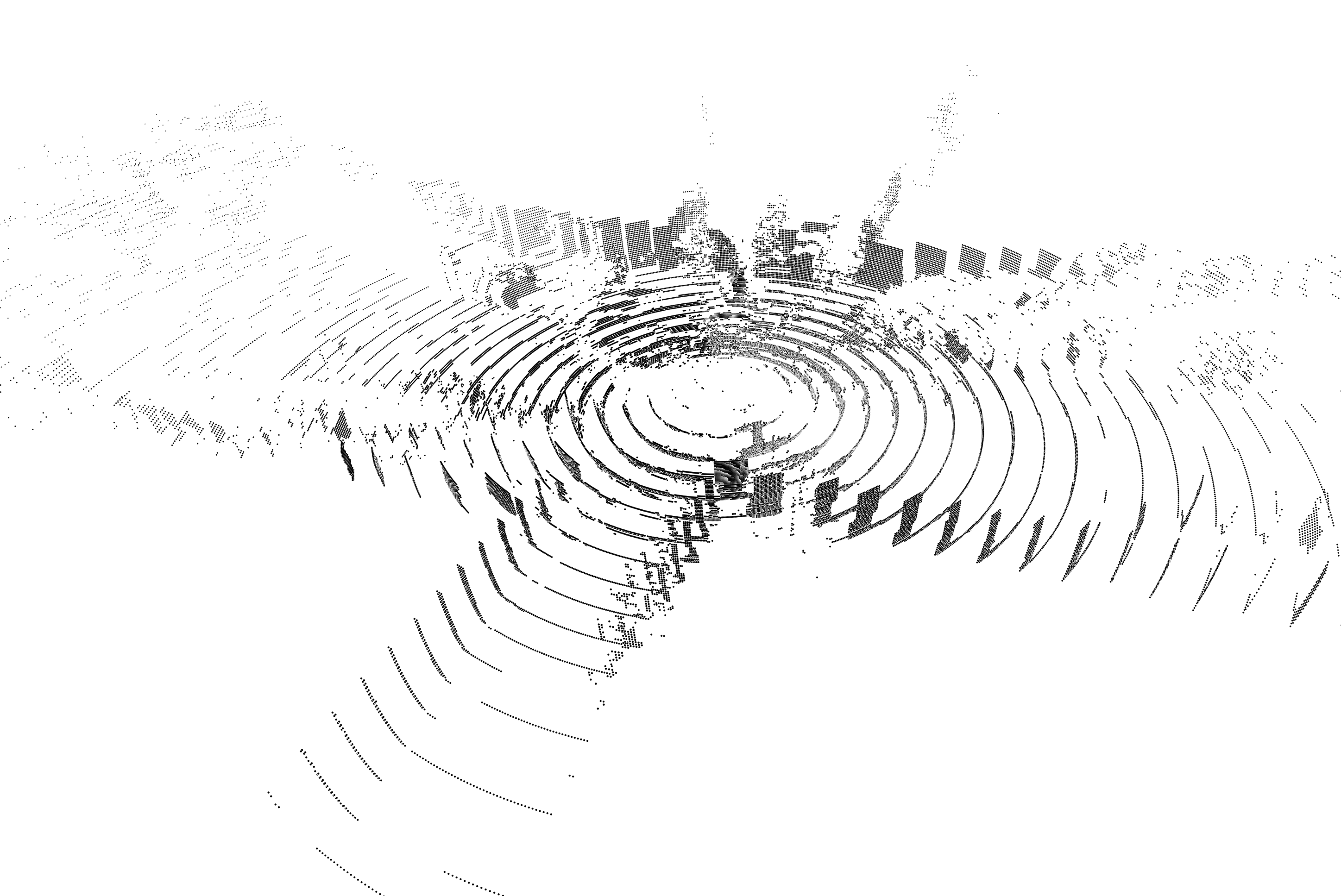}}
        \end{minipage} &

        \begin{minipage}[t]{0.142\hsize}
            \centering
            \subfloat[AVIF \\
            bpp:1.18 \\
            CD:0.308]
            {\includegraphics[width=\hsize]{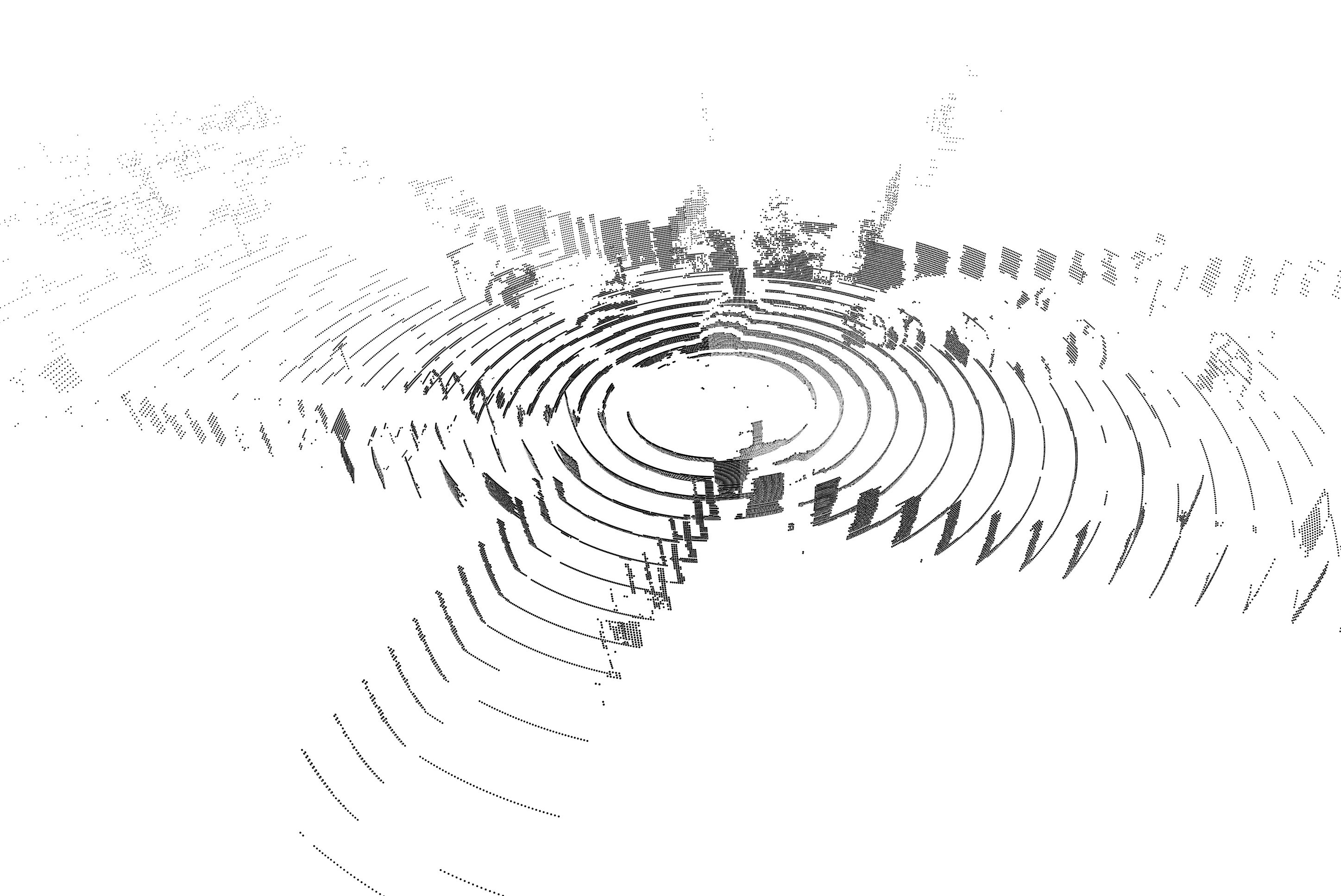}}
        \end{minipage} &

        \begin{minipage}[t]{0.142\hsize}
            \centering
            \subfloat[COIN \\
            bpp:1.33 \\
            CD:1.01]
            {\includegraphics[width=\hsize]{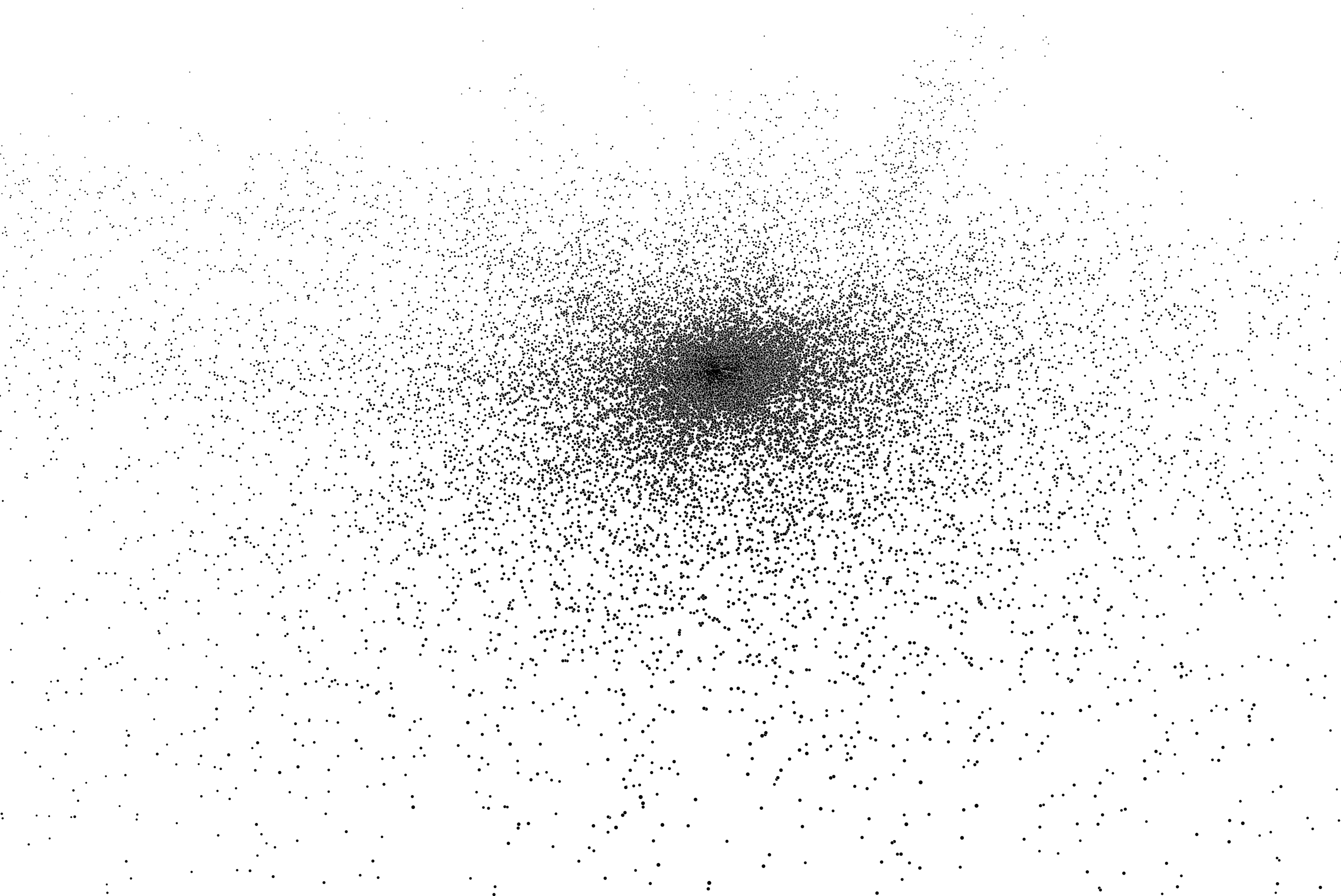}}
        \end{minipage} \\
        
        \begin{minipage}[t]{0.142\hsize}
            \centering
            \subfloat[\ac{G-PCC} \\
            bpp:1.89 \\
            CD:0.109]
            {\includegraphics[width=\hsize]{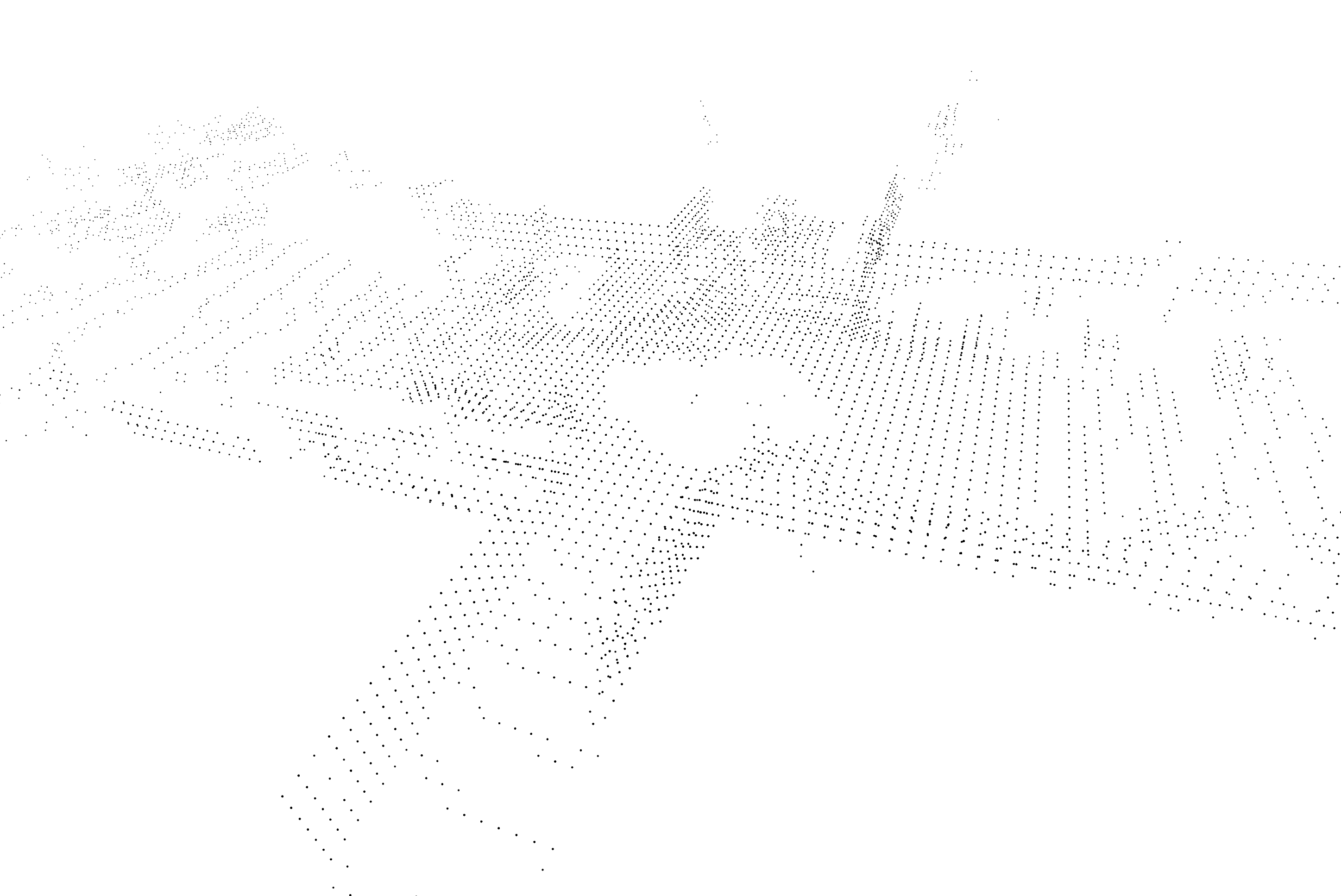}}
        \end{minipage} &

        \begin{minipage}[t]{0.142\hsize}
            \centering
            \subfloat[Draco \\
            bpp:1.98 \\
            CD:0.128]
            {\includegraphics[width=\hsize]{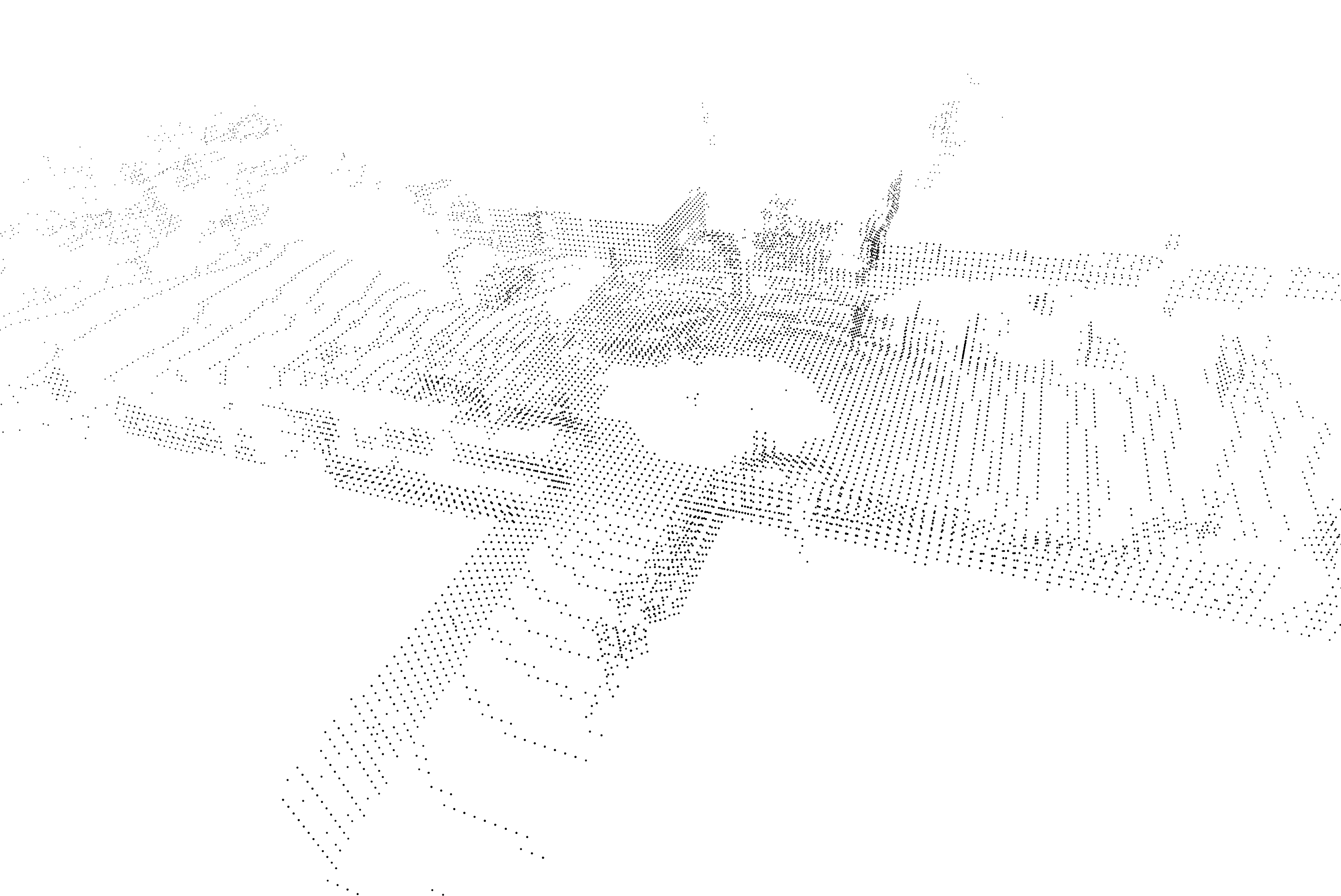}}
        \end{minipage} &

        \begin{minipage}[t]{0.142\hsize}
            \centering
            \subfloat[OctAttention \\
            bpp:2.44 \\
            CD:0.036]
            {\includegraphics[width=\hsize]{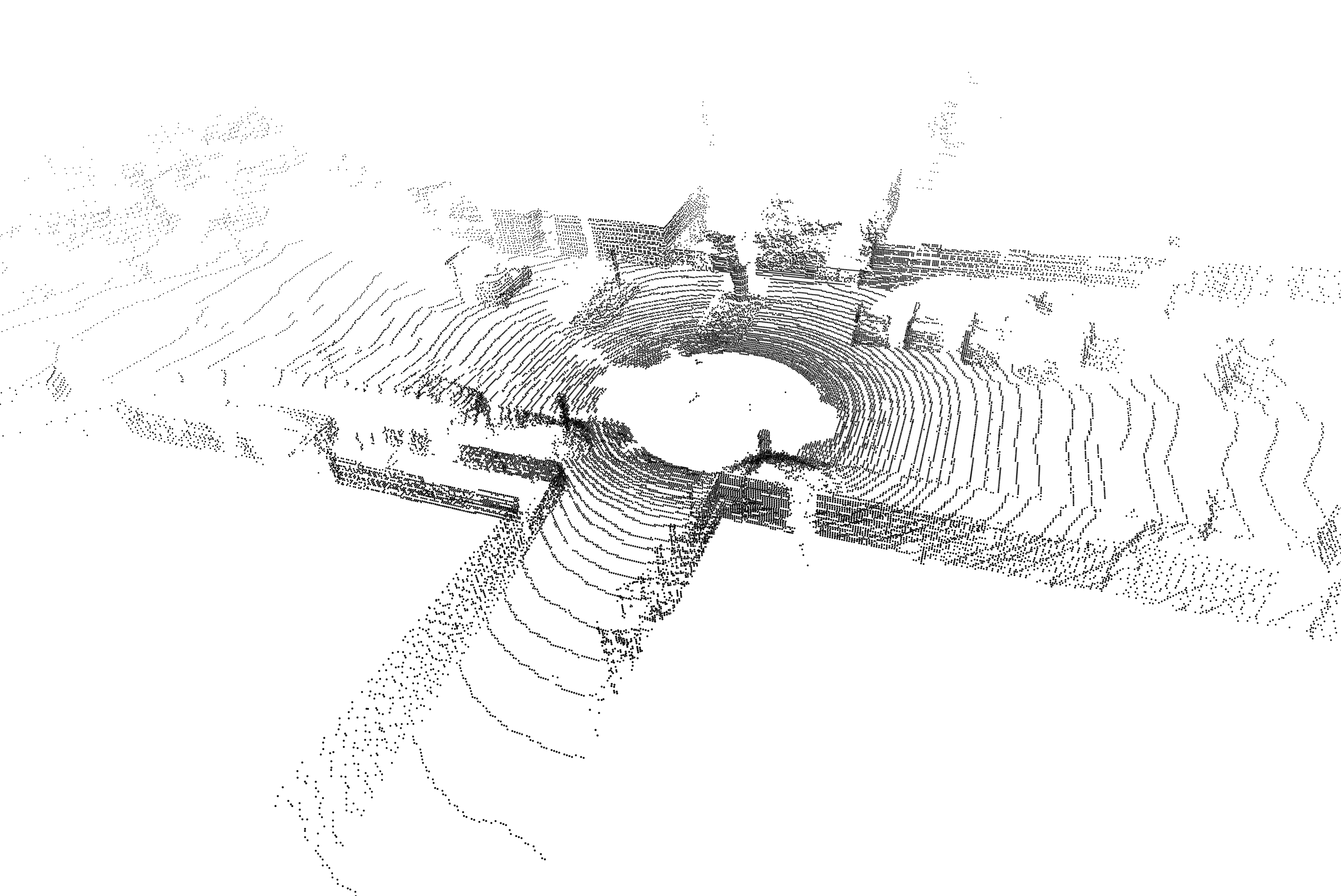}}
        \end{minipage} &

        \begin{minipage}[t]{0.142\hsize}
            \centering
            \subfloat[R-PCC~(Deflate) \\
            bpp:2.24 \\
            CD:0.082]
            {\includegraphics[width=\hsize]{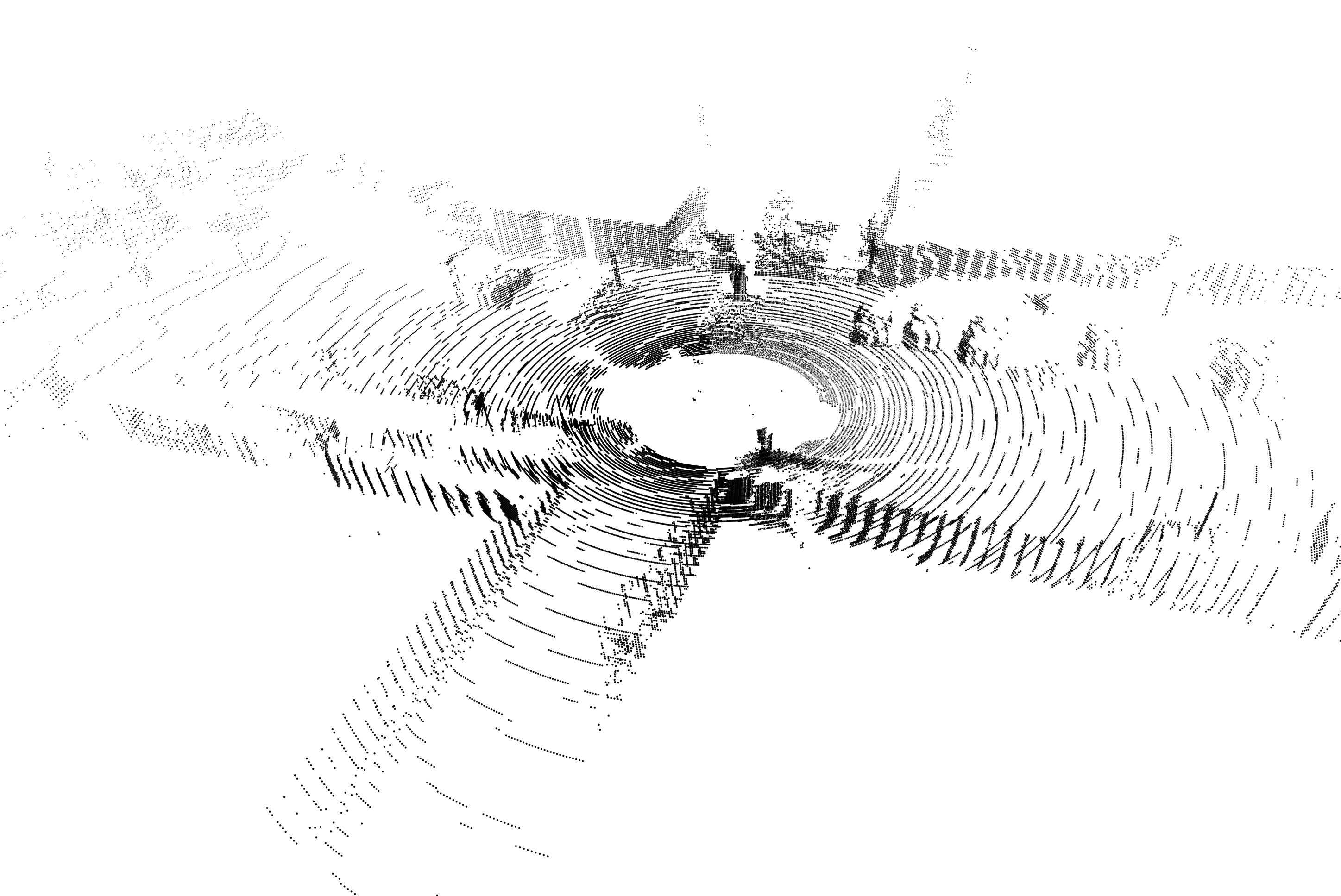}}
        \end{minipage} &

        \begin{minipage}[t]{0.142\hsize}
            \centering
            \subfloat[R-PCC~(LZ4) \\
            bpp:2.90 \\
            CD:0.235]
            {\includegraphics[width=\hsize]{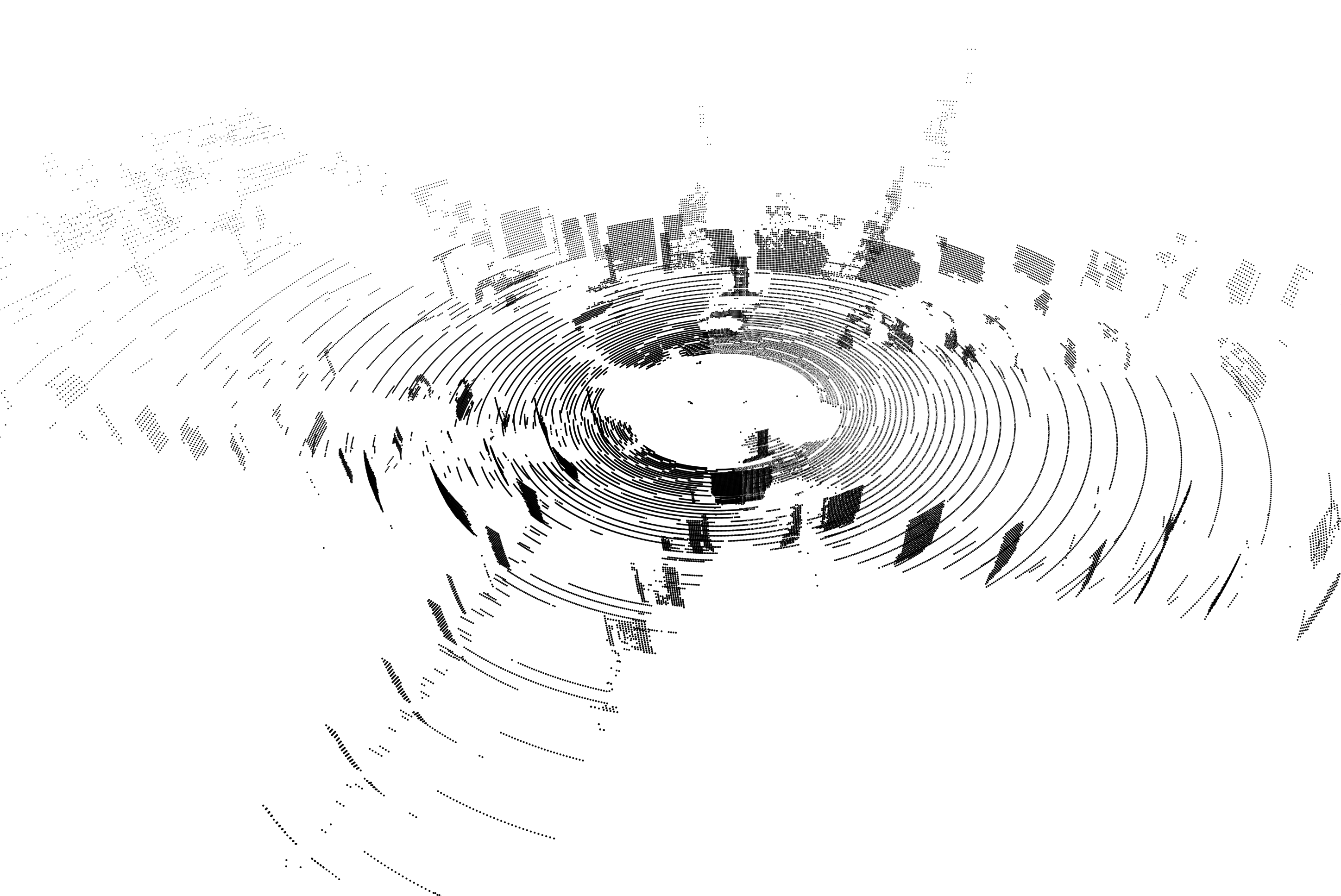}}
        \end{minipage} &

        \begin{minipage}[t]{0.142\hsize}
            \centering
            \subfloat[Proposed \\
            bpp:2.16 \\
            CD:0.058]
            {\includegraphics[width=\hsize]{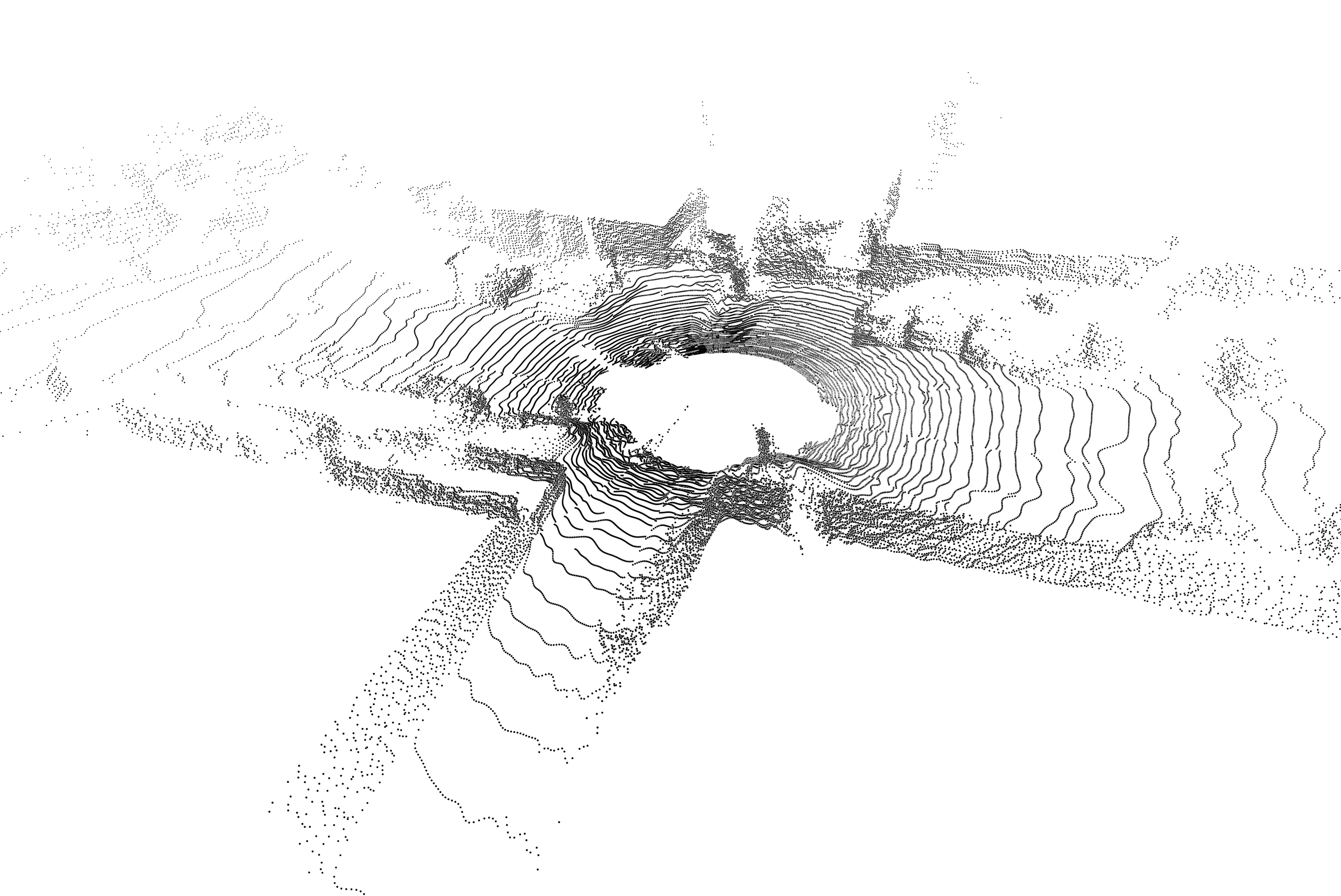}}
        \end{minipage} \\
    \end{tabular}
    \caption{Snapshot of the reconstructed {LiDAR} point clouds in proposed and baseline methods. Here, (b)-(l) and (n)-(x) show the reconstructed point clouds of {frames 00 and 25 of sequence 00, respectively.}}
    \label{fig:snaps}
\end{figure*}

\begin{table*}[t]
    \caption{The list of BD-CD $\uparrow$ for the KITTI dataset across the different sequences. {Note that BD-CD is evaluated for each baseline using the proposed method as the reference.
   Positive values indicate that the proposed method achieves a lower chamfer distance than the corresponding baseline.}}
    \centering
    \begin{threeparttable}
        \begin{tabular}{c c c c c c c c c c c} \toprule
            \textbf{Seq.} & \textbf{JPEG}$\dagger$ & \textbf{JPEG2000}$\dagger$ & \textbf{HEIF}$\dagger$ & \textbf{AVIF}$\dagger$ & \textbf{COIN}$\ddag$ & \textbf{G-PCC}$\S$ & \textbf{Draco}$\S$ & \textbf{\begin{tabular}{c}Oct \\ Attention\end{tabular}}$\S$ & \textbf{\begin{tabular}{c}R-PCC \\ (Deflate)\end{tabular}}$\P$ & \textbf{\begin{tabular}{c}R-PCC \\ (LZ4)\end{tabular}}$\P$ \\ \midrule
            00 & 1.393 & 0.801 & 0.535 & 0.256 & 1.103 & 0.081 & 0.157 & -0.025 & 0.023 & 0.110 \\
            01 & 1.234 & 0.686 & 0.515 & 0.259 & 1.095 & 0.097 & 0.234 & -0.017 & 0.019 & 0.097 \\
            02 & 1.120 & 0.612 & 0.447 & 0.254 & 1.039 & 0.083 & 0.226 & -0.026 & 0.006 & 0.094 \\
            03 & 1.322 & 0.733 & 0.497 & 0.232 & 1.074 & 0.059 & 0.176 & -0.045 & -0.007 & 0.078 \\
            04 & 1.343 & 0.748 & 0.515 & 0.237 & 1.092 & 0.074 & 0.192 & -0.039 & 0.010 & 0.097 \\
            05 & 1.484 & 0.877 & 0.576 & 0.248 & 1.041 & 0.065 & 0.153 & -0.040 & 0.008 & 0.095 \\
            06 & 1.324 & 0.785 & 0.527 & 0.224 & 0.993 & 0.053 & 0.153 & -0.070 & -0.024 & 0.062 \\
            \midrule
            Average & 1.317 & 0.749 & 0.516 & 0.244 & 1.063 & 0.073 & 0.185 & -0.037 & 0.005 & 0.090 \\ \bottomrule
        \end{tabular}
        \begin{tablenotes}
            \item[]$\dagger$: Image based method(s), \quad $\ddag$: INR based method, \quad $\S$: Point Cloud based method(s), \quad $\P$: Range Image based method(s).
        \end{tablenotes}
    \end{threeparttable}
    \label{tab:bd-cd}
\end{table*}

\begin{figure*}[t]
  \centering
  \subfloat[PointPillars]{
    \includegraphics[width=.33\linewidth]{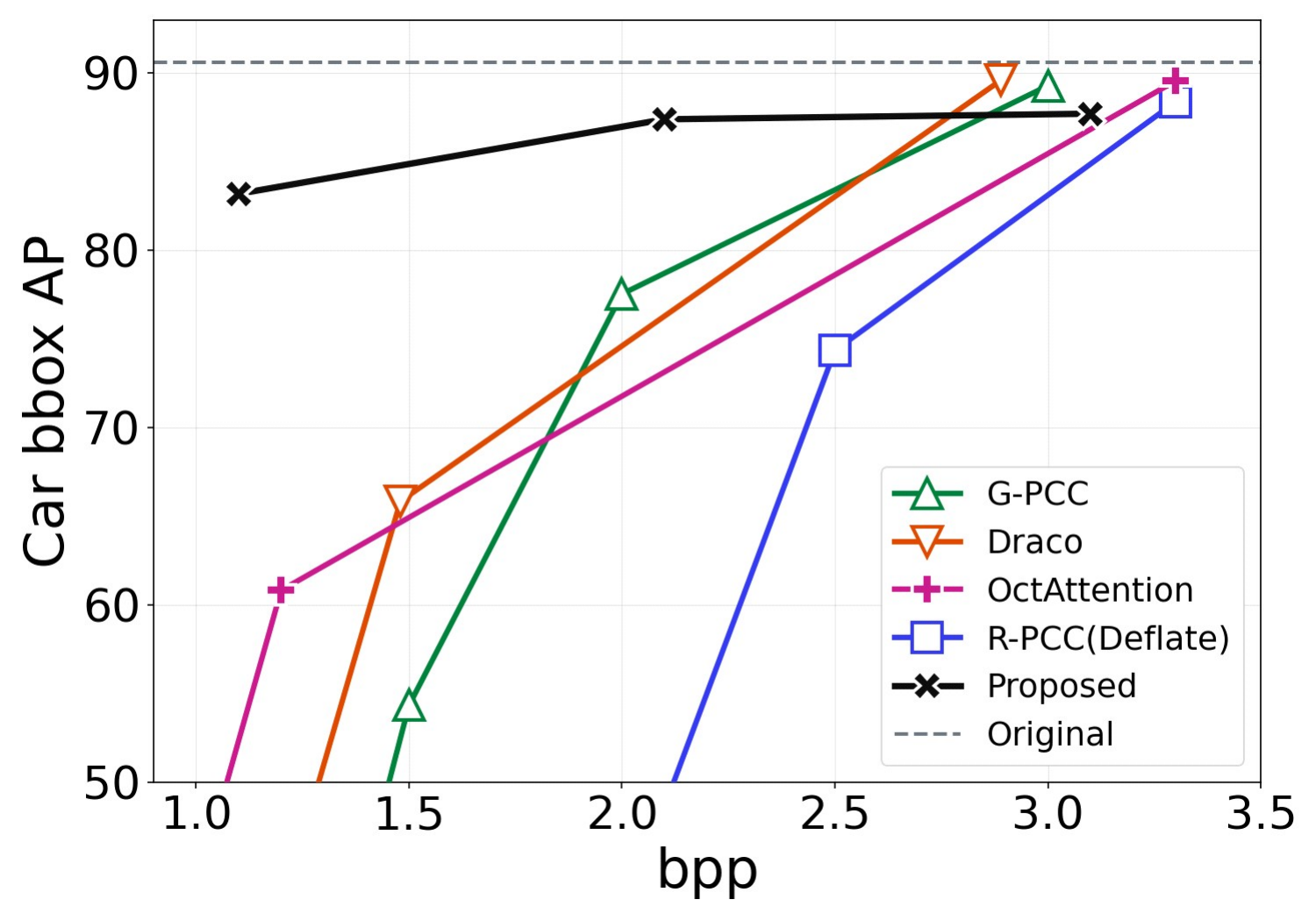}
  }
  \subfloat[SECOND]{
    \includegraphics[width=.33\linewidth]{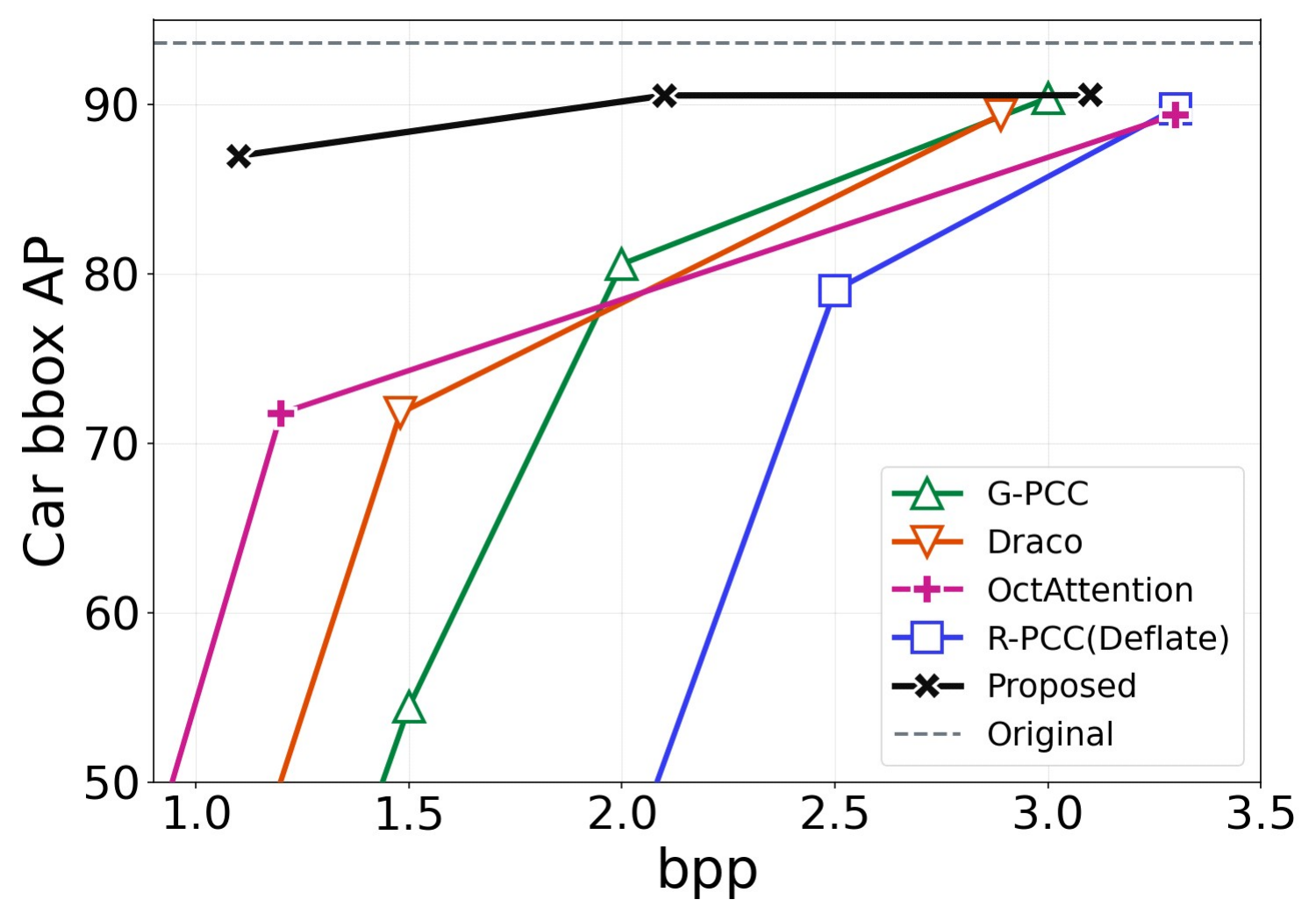}
  }
  \subfloat[PointRCNN]{
    \includegraphics[width=.33\linewidth]{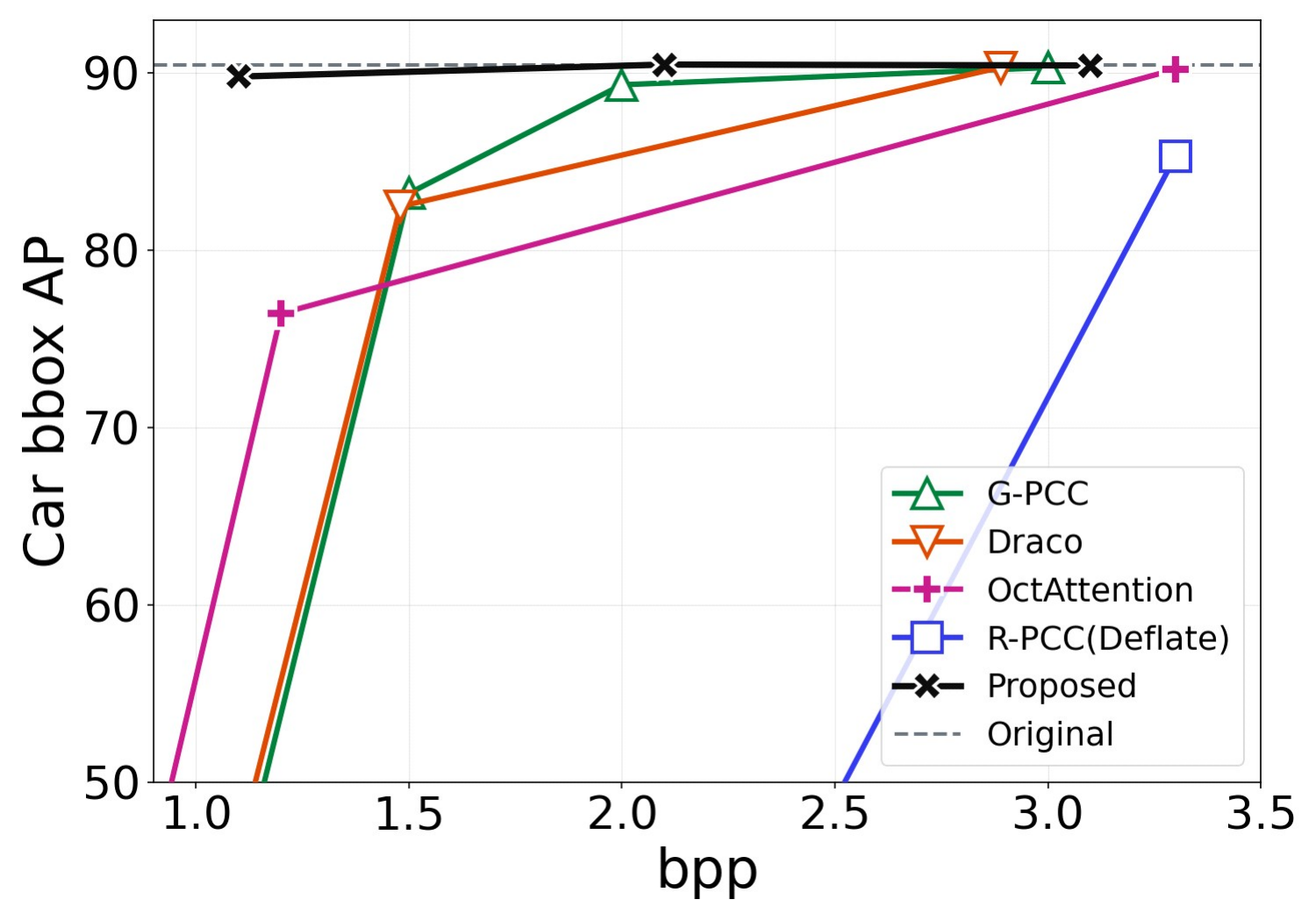}
  }
\caption{{Quantitative results of 3D object detection on the KITTI detection dataset. Car 3D bounding-box AP@0.7, 0.7, 0.7 as a function of bitrate (bits per point), for three downstream detectors (left to right): PointPillars, SECOND, and PointRCNN.}}

  \label{fig:downstreaming-task}
\end{figure*}

\begin{table}[t]
    \centering
    \normalsize
    \caption{Average decoding latency $\downarrow$}
    \label{tab:decoding-time}
    \begin{tabular}{l|c} \toprule
        \textbf{Method} & \textbf{Latency per frame} \\ \midrule
        JPEG           & 0.49~ms                                    \\ 
        JPEG2000       & 0.54~ms                                    \\ 
        HEIF           & 0.51~ms                                    \\ 
        AVIF           & 0.48~ms                                    \\ 
        COIN           & 0.71~ms                                    \\
        G-PCC          & 2.00~ms                                    \\ 
        Draco          & 3.00~ms                                    \\ 
        OctAttention   & 10.6~s                                     \\             
        R-PCC~(Deflate) & 11.5~ms                                    \\
        R-PCC~(LZ4)     & 10.3~ms                                    \\ 
        Proposed       & 0.69~ms                                    \\ \bottomrule
    \end{tabular}
\end{table}

\begin{table}[t]
    \centering
    \normalsize
    \caption{Average encoding latency $\downarrow$}
    \begin{tabular}{l|c} \toprule
            \textbf{Method} & \textbf{Latency per frame} \\ \midrule
        JPEG       &    9~ms \\
        JPEG2000       &    10~ms                                 \\ 
        HEIF           &    55~ms                              \\ 
        AVIF           &    94~ms                                 \\ 
        COIN           & 30~min                                    \\
        G-PCC          &   65~ms                                \\ 
        Draco          &    10~ms                                 \\ 
        OctAttention         &    134~ms                                 \\ 
        R-PCC~(Deflate) &  20~ms                                   \\
        R-PCC~(LZ4)     & 60~ms                                    \\ 
        Proposed       & 180~min                                    \\ \bottomrule
    \end{tabular}
    \label{tab:encoding-time} 
\end{table}


\subsection{Comparison with Baselines}

\subsubsection{RATE-DISTORTION PERFORMANCE}\label{rd-performance}

We show the \ac{R-D} performance of our proposed method and baselines. 
Figs.~\ref{fig:rate-distortion} show the \ac{CD} between the original and reconstructed {LiDAR} point clouds against various bitrates, i.e., bit per point~(bpp).
We observe the following findings:
\begin{itemize}
    \item The proposed method achieves higher 3D reconstruction quality than G-PCC, Draco, image compression, and RI compression methods across the bpp range up to 3.0 for frame 00.
    \item  {OctAttention achieves the best \ac{R-D} performance in frames 00, 25 and 50, whereas it requires long decoding latency, as will be detailed in Table~\ref{tab:decoding-time}.}
    \item Image compression methods suffer from quality saturation due to the precision disparity between \ac{RI} and the typical 8-bit precision image.
    \item \ac{PCC} methods do not have saturation since they compress the geometry information with 10-bit precision. 
    \item In R-PCC, \ac{R-D} performance highly depends on the lossless coding method.
    \item The \ac{INR}-based method requires a large model size for reconstructing high-quality \ac{RI}.
\end{itemize}

Figs.~\ref{fig:snaps}~(a)-(x) show the snapshots of the original and reconstructed {LiDAR} point clouds in each method 
Here, Figs.~\ref{fig:snaps}~(a)-(l) and (m)-(x) use {frames 00 and 25 of sequence 00, respectively.}
The proposed method can reconstruct a clean point cloud at the same bitrate. 
However, some \ac{PCC}, image compression, and \ac{INR}-based compression methods contain circular noises and/or decrease the number of 3D points in the reconstructed {LiDAR} point clouds. 
A circular noise still remains in R-PCC methods as well.

Table~\ref{tab:bd-cd} lists the average \ac{BD-CD} performance of the proposed method against the baselines in each sequence of {LiDAR} point clouds. {Here, BD-CD is evaluated for each baseline using the proposed method as the reference.
It shows that the proposed method achieves the best 3D reconstruction quality in the same bitrate range against G-PCC, Draco, image compression, and INR-based compression methods irrespective of LiDAR sequences. 
For OctAttention and R-PCC~(Deflate), the proposed method can be comparable or slightly worse in some cases.}
In addition, the performance gap between R-PCC and the proposed methods depends on the lossless coding method. 
When R-PCC uses a low-efficiency coding method, such as LZ4, for fast decoding, the proposed scheme achieves better R-D performance than R-PCC.

\subsubsection{DOWNSTREAM TASK}\label{downstream}
We then discuss the impact of our RI compression method on the performance of downstream tasks on the LiDAR point cloud.
We selected 3D object detection as a representative example of downstream tasks.
{
To evaluate robustness across different perception architectures, we consider three representative LiDAR detectors: PointPillars (BEV-based), SECOND (voxel-based), and PointRCNN (point-based).}

{Fig.~\ref{fig:downstreaming-task} shows the Car 3D bounding-box AP@0.7, 0.7, 0.7 as a function of bitrate for the original and reconstructed point clouds, evaluated with PointPillars, SECOND, and PointRCNN.
The dashed line indicates the detection accuracy on the uncompressed point clouds.
The results demonstrate that the proposed method achieves higher detection accuracy than the baselines across all three detectors in low-bpp regimes, i.e., bpp from 1.0 to 2.0.}

\subsubsection{DECODING LATENCY}
Table~\ref{tab:decoding-time} shows the average decoding latency of the proposed and baseline methods for LiDAR {frame 00 of sequence 00.} 
The decoding latency values for the proposed method and the baselines are the total time required from \ac{RI} decoding to the 2D-to-3D mapping. 
The decoding latency of the proposed method is comparable to that of image compression methods and has more than 65.5\% and 93.3\% reduction compared to G-PCC, Draco and R-PCC methods, respectively. 
{
In addition, the proposed scheme achieves a speedup of over four orders of magnitude compared to OctAttention. }
This means that the proposed method approaches the decoding latency of image compression methods and achieves \ac{3D} reconstruction quality comparable to \ac{PCC}/R-PCC methods.

\subsubsection{ENCODING LATENCY}
Table~\ref{tab:encoding-time} lists the average encoding latency of the proposed and baseline methods for LiDAR {frame 00 of sequence 00.} 
Here, the encoding latency for the RI-based schemes contains the conversion time from the point cloud to the RI. 
It can be seen that INR-based approaches, including the proposed one, involve significantly longer encoding time than both 3D point cloud and 2D image compression methods.
However, as shown in Table~\ref{tab:decoding-time}, INR-based compression drastically reduces decoding latency, which is advantageous for on-demand, quality-driven services.

{We note that the proposed scheme has a trade-off between reconstruction quality and encoding latency depending on the learning rate schedule. 
Here, we use initial and minimum learning rates of $1\times10^{-4}$ and $1\times10^{-12}$ for quality-sensitive users.
When we use initial and minimum learning rates of $1\times10^{-6}$ and $1\times10^{-8}$ for encoding, the encoding latency is reduced to 30 minutes with a slight degradation in reconstruction quality.
This means the proposed scheme can select quality-oriented or latency-oriented configurations based on application requirements.}

\begin{figure}[t]
  \centering 
  \subfloat[\ac{R-D} performance.]
    {\includegraphics[width=\linewidth]{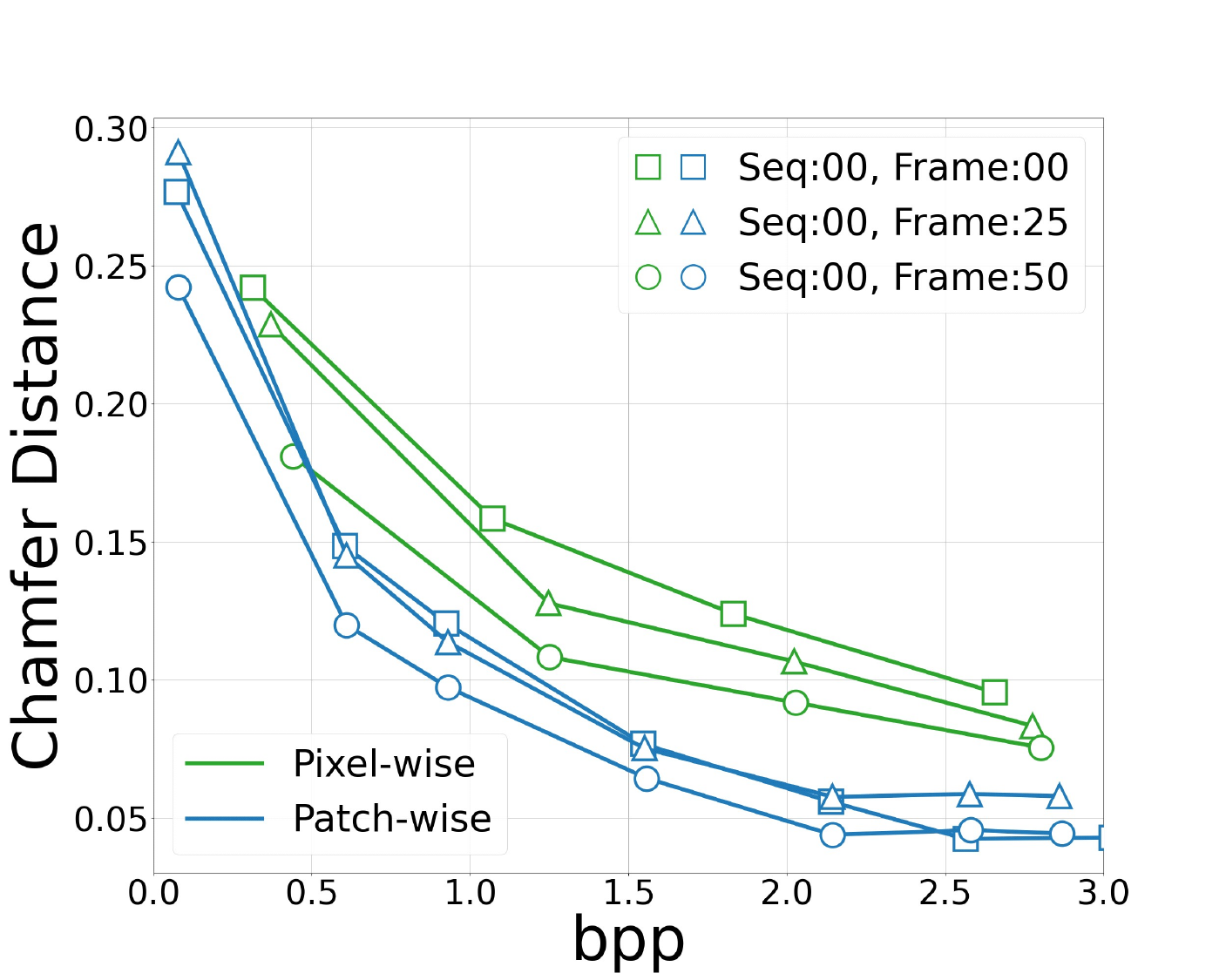}}
  \\ 
  \subfloat[Convergence speed.]
    {\includegraphics[width=\linewidth]{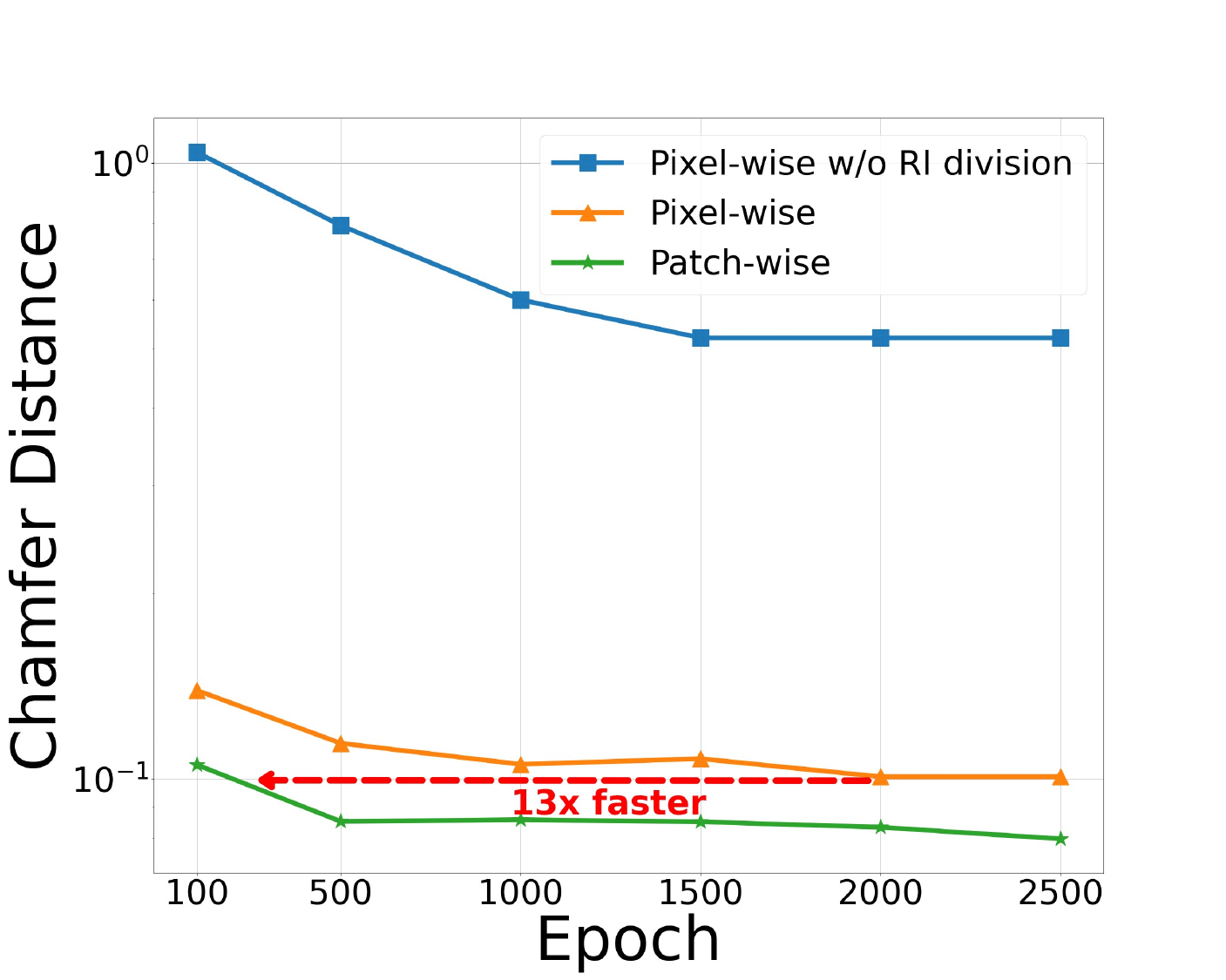}}
  \caption{Patch-wise vs. pixel-wise.}
  \label{fig:patch2}
\end{figure}

\begin{figure}[t!]
    \centering
    \begin{tabular}{cc}
        \begin{minipage}[t]{0.445\hsize}
            \centering
            \subfloat[Pixel-wise \\
            bpp:1.83 \\
            CD:0.123]
            {\includegraphics[width=\hsize]{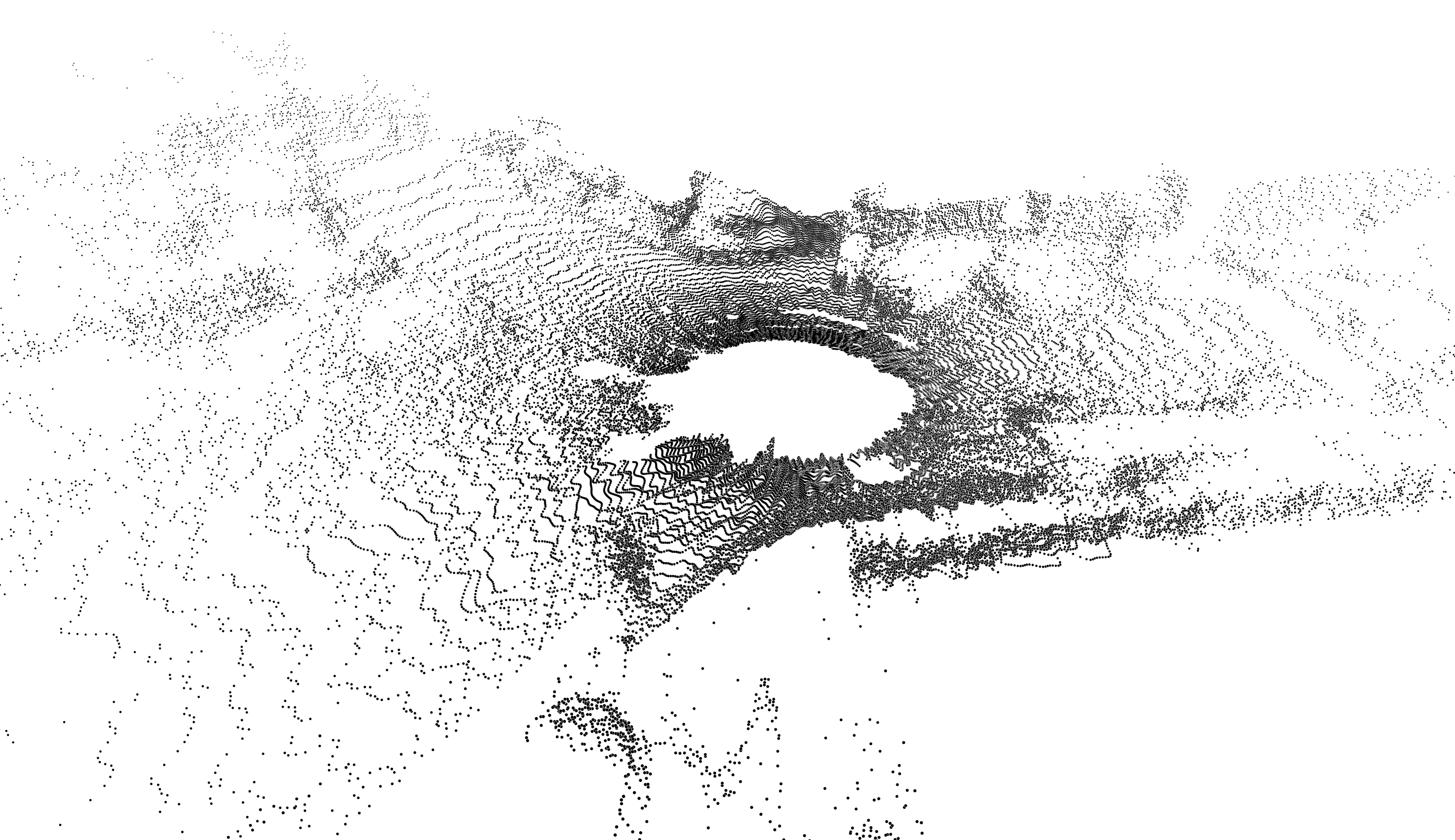}}
        \end{minipage} &

        \begin{minipage}[t]{0.445\hsize}
            \centering
            \subfloat[Patch-wise \\
            bpp:2.14 \\
            CD:0.056]
            {\includegraphics[width=\hsize]{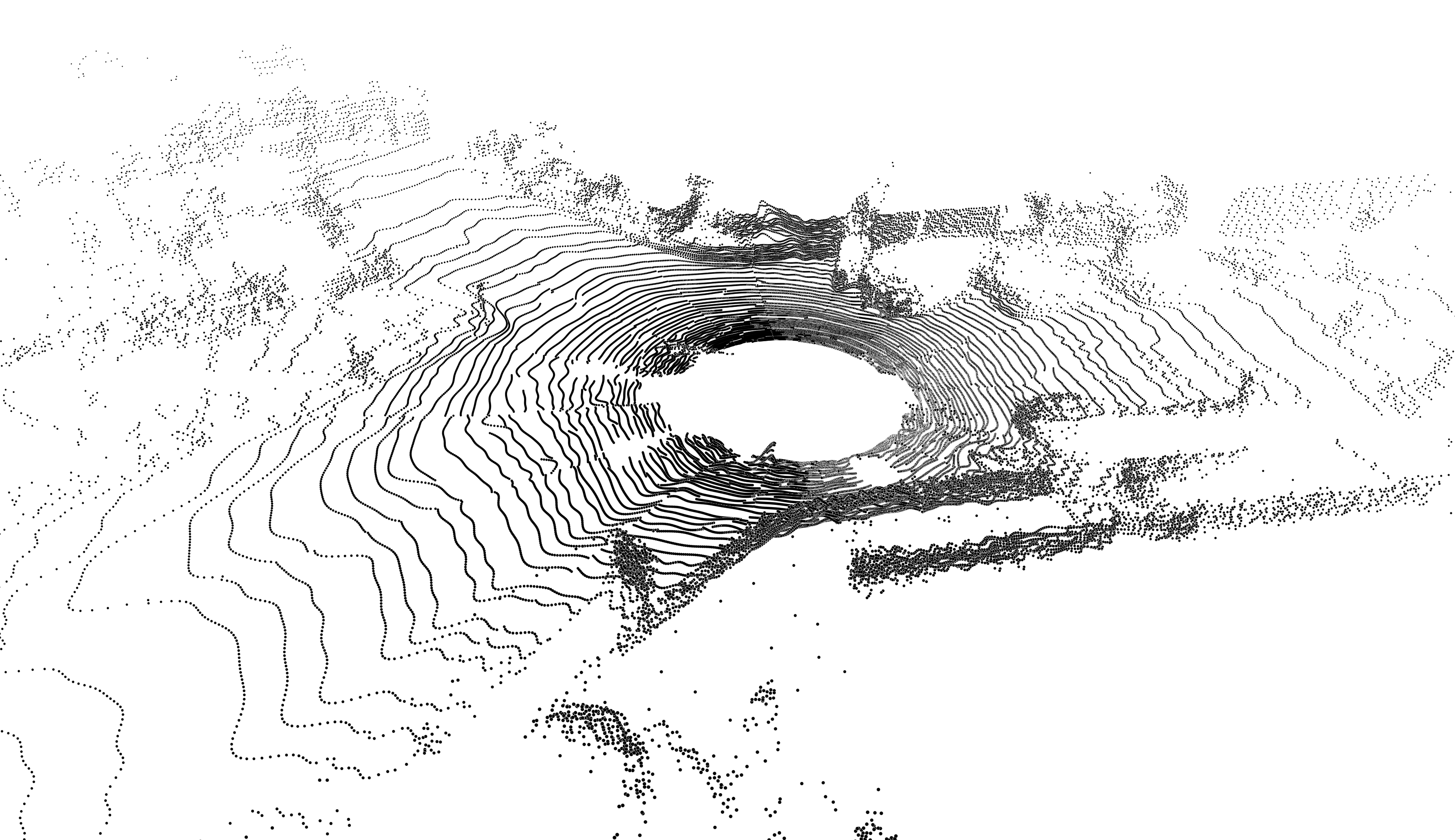}}
        \end{minipage} \\

        \begin{minipage}[t]{0.445\hsize}
            \centering
            \subfloat[Pixel-wise \\
            bpp:2.02 \\
            CD:0.107]
            {\includegraphics[width=\hsize]{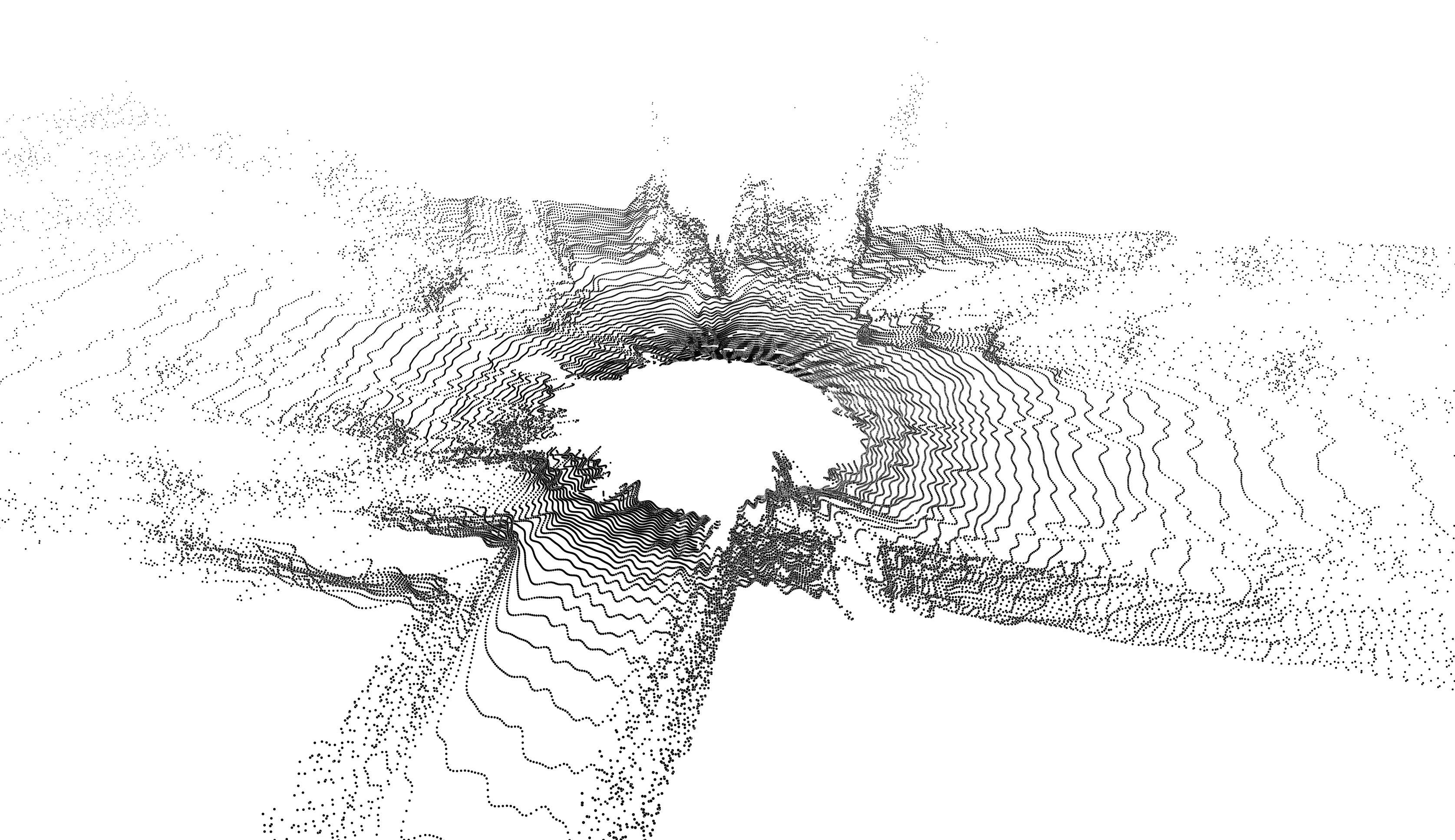}}
        \end{minipage} &

        \begin{minipage}[t]{0.445\hsize}
            \centering
            \subfloat[Patch-wise \\
            bpp:2.16 \\
            CD:0.058]
            {\includegraphics[width=\hsize]{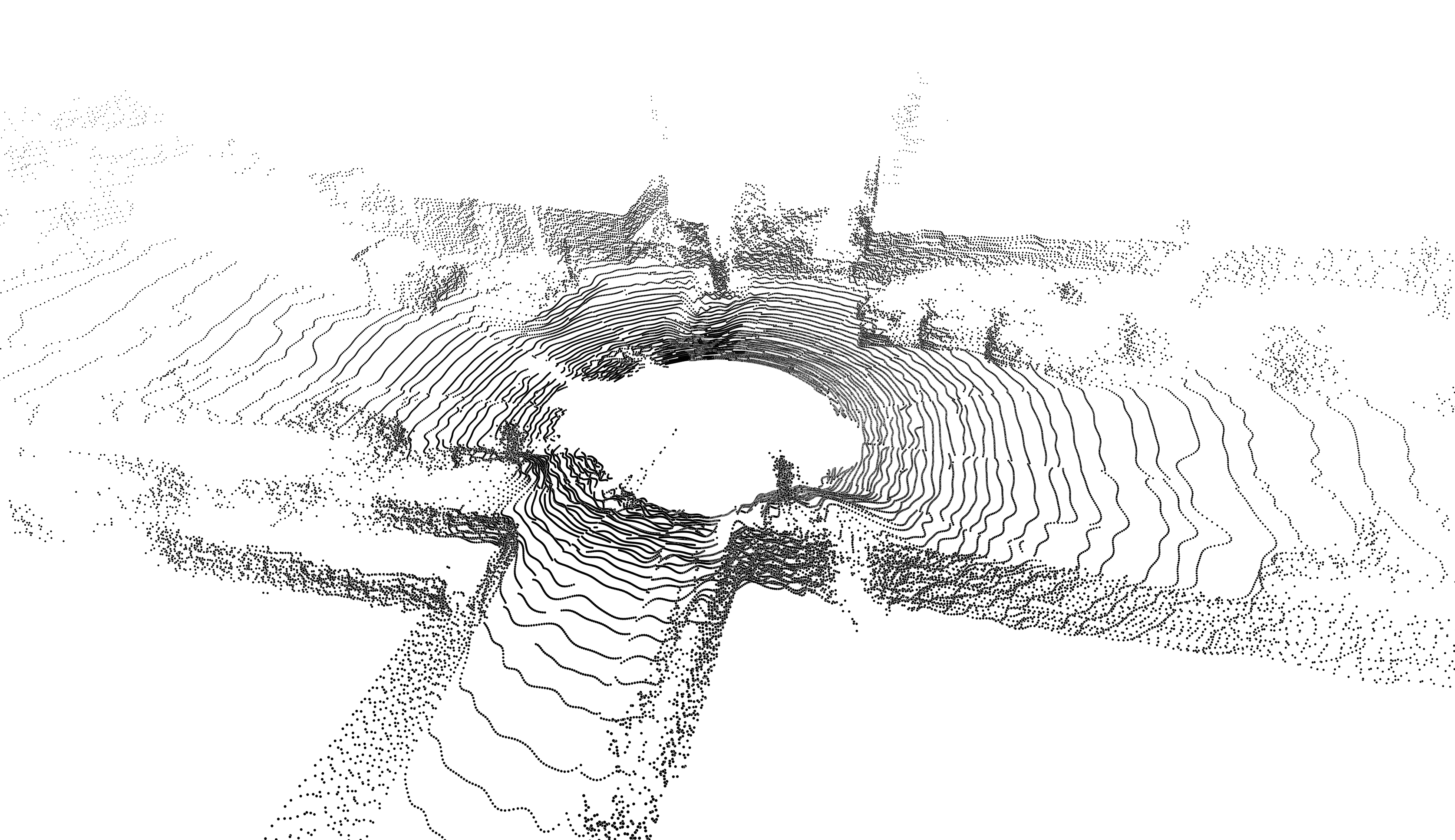}}
        \end{minipage} \\
        
    \end{tabular}
    \caption{ {Snapshot of the reconstructed LiDAR point clouds in pixel-wise and patch-wise proposed methods. Here, (a)-(b) and (c)-(d) show the reconstructed point clouds of frames 00 and 25 of sequence 00, respectively.}}
    \label{fig:patch_pixel_snap}

\end{figure}

\subsection{Ablation Study}

\subsubsection{IMPACT OF PATCH-WISE INR ARCHITECTURE}

The proposed depth INR exploits the patch-wise architecture, whereas the pixel-wise architecture can be used for the depth INR. 
Fig.~\ref{fig:patch2}~(a) shows the \ac{CD} of the proposed patch-wise INR and pixel-wise INR architectures as a function of bitrates under the different sequences of KITTI's {LiDAR} point cloud.
We can see that the patch-wise depth INR achieves better \ac{CD} than the pixel-wise architecture at large bitrate regimes in every {LiDAR} sequence.
Specifically, \ac{BD-CD} between the patch-wise and pixel-wise architectures is 0.047, 0.031, and 0.028 in {frames 00, 25, and 50 of sequence 00, respectively.}
{The effects on the visual quality are shown in Figs.~\ref{fig:patch_pixel_snap}~(a)--(d), respectively.}

Fig.~\ref{fig:patch2}~(b) shows the \ac{CD} performance of \ac{INR}-based image compression methods as a function of the learning epochs.
Our patch-wise architecture boosts the convergence speed by up to 13$\times$ compared to the pixel-wise architecture, and the fast convergence results in a short encoding delay.

\begin{figure}[t]
  \centering
  \subfloat[CD for different pruning sparsity.]
  {\includegraphics[width=\linewidth]{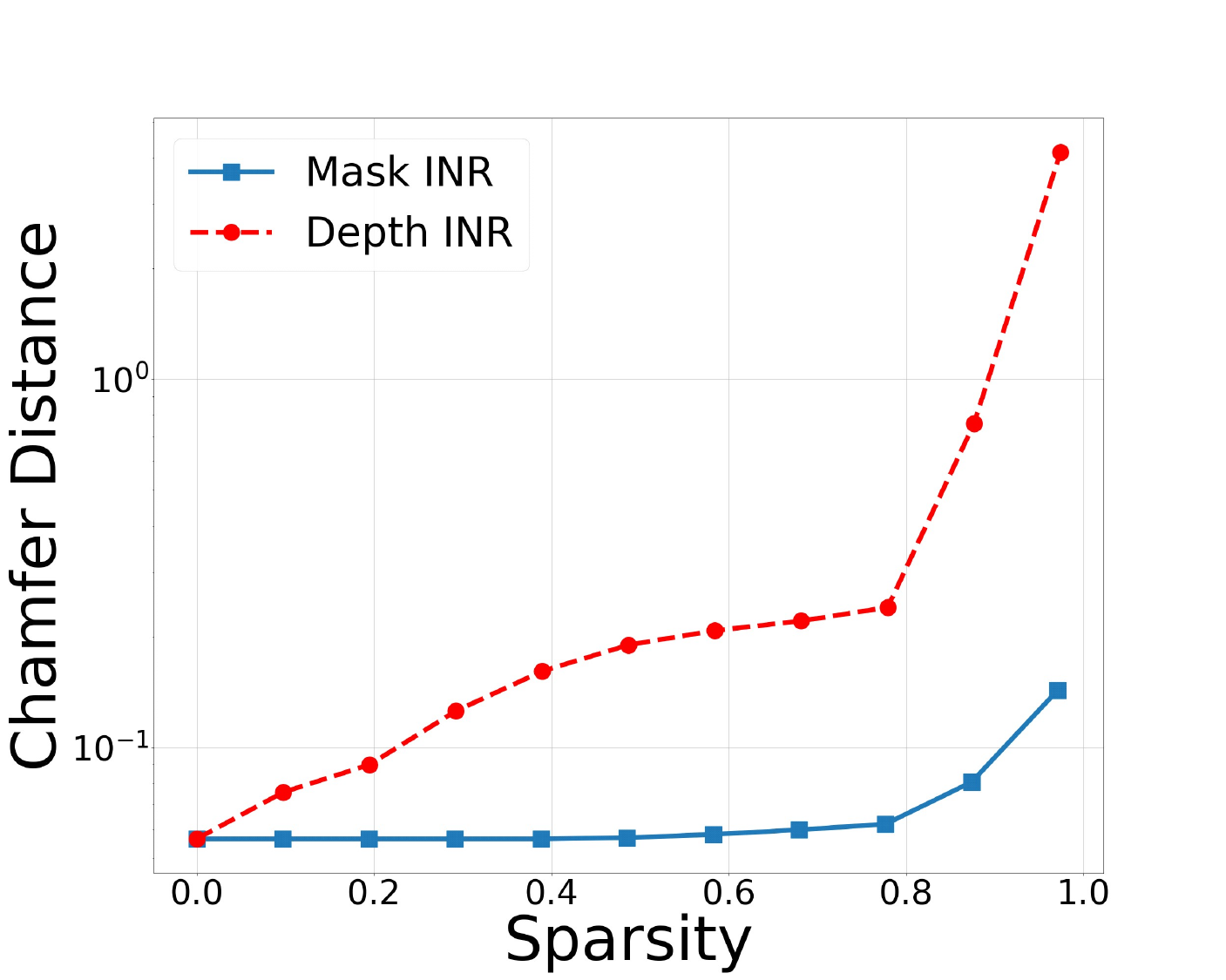}} \\
  \subfloat[Quantization with different $N_b$.]
  {\includegraphics[width=\linewidth]{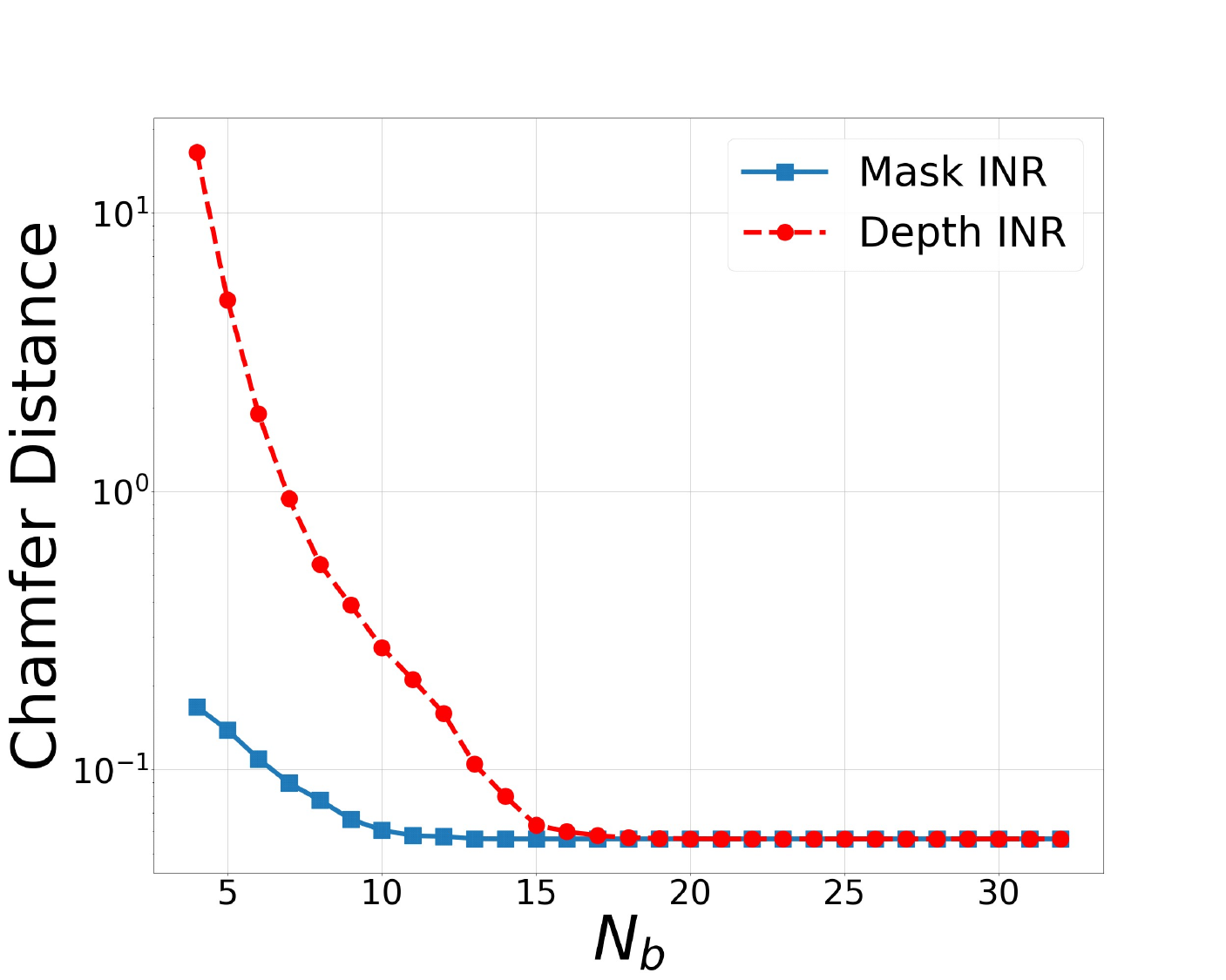}}
   \caption{Model compression performance.}
   \label{fig:compression}
\end{figure}

\subsubsection{IMPACT OF MODEL COMPRESSION}

After the encoder trains the depth and mask INR architectures, the trained weights are pruned and quantized for compression.  
Here, the proposed method can set different pruning ratios and bit depths for the depth and mask INRs. 
This section discusses the impact of model pruning and quantization on both INR model compression. 

Figs.~\ref{fig:compression}~(a) and (b) show the effect of model pruning and quantization for the depth and mask architectures. 
In pruning, the mask INR is similar in performance to the full model, although the sparsity is approximately 70\%.
However, pruning the model for the depth INR causes quality degradation even though the sparsity is only 10\%. 
For quantization, a 16-bit model still retains almost the same \ac{CD} as the original 32-bit model in depth INR, while the mask INR can reduce the number of bits to 11.

\subsubsection{IMPACT OF NETWORK ARCHITECTURE}

\begin{figure}[t]
  \centering
  {\includegraphics[width=.8\linewidth]{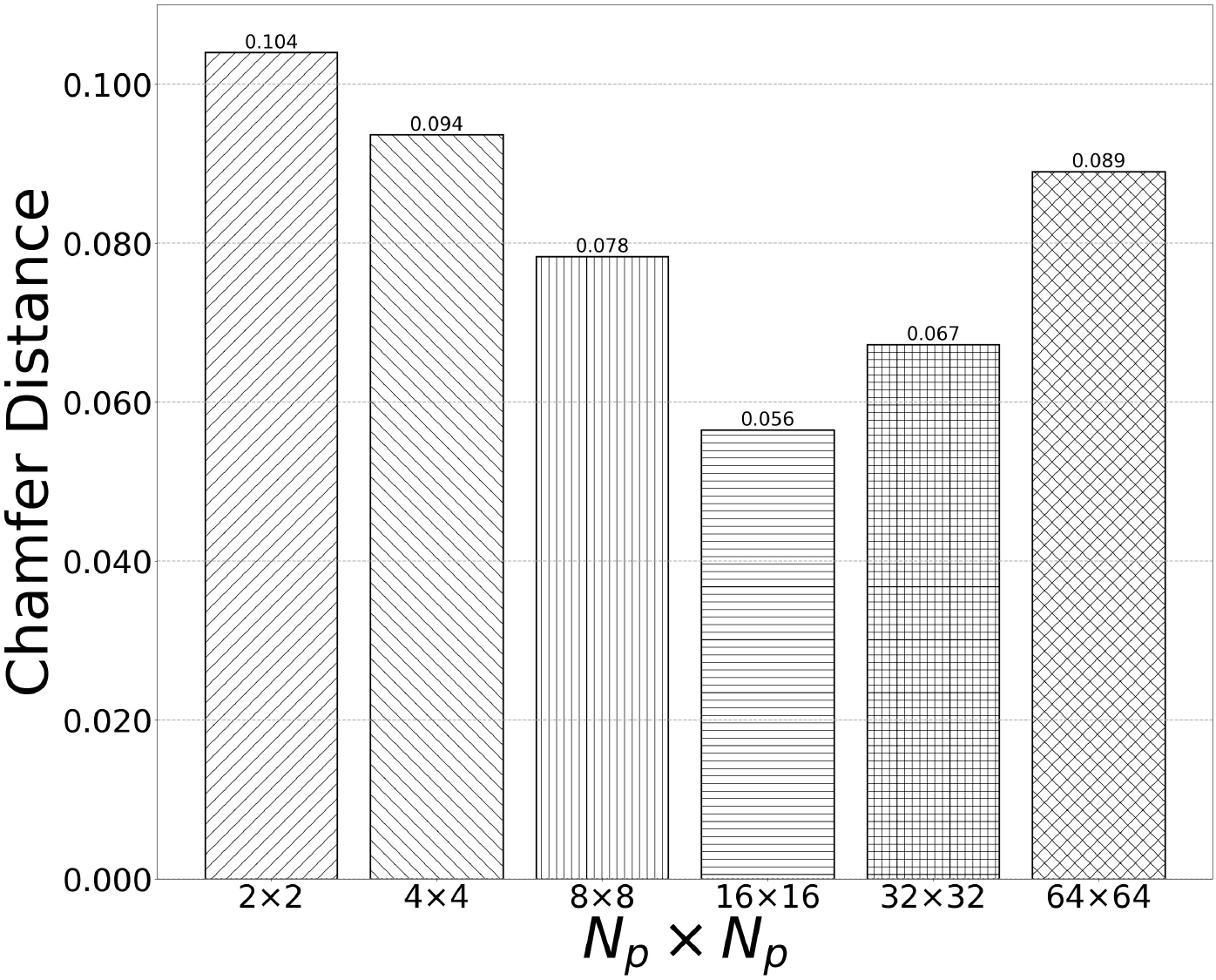}} \\
  \caption{Chamfer distance under the different patch sizes.}
   \label{fig:patch}
\end{figure}

\begin{figure}[t!]
    \centering
    \begin{tabular}{cc}
        \begin{minipage}[t]{0.44\hsize}
            \centering
            \subfloat[{Seq:~00, Frame:~00}]
            {\includegraphics[width=\hsize]{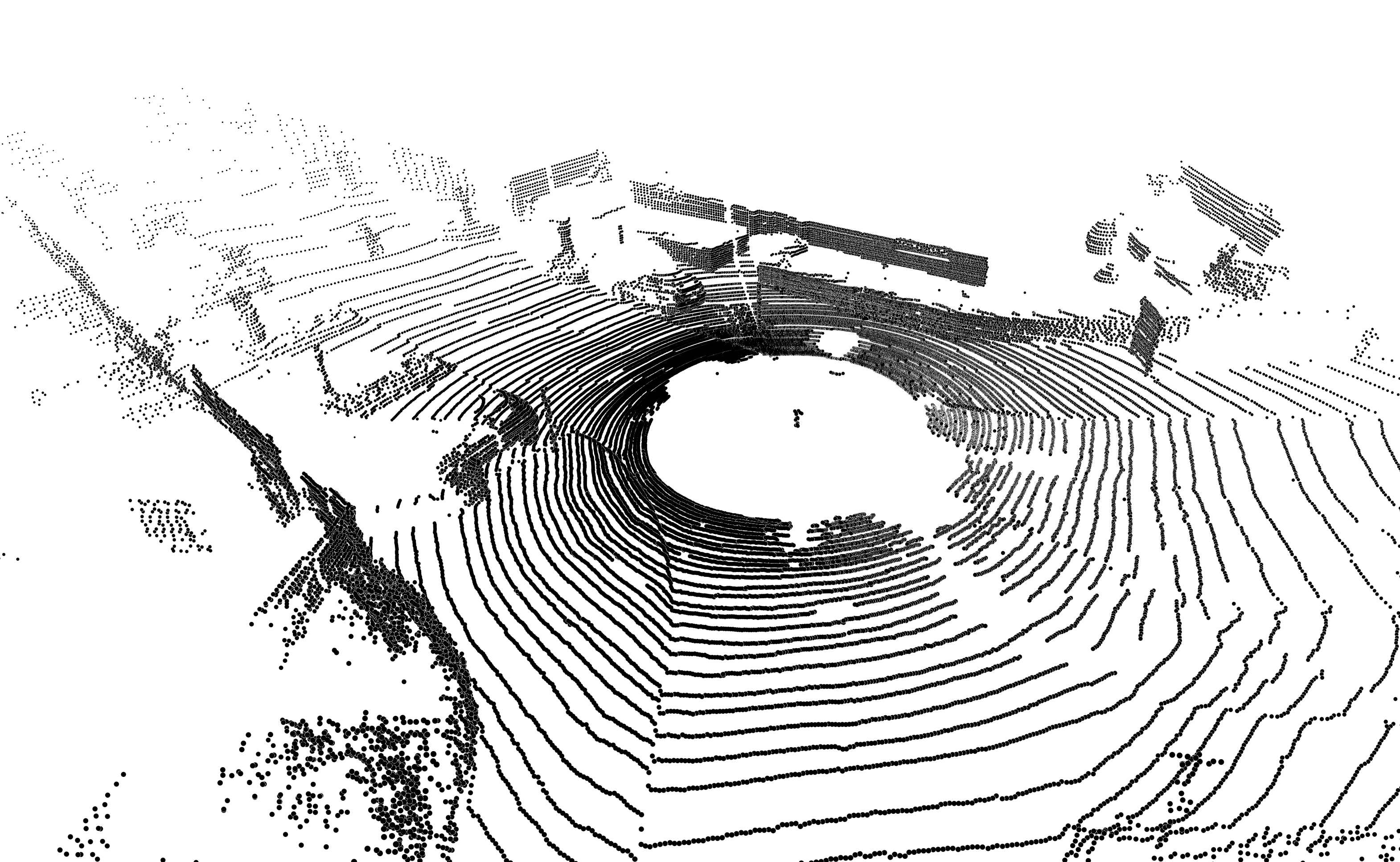}}
        \end{minipage} &

        \begin{minipage}[t]{0.44\hsize}
            \centering
            \subfloat[\centering $N_p$:~2, CD:~0.104]
            {\includegraphics[width=\hsize]{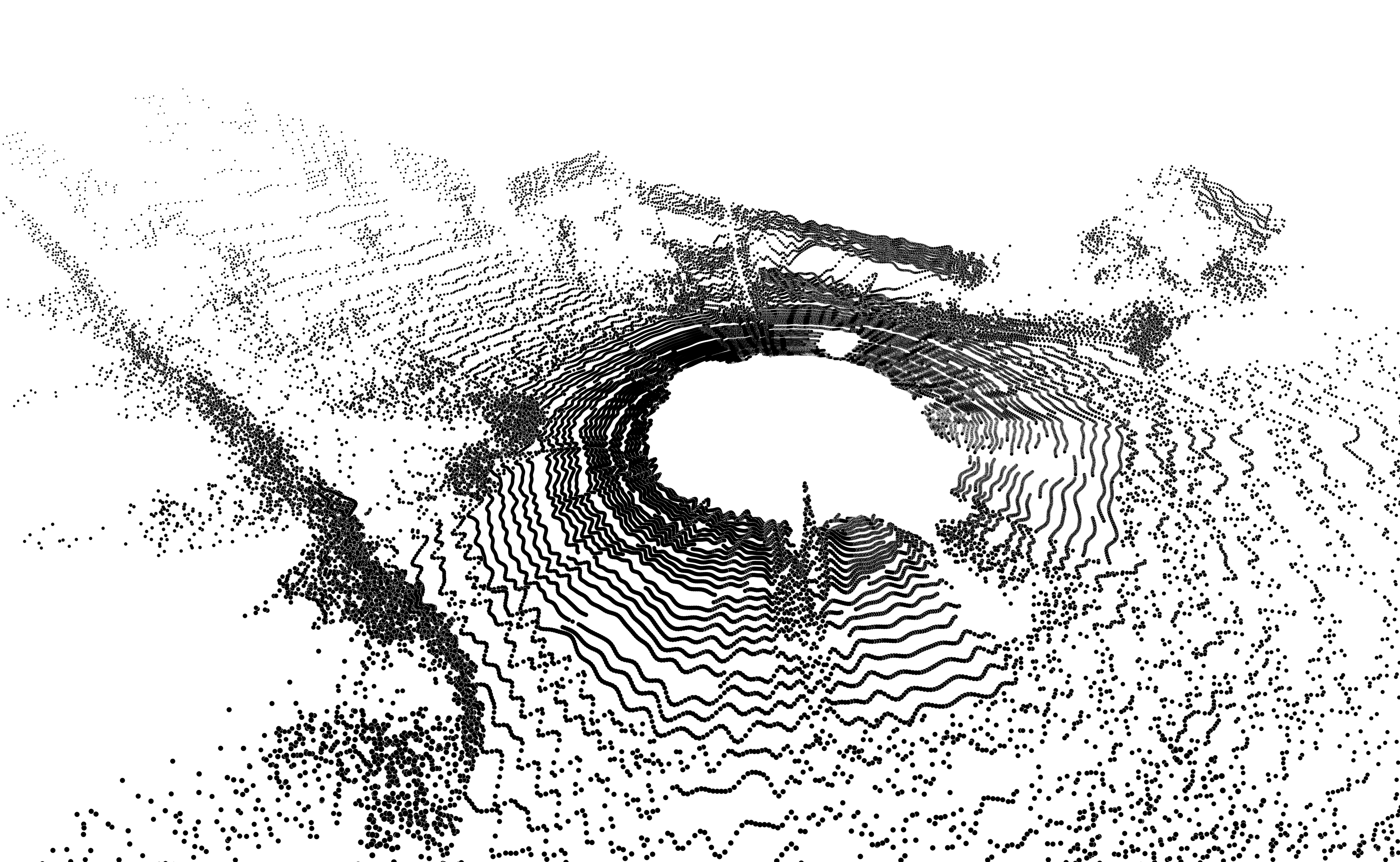}}
        \end{minipage} \\

        \begin{minipage}[t]{0.44\hsize}
            \centering
            \subfloat[\centering $N_p$:~4, CD:~0.094]
            {\includegraphics[width=\hsize]{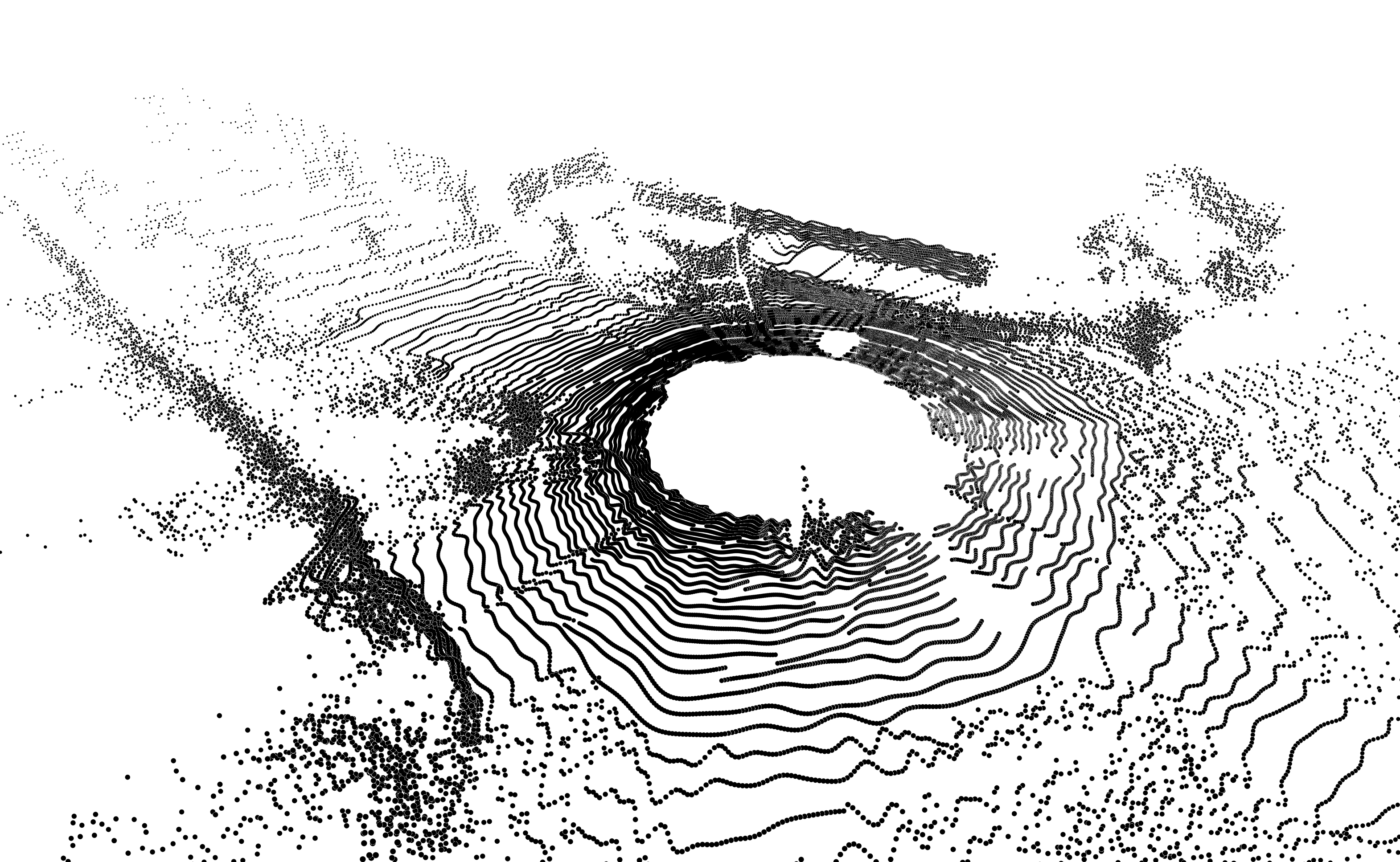}}
        \end{minipage} &

        \begin{minipage}[t]{0.44\hsize}
            \centering
            \subfloat[\centering $N_p$:~8, CD:~0.078]
            {\includegraphics[width=\hsize]{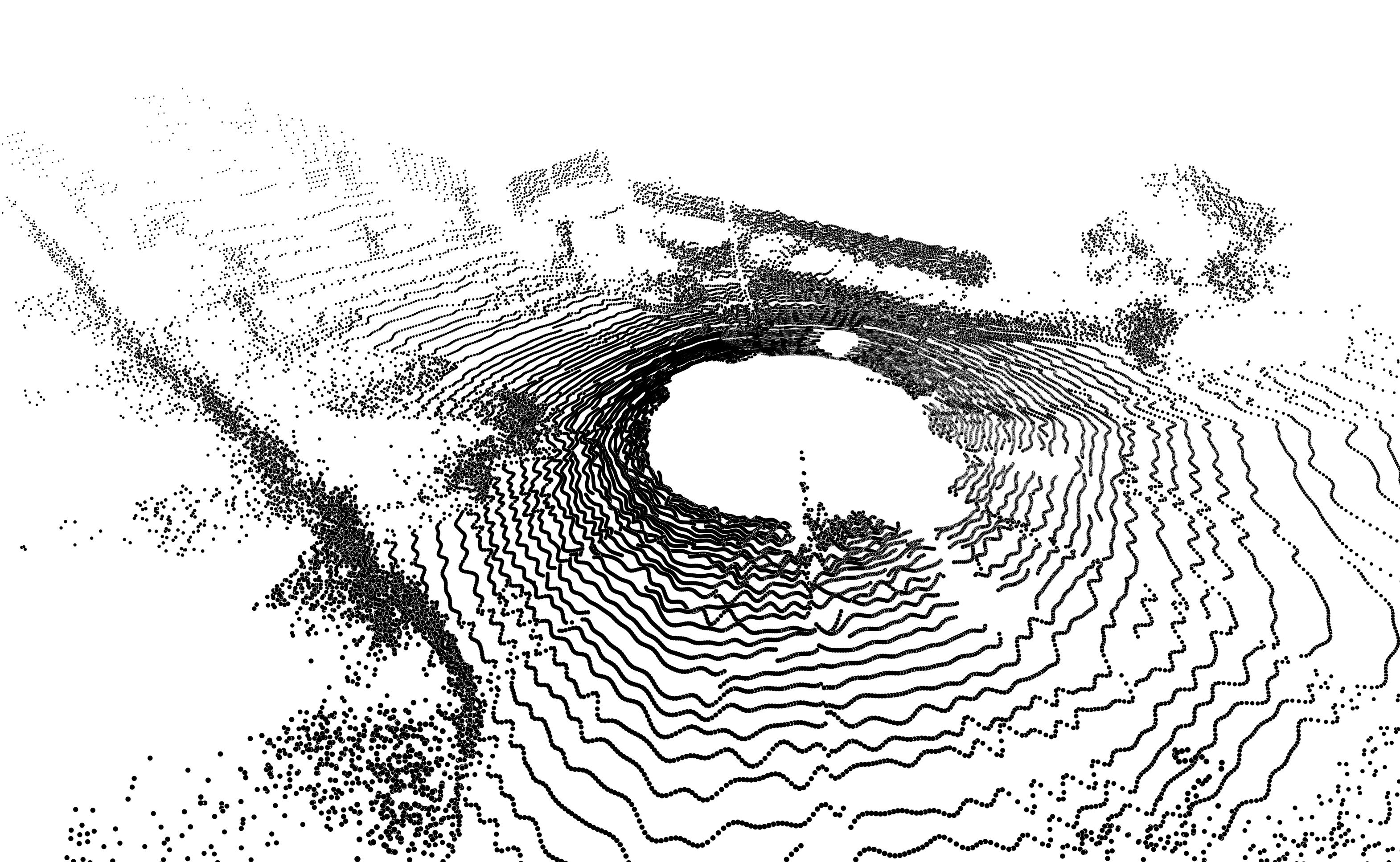}}
        \end{minipage} \\

        \begin{minipage}[t]{0.44\hsize}
            \centering
            \subfloat[\centering $N_p$:~\textbf{16}, CD:~\textbf{0.056}]
            {\includegraphics[width=\hsize]{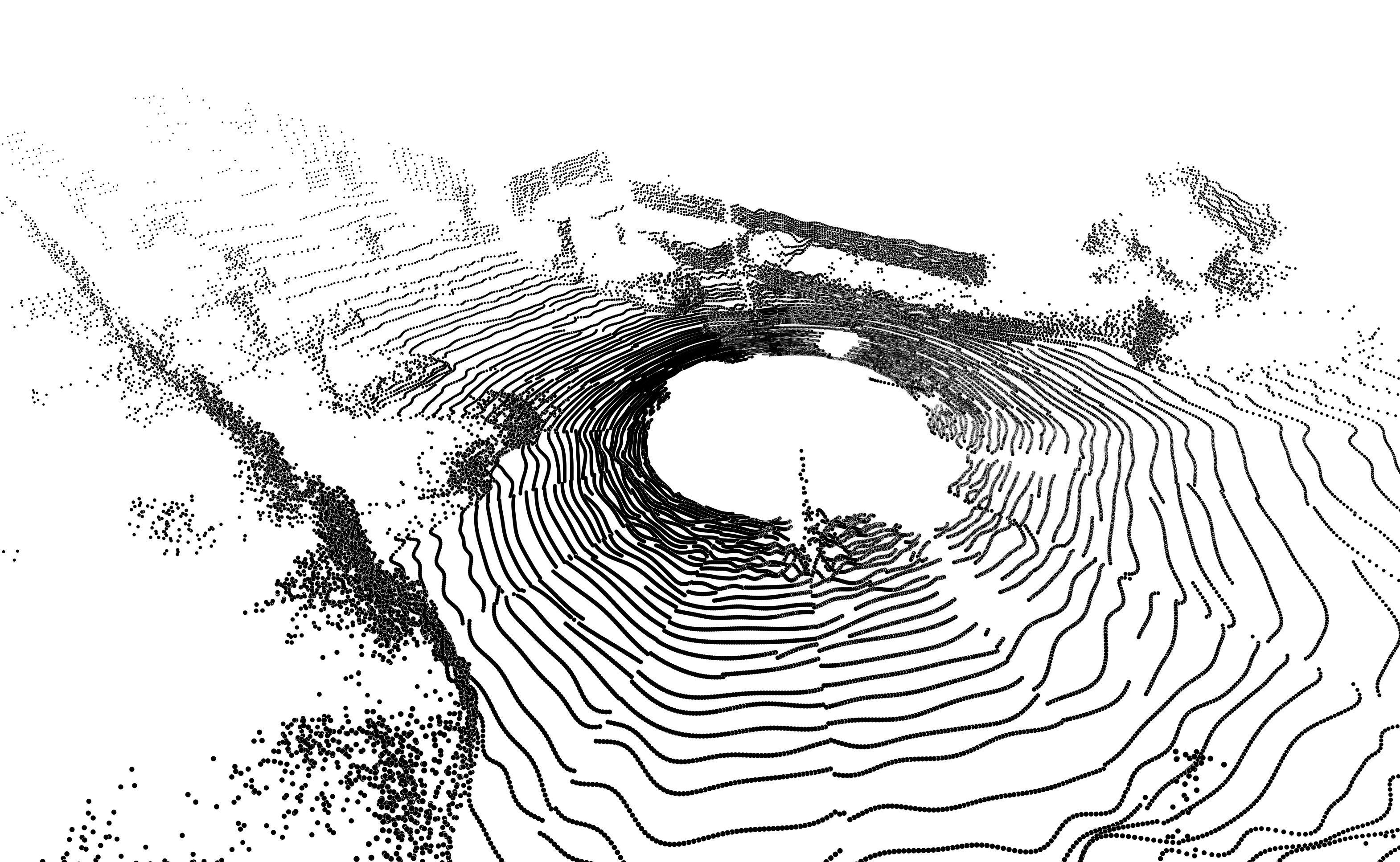}}
        \end{minipage} &

        \begin{minipage}[t]{0.44\hsize}
            \centering
            \subfloat[\centering $N_p$:~32, CD:~0.067]
            {\includegraphics[width=\hsize]{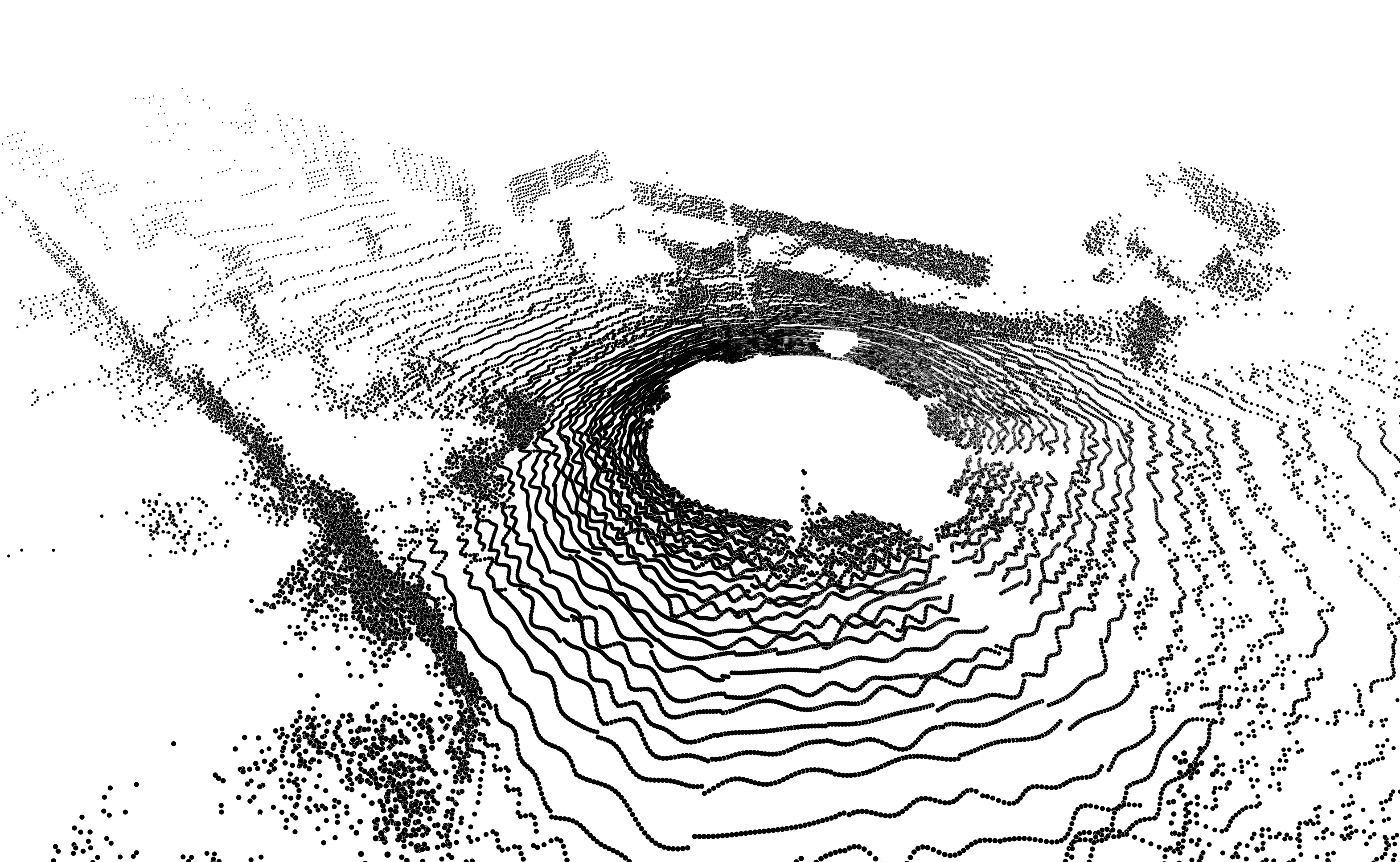}}
        \end{minipage} \\
        
        \begin{minipage}[t]{0.44\hsize}
            \centering
            \subfloat[\centering $N_p$:~64, CD:~0.089]
            {\includegraphics[width=\hsize]{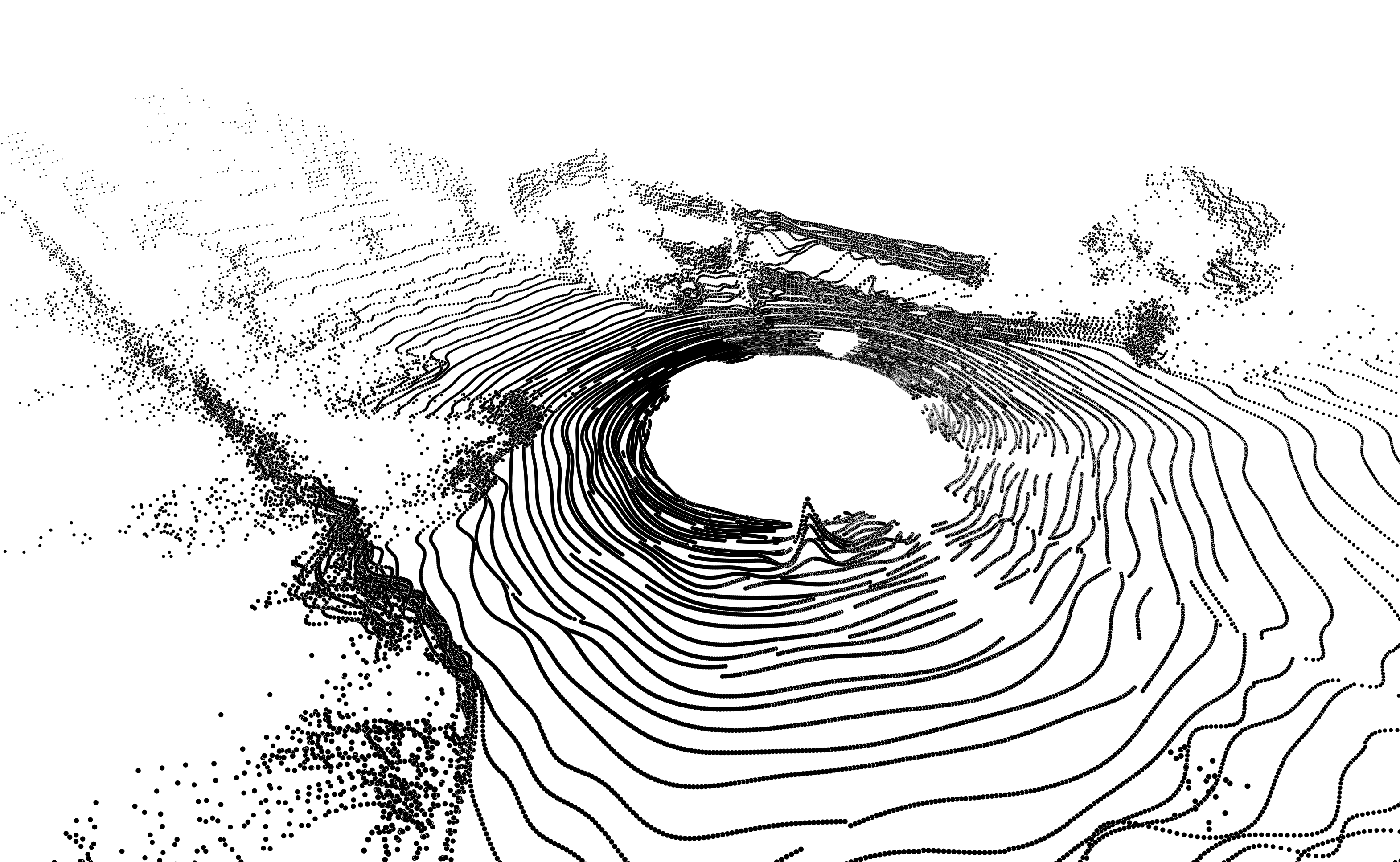}}
        \end{minipage} 
    \end{tabular}
    \caption{Snapshots of the reconstructed {LiDAR} point clouds in proposed methods under the different patch sizes $N_p \times N_p$. Here, (b)-(g) show the reconstructed point clouds of {frame 00 of sequence 00.}}
    \label{fig:patch_snap}

\end{figure}

\begin{figure}[t]
  \centering
  {\includegraphics[width=.8\linewidth]{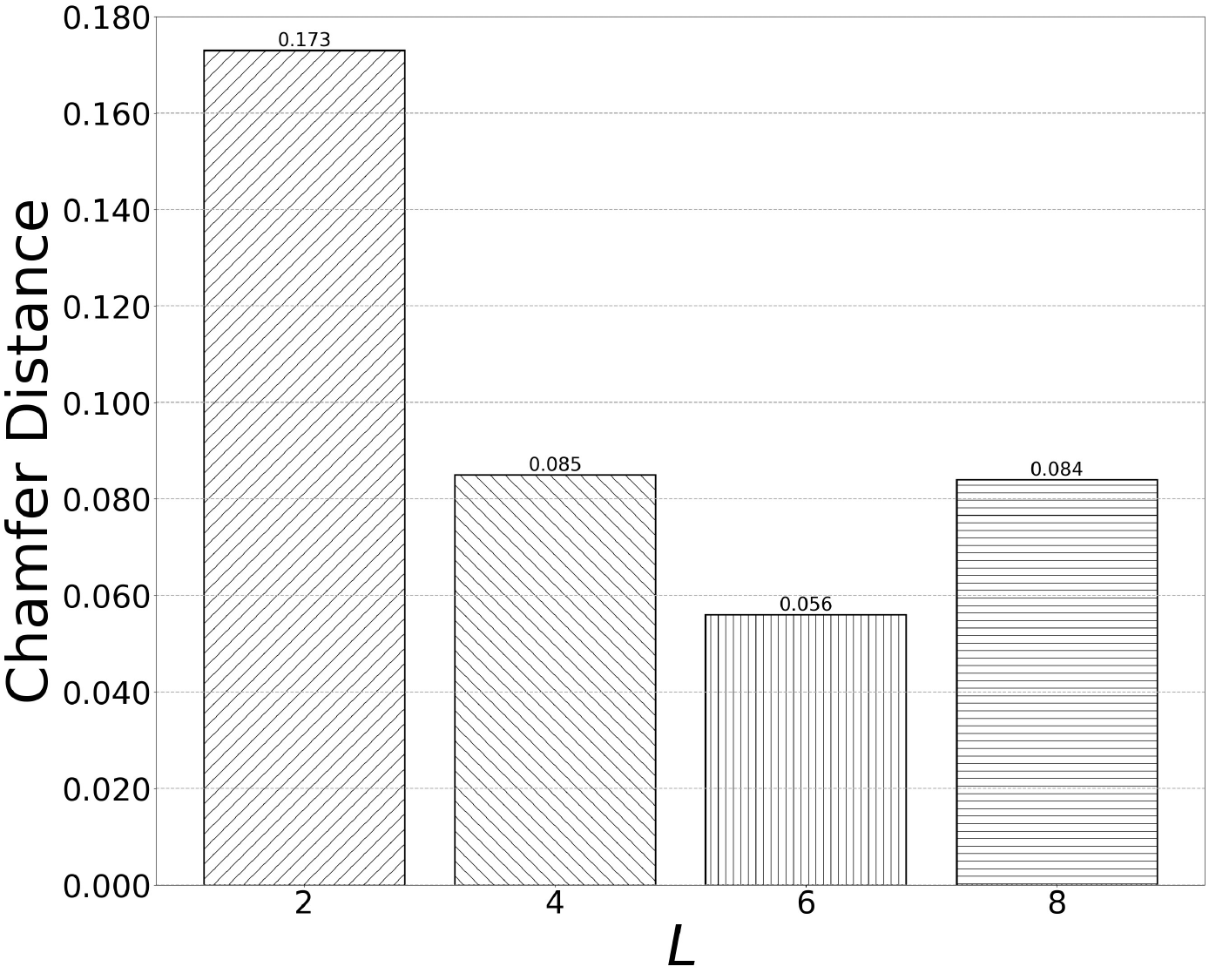}} \\
  \caption{Chamfer distance under the different layer sizes.}
   \label{fig:nl}
\end{figure}

\begin{figure}[t!]
    \centering
    \begin{tabular}{cc}
        \begin{minipage}[t]{0.44\hsize}
            \centering
            \subfloat[{Seq:~00, Frame:~00}]
            {\includegraphics[width=\hsize]{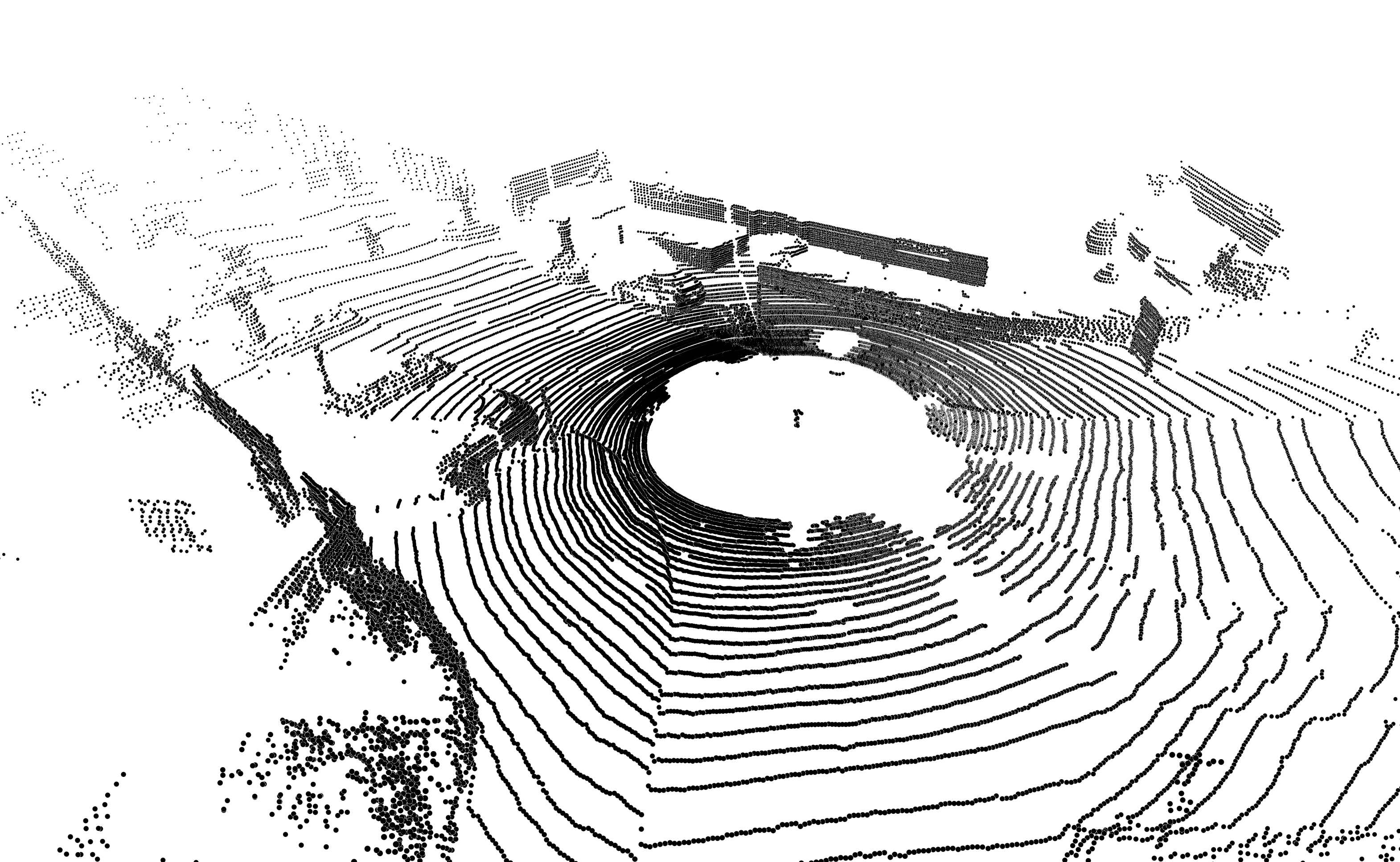}}
        \end{minipage} &

        \begin{minipage}[t]{0.44\hsize}
            \centering
            \subfloat[\centering $L$:~2, CD:~0.173]
            {\includegraphics[width=\hsize]{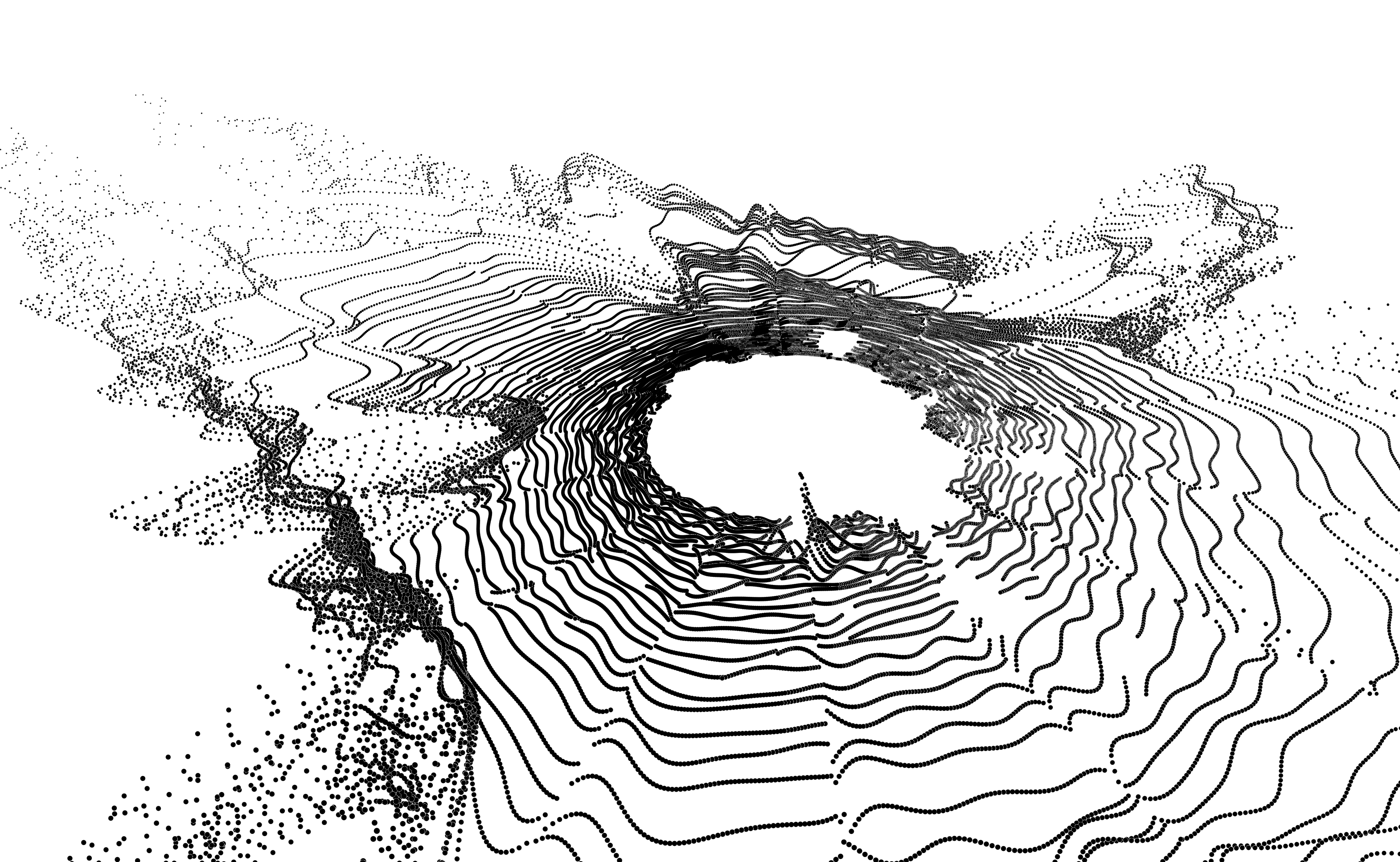}}
        \end{minipage} \\

        \begin{minipage}[t]{0.44\hsize}
            \centering
            \subfloat[\centering $L$:~4, CD:~0.085]
            {\includegraphics[width=\hsize]{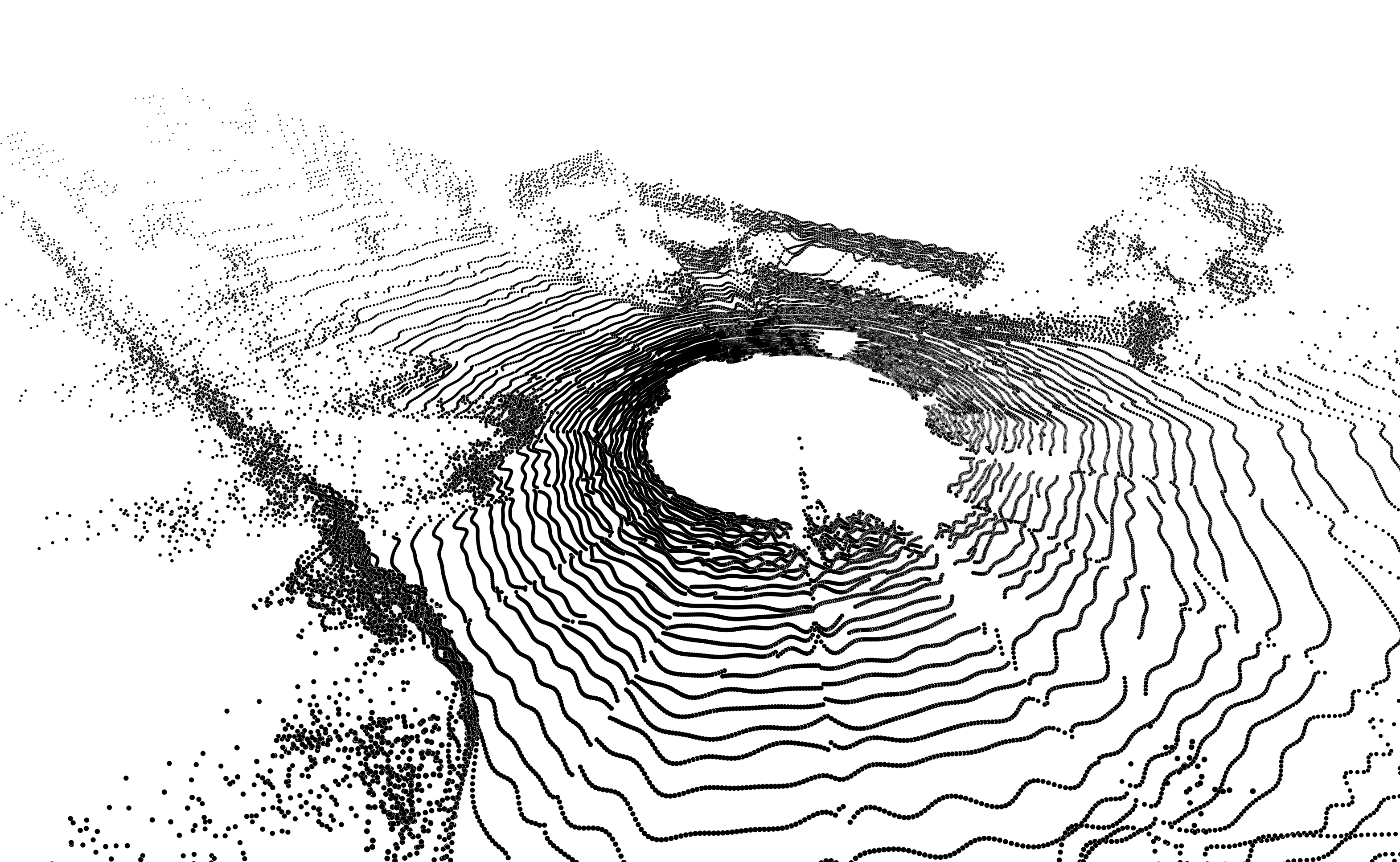}}
        \end{minipage} &

        \begin{minipage}[t]{0.44\hsize}
            \centering
            \subfloat[\centering $L$:~\textbf{6}, CD:~\textbf{0.056}]
            {\includegraphics[width=\hsize]{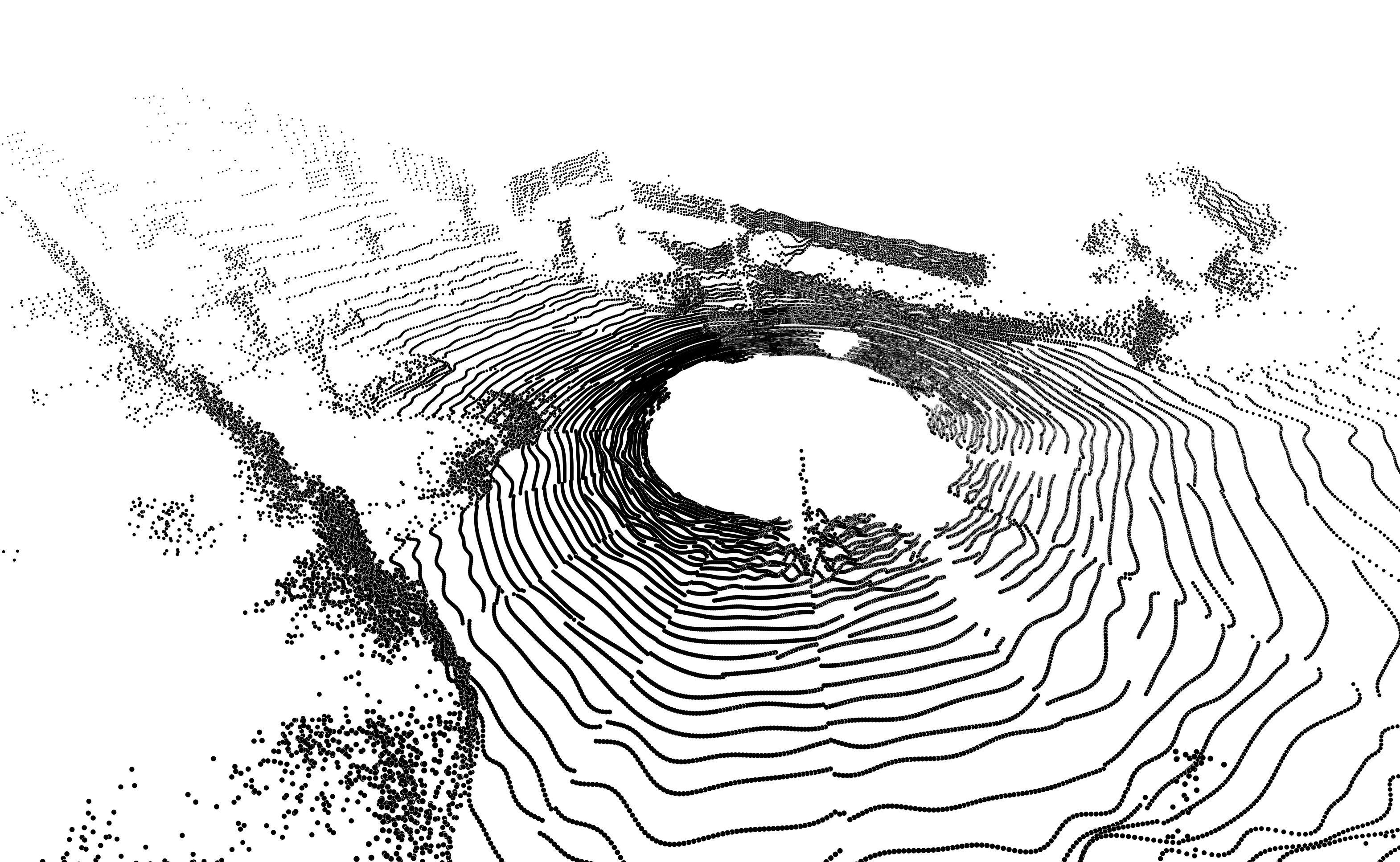}}
        \end{minipage} \\

        \begin{minipage}[t]{0.44\hsize}
            \centering
            \subfloat[\centering $L$:~8, CD:~0.084]
            {\includegraphics[width=\hsize]{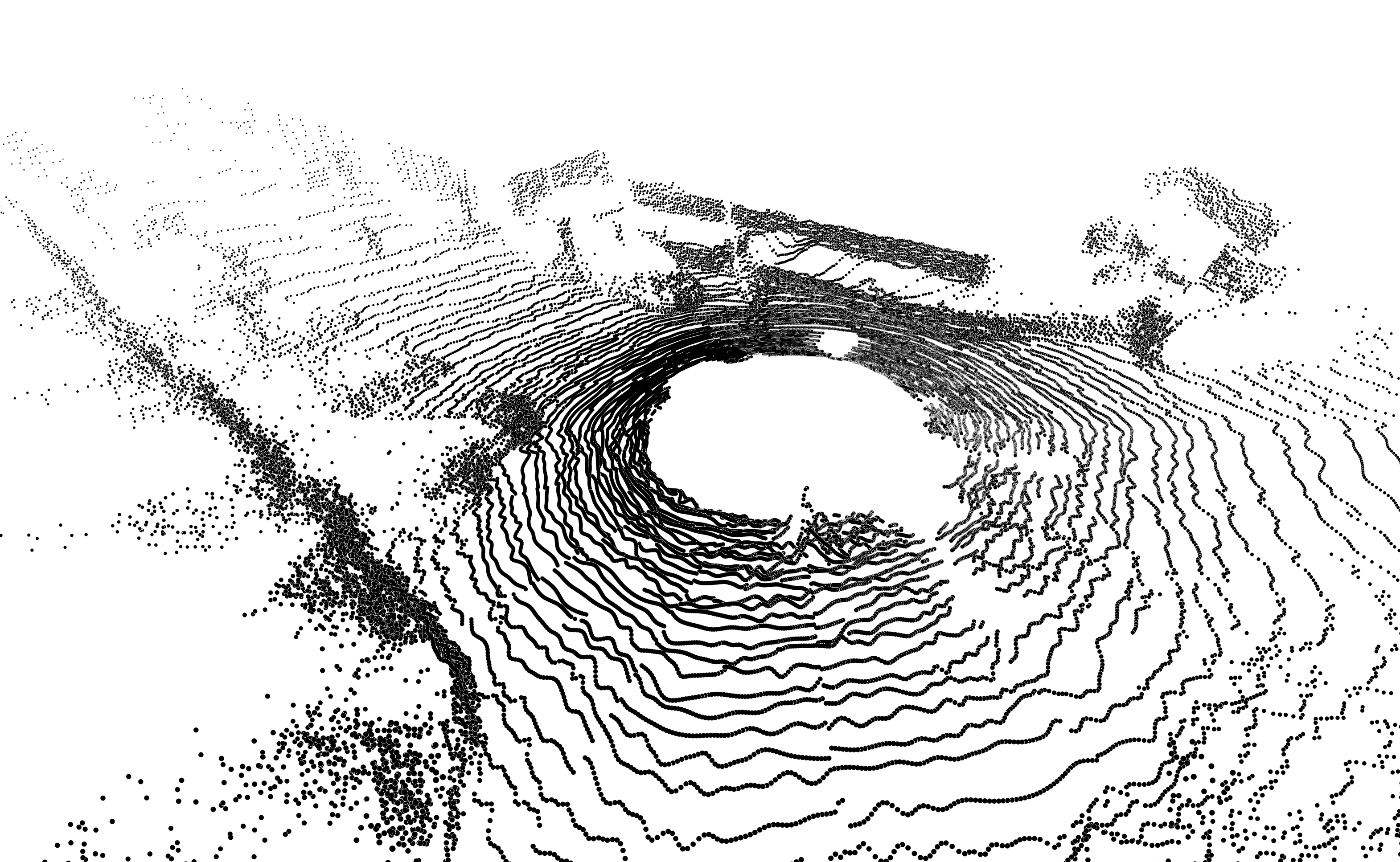}}
        \end{minipage} \\
    \end{tabular}
    \caption{Snapshots of the reconstructed {LiDAR} point clouds in proposed methods under the different layer sizes $L$. Here, (b)-(e) show the reconstructed point clouds of {frame 00 of sequence 00.}}
    \label{fig:nl_snap}

\end{figure}

This section discusses the effect of the configurations for the depth INR architecture, specifically the patch size $N_p$ and layer size $L$, on the quality of the reconstructed {LiDAR} point cloud.
The proposed depth INR uses the patch-wise architecture, and thus the depth image is divided into patches of size \(N_p \times N_p\). 
Here, a small patch size increases the complexity of intra-patch learning, while a large patch size increases the complexity of inter-patch learning.

Fig.~\ref{fig:patch} shows \ac{CD} performance varying \(N_p\), and Fig.~\ref{fig:patch_snap} demonstrates the corresponding snapshots of the reconstructed point cloud. 
We consider all the variants using the same model size in Fig.~\ref{fig:compression} with a sparsity of 0.0 and $N_b$ of 32. 
The evaluation results demonstrated that the patch size of $N_p=16$ yields the best CD performance. 
{However, using larger or smaller $N_p$ values leads to performance degradation.
While a large $N_p$ reduces the effectiveness of patch-wise modeling due to the limited pixel count, a small $N_p$ requires covering a wider area. 
This makes it challenging to capture sharp depth transitions and preserve geometric details near boundaries.}

Similarly, Fig.~\ref{fig:nl} and \ref{fig:nl_snap} show the 3D reconstruction quality of the proposed scheme and the corresponding snapshots for different layer sizes $L$.
The results indicate that a layer size of $L=6$ is the most effective for CD performance.




\section{Conclusion and Future Work}
We proposed a novel \ac{RI}-based {LiDAR} point cloud compression method. 
The proposed method is designed to efficiently compress floating-point RIs using INR-based techniques and features a sophisticated architecture that combines separated learning for mask and depth images, patch-wise learning for depth images, and model compression.
{Experiments on the KITTI dataset show that the proposed method improves 3D reconstruction quality at low bitrates compared with conventional image codecs and representative baselines such as G-PCC, Draco, and COIN, and it also achieves strong 3D object detection accuracy in the low-bpp regime.}

The proposed method has two limitations: encoding delay and transformation loss from RIs to 3D point clouds.
While existing baselines require only a few milliseconds for encoding, implicit neural compression, including the proposed method, takes from tens of minutes to several hours.
In summary, the proposed method's long encoding delay and short decoding delay make it well-suited for offline applications of LiDAR point clouds.
{To further reduce encoding latency, recent findings on learned initializations for coordinate-based neural representations~\cite{tancik2021learned} and meta-learned sparse INRs~\cite{lee2021meta} can be integrated into our depth/mask INR architecture. We leave the implementation and evaluation of such integration as future work.}

In addition, RIs with limited spatial resolution will lead to irreversible point loss during 2D-to-3D decoding, potentially degrading the performance of downstream tasks. 
In future work, we will consider integrating a point cloud generator~\cite{2024arXiv240406715K} to obtain a denser point cloud from the limited resolution of RIs.

\section*{APPENDIX}
This appendix provides further details for Table~\ref{tab:bd-cd}.
Table~\ref{tab:bd-cd-detail} shows the detailed BD-CD performance across the different LiDAR frames.

\begin{table*}[t]
    \caption{The list of BD-CD $\uparrow$ for the KITTI dataset across the different frames. {Note that BD-CD is evaluated for each baseline using the proposed method as the reference. 
   Positive values indicate that the proposed method achieves a lower chamfer distance than the corresponding baseline.}}
    \scriptsize
    \centering
    \begin{threeparttable}
        \begin{tabular}{cc c c c c c c c c c c} \toprule
            \textbf{Seq.} & \textbf{Frame} & \textbf{JPEG}$\dagger$ & \textbf{JPEG2000}$\dagger$ & \textbf{HEIF}$\dagger$ & \textbf{AVIF}$\dagger$ & \textbf{COIN}$\ddag$ & \textbf{G-PCC}$\S$ & \textbf{Draco}$\S$ & \textbf{\begin{tabular}{c}Oct \\ Attention\end{tabular}}$\S$ & \textbf{\begin{tabular}{c}R-PCC \\ (Deflate)\end{tabular}}$\P$ & \textbf{\begin{tabular}{c}R-PCC \\ (LZ4)\end{tabular}}$\P$ \\ \midrule
            \multirow{5}{*}{00} & 00 & 1.197 & 0.661 & 0.465 & 0.260 & 1.039 & 0.099 & 0.146 & -0.022 & 0.023 & 0.108 \\
            & 25 & 1.277 & 0.719 & 0.490 & 0.237 & 0.969 & 0.071 & 0.146 & -0.032 & 0.028 & 0.118 \\
            & 50 & 1.588 & 0.923 & 0.607 & 0.266 & 1.264 & 0.081 & 0.153 & -0.025 & 0.050 & 0.136 \\
            & 75 & 1.696 & 1.036 & 0.661 & 0.270 & 0.983 & 0.068 & 0.153 & -0.021 & 0.016 & 0.100 \\
            & 100 & 1.208 & 0.667 & 0.450 & 0.247 & 1.261 & 0.088 & 0.189 & -0.027 & 0.000 & 0.085 \\
            \midrule
            \multirow{5}{*}{01} & 00 & 1.118 & 0.630 & 0.474 & 0.258 & 1.072 & 0.084 & 0.265 & -0.036 & -0.005 & 0.074 \\
            & 25 & 1.040 & 0.581 & 0.462 & 0.285 & 1.073 & 0.091 & 0.292 & -0.013 & 0.005 & 0.082 \\
            & 50 & 1.128 & 0.624 & 0.485 & 0.271 & 1.113 & 0.121 & 0.238 & -0.011 & 0.019 & 0.088 \\
            & 75 & 1.411 & 0.754 & 0.546 & 0.244 & 0.969 & 0.115 & 0.177 & -0.005 & 0.021 & 0.080 \\
            & 100 & 1.472 & 0.842 & 0.606 & 0.239 & 1.250 & 0.075 & 0.199 & -0.018 & 0.057 & 0.159 \\
            \midrule
            \multirow{5}{*}{02} & 00 & 1.248 & 0.704 & 0.497 & 0.229 & 0.970 & 0.082 & 0.217 & -0.056 & -0.014 & 0.137 \\
            & 25 & 1.037 & 0.561 & 0.419 & 0.250 & 0.986 & 0.063 & 0.215 & -0.031 & 0.002 & 0.070 \\
            & 50 & 1.039 & 0.558 & 0.418 & 0.266 & 0.996 & 0.099 & 0.272 & -0.022 & 0.006 & 0.076 \\
            & 75 & 1.101 & 0.607 & 0.458 & 0.273 & 0.968 & 0.075 & 0.203 & -0.017 & 0.009 & 0.087 \\
            & 100 & 1.176 & 0.629 & 0.446 & 0.250 & 1.276 & 0.095 & 0.223 & -0.005 & 0.027 & 0.101 \\
            \midrule
            \multirow{5}{*}{03} & 00 & 1.524 & 0.847 & 0.551 & 0.228 & 1.037 & 0.062 & 0.164 & -0.031 & 0.036 & 0.124 \\
            & 25 & 1.559 & 0.973 & 0.615 & 0.244 & 0.960 & 0.061 & 0.139 & -0.040 & 0.019 & 0.109 \\
            & 50 & 1.293 & 0.737 & 0.496 & 0.222 & 1.215 & 0.050 & 0.203 & -0.055 & -0.019 & 0.071 \\
            & 75 & 1.129 & 0.560 & 0.405 & 0.232 & 0.927 & 0.060 & 0.202 & -0.055 & -0.041 & 0.036 \\
            & 100 & 1.103 & 0.548 & 0.421 & 0.237 & 1.232 & 0.063 & 0.174 & -0.045 & -0.029 & 0.052 \\
            \midrule
            \multirow{5}{*}{04} & 00 & 1.078 & 0.566 & 0.420 & 0.223 & 1.046 & 0.043 & 0.210 & -0.072 & -0.002 & 0.115 \\
            & 25 & 1.348 & 0.806 & 0.546 & 0.239 & 1.045 & 0.085 & 0.199 & -0.052 & -0.012 & 0.065 \\
            & 50 & 1.594 & 0.876 & 0.607 & 0.241 & 1.131 & 0.090 & 0.196 & -0.020 & 0.028 & 0.105 \\
            & 75 & 1.391 & 0.789 & 0.518 & 0.230 & 0.951 & 0.062 & 0.154 & -0.044 & 0.005 & 0.089 \\
            & 100 & 1.307 & 0.704 & 0.483 & 0.252 & 1.288 & 0.088 & 0.199 & -0.005 & 0.030 & 0.109 \\
            \midrule
            \multirow{5}{*}{05} & 00 & 1.108 & 0.620 & 0.439 & 0.252 & 1.007 & 0.083 & 0.179 & -0.038 & 0.032 & 0.135 \\
            & 25 & 1.198 & 0.661 & 0.464 & 0.256 & 1.009 & 0.077 & 0.187 & -0.027 & -0.005 & 0.071 \\
            & 50 & 1.774 & 1.078 & 0.679 & 0.259 & 0.997 & 0.059 & 0.125 & -0.043 & 0.002 & 0.084 \\
            & 75 & 1.752 & 1.100 & 0.698 & 0.241 & 0.935 & 0.051 & 0.109 & -0.058 & -0.002 & 0.084 \\
            & 100 & 1.589 & 0.924 & 0.601 & 0.233 & 1.258 & 0.055 & 0.166 & -0.036 & 0.012 & 0.102 \\
            \midrule
            \multirow{5}{*}{06} & 00 & 1.161 & 0.686 & 0.471 & 0.224 & 0.954 & 0.052 & 0.123 & -0.069 & 0.010 & 0.122 \\
            & 25 & 1.439 & 0.824 & 0.530 & 0.218 & 0.950 & 0.048 & 0.184 & -0.068 & -0.027 & 0.059 \\
            & 50 & 1.359 & 0.814 & 0.531 & 0.226 & 0.951 & 0.046 & 0.159 & -0.070 & -0.031 & 0.045 \\
            & 75 & 1.532 & 0.976 & 0.654 & 0.223 & 0.905 & 0.038 & 0.161 & -0.079 & -0.028 & 0.047 \\
            & 100 & 1.128 & 0.623 & 0.451 & 0.230 & 1.205 & 0.082 & 0.141 & -0.061 & -0.045 & 0.035 \\
            \midrule
            \multicolumn{2}{c}{Average} & 1.317 & 0.749 & 0.516 & 0.244 & 1.063 & 0.073 & 0.185 & -0.037 & 0.005 & 0.090 \\ \bottomrule
        \end{tabular}
        \begin{tablenotes}
            \item[]$\dagger$: Image based method(s), \quad $\ddag$: INR based method, \quad $\S$: Point Cloud based method(s), \quad $\P$: Range Image based method(s).
        \end{tablenotes}
    \end{threeparttable}
    \label{tab:bd-cd-detail}
\end{table*}

\section*{Acknowledgment}
This work was supported by JST\textperiodcentered ASPIRE Grant Number JPMJAP2432 and JSPS KAKENHI Grant Number JP22H03582.

\bibliographystyle{IEEEtran}
\bibliography{./access25_kuwabara}

\begin{IEEEbiography}[{\includegraphics[width=1in,height=1.25in,clip,keepaspectratio]{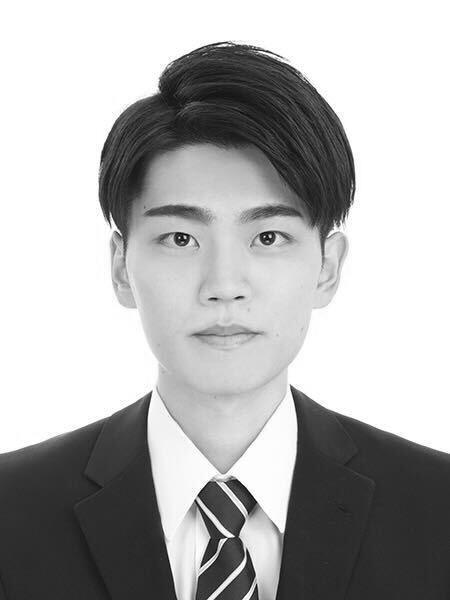}}]{AKIHIRO KUWABARA }
received the B.S.\ degree from Osaka University, Osaka, Japan, in 2024.
He is currently an M.S. student in the Graduate School of Information Science and Technology, Osaka University.
His research interests include point cloud compression and delivery. 
\end{IEEEbiography}

\begin{IEEEbiography}[{\includegraphics[width=1in,height=1.25in,clip,keepaspectratio]{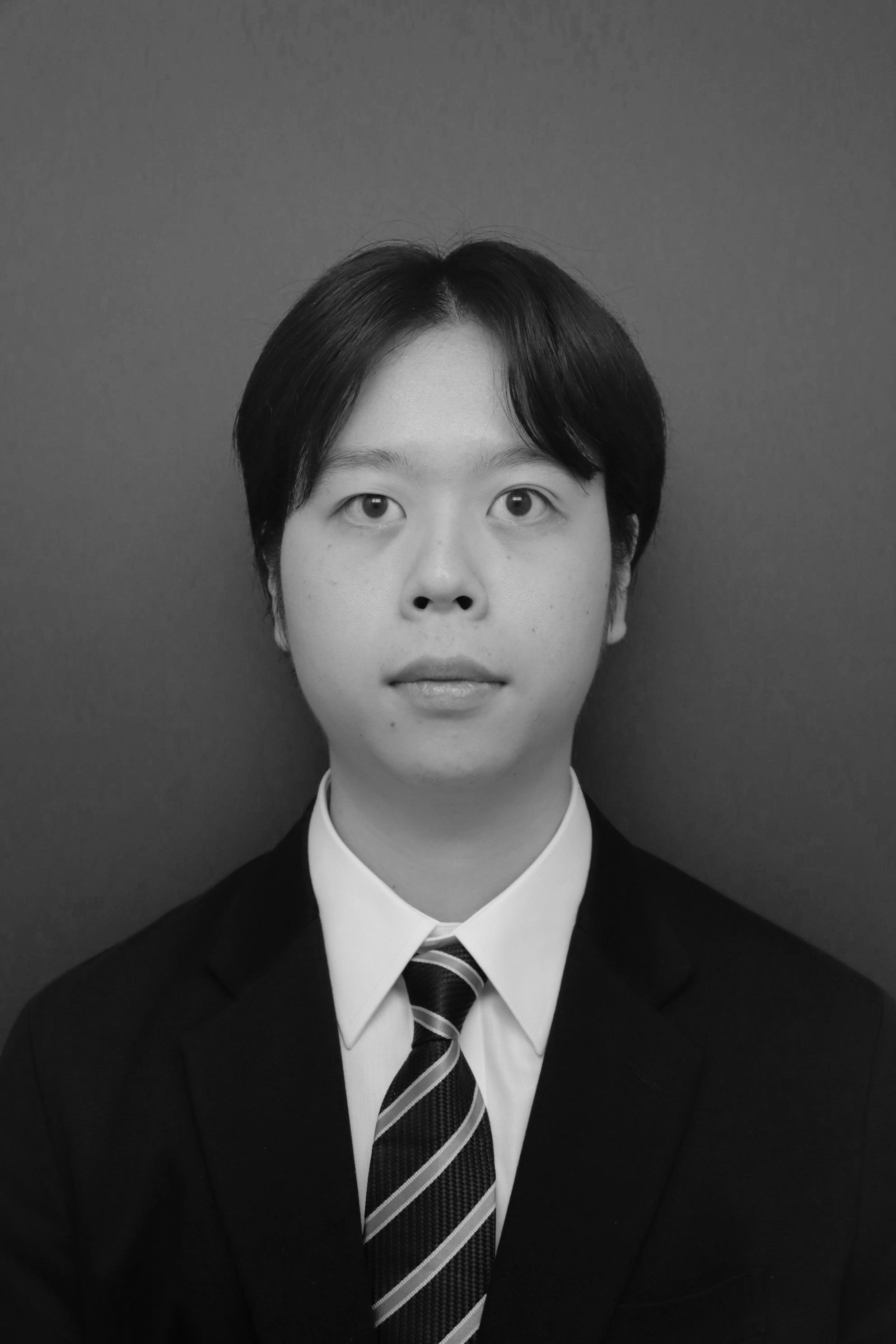}}]{SORACHI KATO }
received B.E. and M.E. degrees from Osaka University, Japan, in 2021 and 2023, respectively. He is currently pursuing his Ph.D. degree in the Graduate School of Information Science and Technology, Osaka University, from April 2023. He is a research fellow~(DC1) of Japan Society for the Promotion of Science from 2023. From 2023 to 2024, he was an intern at Mitsubishi Electric Research Labs.~(MERL) working with the signal processing group. He received the Outstanding Paper Award from the Information Processing Society of Japan~(JSPS) in 2022. His research interests are in the areas of RF sensing, deep neural signal processing, and multimedia neural compression.
\end{IEEEbiography}

\begin{IEEEbiography}{TOSHIAKI KOIKE-AKINO }
(M'05--SM'11) received the B.S.\ degree in electrical and electronics
engineering, M.S.\ and Ph.D.\ degrees in communications and computer
engineering from Kyoto University, Kyoto, Japan, in 2002, 2003, and
2005, respectively. 
During 2006--2010 he was a Postdoctoral Researcher at Harvard University, and is currently a Distinguished Research Scientist at Mitsubishi Electric
Research Laboratories (MERL), Cambridge, MA, USA. 
He received the YRP Encouragement Award 2005, the 21st TELECOM
System Technology Award, the 2008 Ericsson Young Scientist Award, the
IEEE GLOBECOM'08 Best Paper Award in Wireless Communications Symposium,
the 24th TELECOM System Technology Encouragement Award, and the IEEE
GLOBECOM'09 Best Paper Award in Wireless Communications Symposium.
He is a Fellow of Optica.
\end{IEEEbiography}

\begin{IEEEbiography}[{\includegraphics[width=1in,height=1.25in,clip,keepaspectratio]{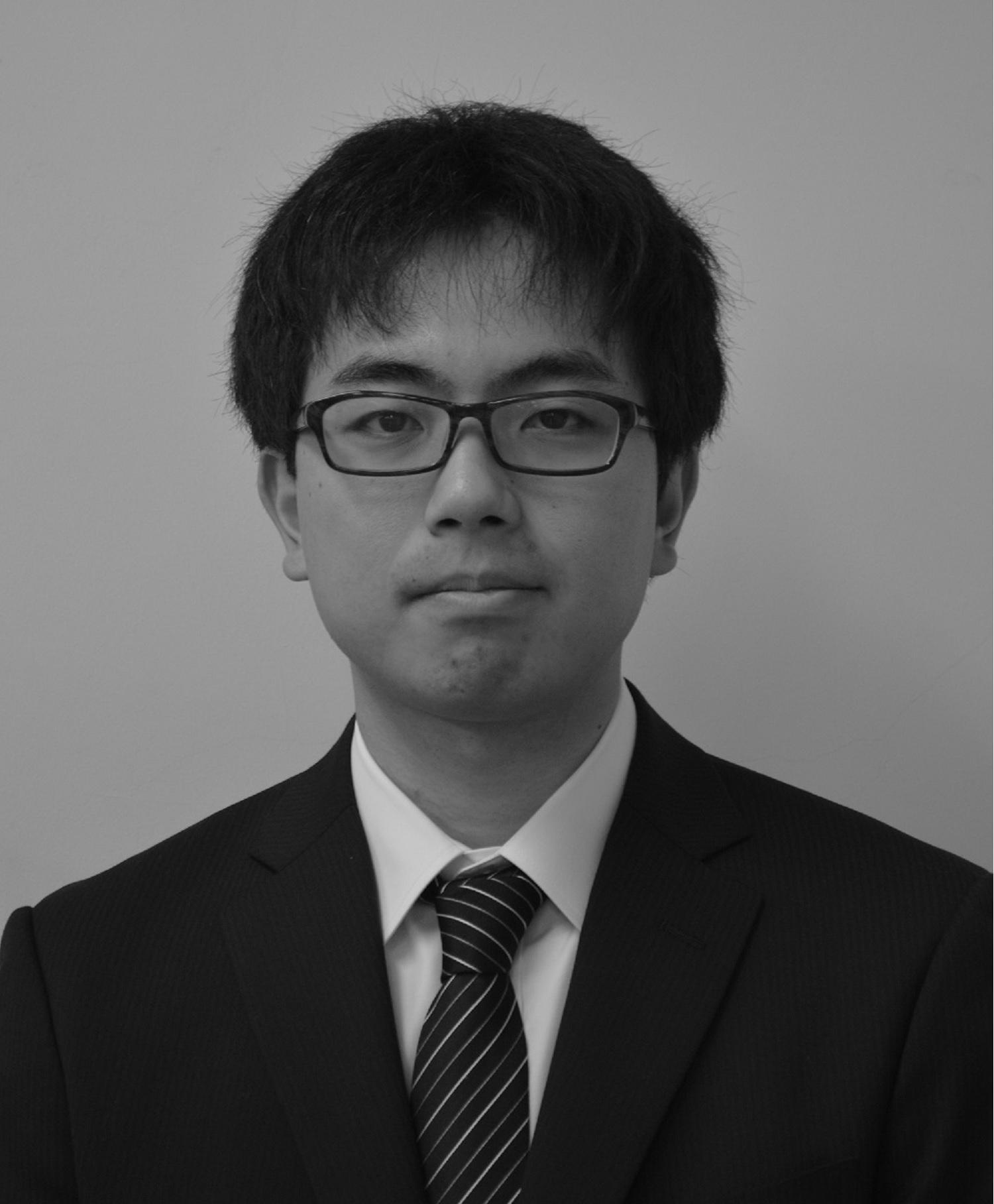}}]{TAKUYA FUJIHASHI } 
received his B.E. degree in
2012 and his M.S. degree in 2013 from Shizuoka
University, Japan. In 2016, he received his Ph.D. degree
from the Graduate School of Information Science
and Technology, Osaka University, Japan. He
is currently an Assistant Professor at the Graduate
School of Information Science and Technology,
Osaka University since April 2019. He was a research
fellow (PD) of Japan Society for the Promotion
of Science in 2016. From 2014 to 2016, he
was a research fellow (DC1) of Japan Society for the Promotion of Science.
From 2014 to 2015, he was an intern at the Mitsubishi Electric Research
Labs. (MERL) working with the Electronics and Communications group.
He was selected as one of the Best Paper candidates in IEEE ICME (International
Conference on Multimedia and Expo) 2012. His research interests are in the
area of video compression and communications, with a focus on multi-view
video coding and streaming.
\end{IEEEbiography}

\EOD

\end{document}